\documentclass[11pt, a4paper, goog]{google}

\usepackage[authoryear, sort&compress, round]{natbib}
\usepackage{multirow}
\usepackage[table]{xcolor}
\usepackage{xcolor}
\usepackage{pifont}
\usepackage{tikz}
\usepackage[most]{tcolorbox}
\usepackage{listings}
\usepackage{xcolor}
\usepackage{xspace}
\usepackage{graphicx}
\usepackage{caption}
\usepackage{subcaption}
\usepackage{booktabs}
\usepackage{multirow}
\usepackage{lineno}
\usepackage{cleveref}
\usepackage{enumitem}
\usepackage{markdown}
\setlist[itemize]{leftmargin=5mm}

\newcommand{\ourmethod}{\textsc{PaperOrchestra}}
\newcommand{\ourdataset}{\texttt{PaperWritingBench}}
\newcommand{\mylatex}{L\kern-.36em\raisebox{0.3ex}{\textsc{a}}\kern-.15em T\kern-.1667em\lower.5ex\hbox{E}\kern-.125em X}

\definecolor{best_color}{HTML}{90EE90}  
\definecolor{second_color}{HTML}{FFFF99}

\definecolor{promptbg}{RGB}{248,248,248}
\definecolor{promptborder}{RGB}{210,210,210}

\newcommand{\cmark}{\textcolor{green!70!black}{\ding{51}}} 
\newcommand{\xmark}{\textcolor{red!80!black}{\ding{55}}}   

\definecolor{rawbg}{HTML}{F4F8FC}       
\definecolor{rawborder}{HTML}{5D8AA8}   

\newtcolorbox{rawmaterialbox}[1][]{%
  enhanced,
  breakable,
  colback=rawbg,
  colframe=rawborder,
  boxrule=0.5pt,
  arc=3pt,
  left=6pt, right=6pt, top=6pt, bottom=6pt,
  title=#1,
  fonttitle=\bfseries\small,
  coltitle=white 
}

\definecolor{humanevalbg}{HTML}{F2F2F2}      
\definecolor{humanevalborder}{HTML}{A6A6A6}  

\newtcolorbox{humanevalbox}[1][]{%
  enhanced,
  breakable,
  colback=humanevalbg,
  colframe=humanevalborder,
  boxrule=0.5pt,
  arc=3pt,
  left=6pt, right=6pt, top=6pt, bottom=6pt,
  title=#1,
  fonttitle=\bfseries\small,
  coltitle=white 
}

\definecolor{leakbg}{HTML}{FDECF2}     
\definecolor{leakborder}{HTML}{E8B4C8} 

\definecolor{paperbg}{HTML}{E8F0ED}       
\definecolor{paperborder}{HTML}{9EB4AD} 

\definecolor{extractbg}{HTML}{F6F0F2}    
\definecolor{extractborder}{HTML}{C5A8B0}

\definecolor{raterbg}{HTML}{FEF5E7}
\definecolor{raterborder}{HTML}{EBD1A7}  

\newtcolorbox{leakpromptbox}[1][]{%
  enhanced,
  breakable,
  colback=leakbg,
  colframe=leakborder,
  boxrule=0.5pt,
  arc=3pt,
  left=6pt,right=6pt,top=6pt,bottom=6pt,
  title=#1,
  fonttitle=\bfseries\small,
  fontupper=\ttfamily\small\raggedright,
  coltitle=white
}

\newtcolorbox{paperpromptbox}[1][]{%
  enhanced,
  breakable,
  colback=paperbg,
  colframe=paperborder,
  boxrule=0.5pt,
  arc=3pt,
  left=6pt,right=6pt,top=6pt,bottom=6pt,
  title=#1,
  fonttitle=\bfseries\small,
  fontupper=\ttfamily\small\raggedright,
  coltitle=white
}

\newtcolorbox{extractpromptbox}[1][]{%
  enhanced,
  breakable,
  colback=extractbg,
  colframe=extractborder,
  boxrule=0.5pt,
  arc=3pt,
  left=6pt,right=6pt,top=6pt,bottom=6pt,
  title=#1,
  fonttitle=\bfseries\small,
  fontupper=\ttfamily\small\raggedright,
  coltitle=white
}

\newtcolorbox{raterpromptbox}[1][]{%
  enhanced,
  breakable,
  colback=raterbg,
  colframe=raterborder,
  boxrule=0.5pt,
  arc=3pt,
  left=6pt,right=6pt,top=6pt,bottom=6pt,
  title=#1,
  fonttitle=\bfseries\small,
  fontupper=\ttfamily\small\raggedright,
  coltitle=white
}

\definecolor{methodbg}{HTML}{F4F0F8}
\definecolor{methodborder}{HTML}{BFA8D0}

\newtcolorbox{methodologybox}[1][]{%
  enhanced,
  breakable,
  colback=methodbg,
  colframe=methodborder,
  boxrule=0.5pt,
  arc=3pt,
  left=6pt, right=6pt, top=6pt, bottom=6pt,
  title=#1,
  fonttitle=\bfseries\small,
  coltitle=white
}
\uselogo{} 

\title{\textls[-2]{\ourmethod: A Multi-Agent Framework} for Automated AI Research Paper Writing}

\correspondingauthor{yiwensong@google.com, yalesong@google.com, jinsungyoon@google.com}

\reportnumber{0001} 

\renewcommand{\today}{}

\author[1]{Yiwen Song}
\author[1]{Yale Song}
\author[1]{Tomas Pfister}
\author[1]{Jinsung Yoon}

\affil[1]{\thepa{}{}}

\begin{abstract}
Synthesizing unstructured research materials into manuscripts is an essential yet under-explored challenge in AI-driven scientific discovery. Existing autonomous writers are rigidly coupled to specific experimental pipelines, and produce superficial literature reviews. We introduce \ourmethod, a multi-agent framework for automated AI research paper writing. It flexibly transforms unconstrained pre-writing materials into submission-ready \mylatex{} manuscripts, including comprehensive literature synthesis and generated visuals, such as plots and conceptual diagrams. To evaluate performance, we present \ourdataset, the first standardized benchmark of reverse-engineered raw materials from 200 top-tier AI conference papers, alongside a comprehensive suite of automated evaluators. In side-by-side human evaluations, \ourmethod{} significantly outperforms autonomous baselines, achieving an absolute win rate margin of \textbf{50\%--68\%} in literature review quality, and \textbf{14\%--38\%} in overall manuscript quality. (Project Page: \url{https://yiwen-song.github.io/paper_orchestra/})
\end{abstract}

\begin{document}

\maketitle

\section{Introduction}

The rapid advancement of Large Language Models (LLMs) is transitioning AI from an assistive tool to an active participant in scientific discovery~\citep{eger2025transforming}. While recent end-to-end autonomous frameworks~\citep{lu2024aiscientist, yamada2025aiscientistv2} establish the feasibility of automated research loops, realizing their full potential is hindered by a critical step: translating unstructured materials, such as raw ideas and experimental logs, into rigorous, submission-ready manuscripts.

Early attempts at automated academic writing relied on the parametric memory of LLMs, often leading to factual hallucinations. To mitigate this, recent frameworks employ retrieval-augmented generation (RAG) methods. Systems like AutoSurvey2~\citep{wu2025autosurvey2} and LiRA~\citep{go2025lira} decompose the literature review process into structured stages or specialized agent roles that emulate human review workflows. However, these survey-specific frameworks lack the capacity to transform raw experimental data into a full-length research paper.

On the other hand, full-lifecycle autonomous research agents are tightly coupled to their experimental loops, preventing them from functioning as standalone writing tools capable of processing human-provided materials. Empirical evaluations show critical deficits in their literature synthesis~\citep{beel2025evaluating, tang2025large}. Relying on simple keyword searches, these agents produce shallow reviews with insufficient citations. They also lack the capabilities to generate conceptual diagrams, restricting visuals to code-generated data plots. Furthermore, evaluating automated writing independently remains difficult due to the absence of a standardized benchmark.

To bridge these gaps, our core contributions are:
\begin{itemize}
    \item \textbf{\ourmethod{}:} A standalone, multi-agent framework that autonomously authors \mylatex{} manuscripts from unconstrained pre-writing materials. It uses specialized agents to synthesize deep literature reviews, generate plots and conceptual diagrams, and iteratively refine the manuscript for better technical clarity.
    \item \textbf{\ourdataset{}:} The first standardized benchmark for AI research paper writing, which isolates the writing task by providing reverse-engineered raw materials (ideas and experimental logs) derived from 200 papers published in top-tier AI conferences.
    \item \textbf{Performance:} In side-by-side human evaluations, \ourmethod{} significantly outperforms autonomous baselines, achieving absolute win rate margins (the difference of our win rate and the baseline's) of \textbf{50\%--68\%} in literature review synthesis and \textbf{14\%--38\%} in overall manuscript quality.
\end{itemize}
\section{Related Work}
\subsection{AI Researcher Frameworks}

Recent works have introduced end-to-end (E2E) frameworks designed to fully automate the scientific research lifecycle. The AI Scientist-v1~\citep{lu2024aiscientist} introduced automated experimentation and drafting via code templates, while v2~\citep{yamada2025aiscientistv2} increases autonomy using agentic tree-search. Concurrent works expand on these paradigms through distinct technical approaches: Cycle Researcher~\citep{weng2024cycleresearcher} employs an iterative refinement loop with reinforcement learning from reviewer feedback, while OmniScientist~\citep{shao2025omniscientist} couples an agentic framework with structured knowledge and tool-based reasoning. Other systems, such as InternAgent~\citep{team2025internagent} and AI Co-Scientist~\citep{gottweis2025towards}, support human-in-the-loop collaboration in domain-specific applications. While effective at autonomous execution, the writing modules in these frameworks are rigidly coupled to their internal experimental loops. Consequently, they cannot directly function as standalone pipelines to synthesize unconstrained, human-provided pre-writing materials into submission-ready manuscripts.

\begin{table}[tbp]
    \centering
    \resizebox{\linewidth}{!}{%
    \setlength{\tabcolsep}{4pt}
    \begin{tabular}{@{} l c c c c c @{}}
        \toprule
        \textbf{Method} & 
        \textbf{\begin{tabular}{@{}c@{}}E2E \mylatex \\ Manuscript Gen.\end{tabular}} & 
        \textbf{\begin{tabular}{@{}c@{}}Decoupled \\ Standalone Writer\end{tabular}} & 
        \textbf{\begin{tabular}{@{}c@{}}Robust to \\ Unstructured Input\end{tabular}} & 
        \textbf{\begin{tabular}{@{}c@{}}Targeted Lit \\ Review Gen.\end{tabular}} & 
        \textbf{\begin{tabular}{@{}c@{}}Conceptual \\ Diagram Gen.\end{tabular}} \\
        \midrule
        PaperRobot \citep{wang2019paperrobot}            & \xmark & \xmark & \xmark & \xmark & \xmark \\
         data-to-paper \citep{technion2024datatopaper}    & \xmark & \cmark & \cmark & \xmark & \xmark \\
        AI-Researcher \citep{tang2025ai}                 & \cmark & \xmark & \xmark & \cmark & \xmark \\
        Cycle Researcher \citep{weng2024cycleresearcher} & \cmark & \xmark & \xmark & \xmark & \xmark \\
        AI Scientist-v2 \citep{yamada2025aiscientistv2}  & \cmark & \xmark & \cmark & \cmark & \xmark \\
        \ourmethod\ \textbf{(Ours)}                      & \cmark & \cmark & \cmark & \cmark & \cmark \\
        \bottomrule
    \end{tabular}
    }
    \caption{Comparison of \ourmethod{} against existing AI writing systems.}
    \label{tab:comparison}
\end{table}

\subsection{Automated Writing and Literature Synthesis}
Early automated writing systems, such as PaperRobot~\citep{wang2019paperrobot}, generated incremental text sequences conditioned on entity relation graphs but lacked the capacity to synthesize complex, data-driven scientific narratives. More recent LLM-based writing assistants, including Prism~\citep{openai_prism_2026}, excel at stylistic refinement and local text editing, but typically rely on structured inputs or human guidance and do not natively support end-to-end manuscript generation from raw experimental logs. Similarly, data-to-paper~\citep{technion2024datatopaper} focuses on translating structured analytical results into text with backward traceability. However, it primarily operates on structured, well-defined inputs and is less suited to flexibly drafting from early-stage, under-specified raw materials.

Beyond constructing a data-driven narrative from experimental results, synthesizing prior literature remains a distinct and critical challenge in automated authoring. Recent work targets this problem via automated literature review generation. For instance, AutoSurvey2~\citep{wu2025autosurvey2} uses a multi-stage retrieval-augmented pipeline to generate comprehensive surveys, while SurveyGen-I~\citep{chen2025surveygen} focuses on iteratively refining topic outlines. Similarly, LiRA~\citep{go2025lira} employs a multi-agent workflow that structures, drafts, and refines literature reviews from retrieved papers. While highly effective at generating long-form surveys, these systems are not designed for writing targeted related work sections. Consequently, they often lack the contextual awareness needed to selectively contrast prior work and clearly motivate the specific research gap of a new method. 

As summarized in Table~\ref{tab:comparison}, existing pipelines show key limitations when used as standalone manuscript generators. Frameworks like CycleResearcher~\citep{weng2024cycleresearcher} require a structured BibTeX reference list as input, which is rarely available at the start of writing, and fail on unstructured inputs outside this format. While AI-Researcher~\citep{tang2025ai} and AI Scientist-v2~\citep{yamada2025aiscientistv2} accept unstructured topic descriptions, their writing modules rely on artifacts produced earlier in their internal pipelines and do not generate conceptual scientific diagrams. In contrast, \ourmethod{} can process unconstrained pre-writing materials without relying on pre-existing plots or ground-truth references. By orchestrating specialized agents to autonomously synthesize targeted literature, generate comprehensive visuals, and craft coherent scientific narratives, \ourmethod{} establishes a robust, end-to-end pipeline for AI research manuscript generation.
\section{Task and Dataset}
\subsection{Task Formulation}
We formulate the end-to-end AI research paper generation task as a function mapping unconstrained pre-writing materials to a complete submission package. Specifically, the framework operates on the following input components:

\begin{itemize}
    \item \textbf{Idea Summary} ($\mathcal{I}$): A brief overview establishing the proposed methodology, core contributions, and theoretical foundation.
    \item \textbf{Experimental Log} ($\mathcal{E}$): A compilation of experimental results, covering raw data points, ablation studies, and performance metrics.
    \item \textbf{LaTeX Template} ($\mathcal{T}$): The template files provided by the target AI conference.
    \item \textbf{Conference Guidelines} ($\mathcal{G}$): The requirements mandated by the target AI conference.
    \item \textbf{Figures} ($\mathcal{F}$): An optional set of pre-existing visual assets (e.g., diagrams, plots). If no figures are provided ($\mathcal{F} = \emptyset$), the pipeline autonomously synthesizes all relevant visuals.
\end{itemize}

The goal is to produce a finalized submission package $P$. We define $P$ as a tuple consisting of the source \mylatex{} file ($P_\textit{tex}$) and the rendered PDF ($P_{\textit{pdf}}$), generated via the framework $W$:
\begin{equation}
    P = (P_\textit{tex}, P_{\textit{pdf}}) = W(\mathcal{I}, \mathcal{E}, \mathcal{T}, \mathcal{G}, \mathcal{F})
\end{equation}

\subsection{Dataset Construction}
To evaluate our pipeline, we introduce \ourdataset{} (App.~\ref{appendix:data_distribution}), a new dataset comprising 200 accepted papers from CVPR 2025 and ICLR 2025 (100 papers each). These venues ensure high academic standards while testing adaptability to distinct conference formats (double-column CVPR vs. single-column ICLR). To construct the evaluation tuple $(\mathcal{I}, \mathcal{E}, \mathcal{T}, \mathcal{G}, \mathcal{F})$ for each paper, we retrieve the official \mylatex{} templates ($\mathcal{T}$) and guidelines ($\mathcal{G}$) for each venue and execute the following pipeline:

\paragraph{Paper Acquisition and Extraction.}
We randomly sample papers from the OpenReview portal (ICLR) and the CVF Open Access repository (CVPR) to ensure topic diversity. We process the raw PDFs using MinerU~\citep{wang2024mineruopensourcesolutionprecise} to generate structured, mathematically faithful Markdown, and PDFFigures 2.0~\citep{pdffigures2} to extract visual entities ($\mathcal{F}$) and captions. Incomplete or misparsed samples are discarded to maintain corpus quality.

\paragraph{Synthesizing Raw Materials.}
Since pre-writing materials (e.g., lab notes) are unavailable, we prompt an LLM to reverse-engineer two core components from the extracted PDF content (App.~\ref{appendix:raw_material_extraction}). To prevent information leakage, both components are fully anonymized (stripping authors and titles) and rendered strictly self-contained by removing all citations, URLs, and explicit figure or table references. We synthesize the following (App.~\ref{appendix:raw_material_data_example}):
\begin{itemize}
    \item \textbf{Idea Summary ($\mathcal{I}$):} Distills the core methodology while explicitly excluding experimental results. We generate a \textit{Sparse} variant (summarizing only high-level ideas) and a \textit{Dense} variant (retaining formal definitions and \mylatex{} equations) to simulate different degrees of user drafting effort.
    \item \textbf{Experimental Log ($\mathcal{E}$):} Extracts a record of experimental setup and empirical findings, including baselines, datasets, metrics, and tabular data. The LLM further de-contextualizes this data by converting visual insights into standalone factual observations, allowing us to test how well the writing system can reconstruct the narrative purely from raw data.
\end{itemize}

\section{\ourmethod}
\begin{figure}[t]
    \centering
    \includegraphics[width=0.95\linewidth]{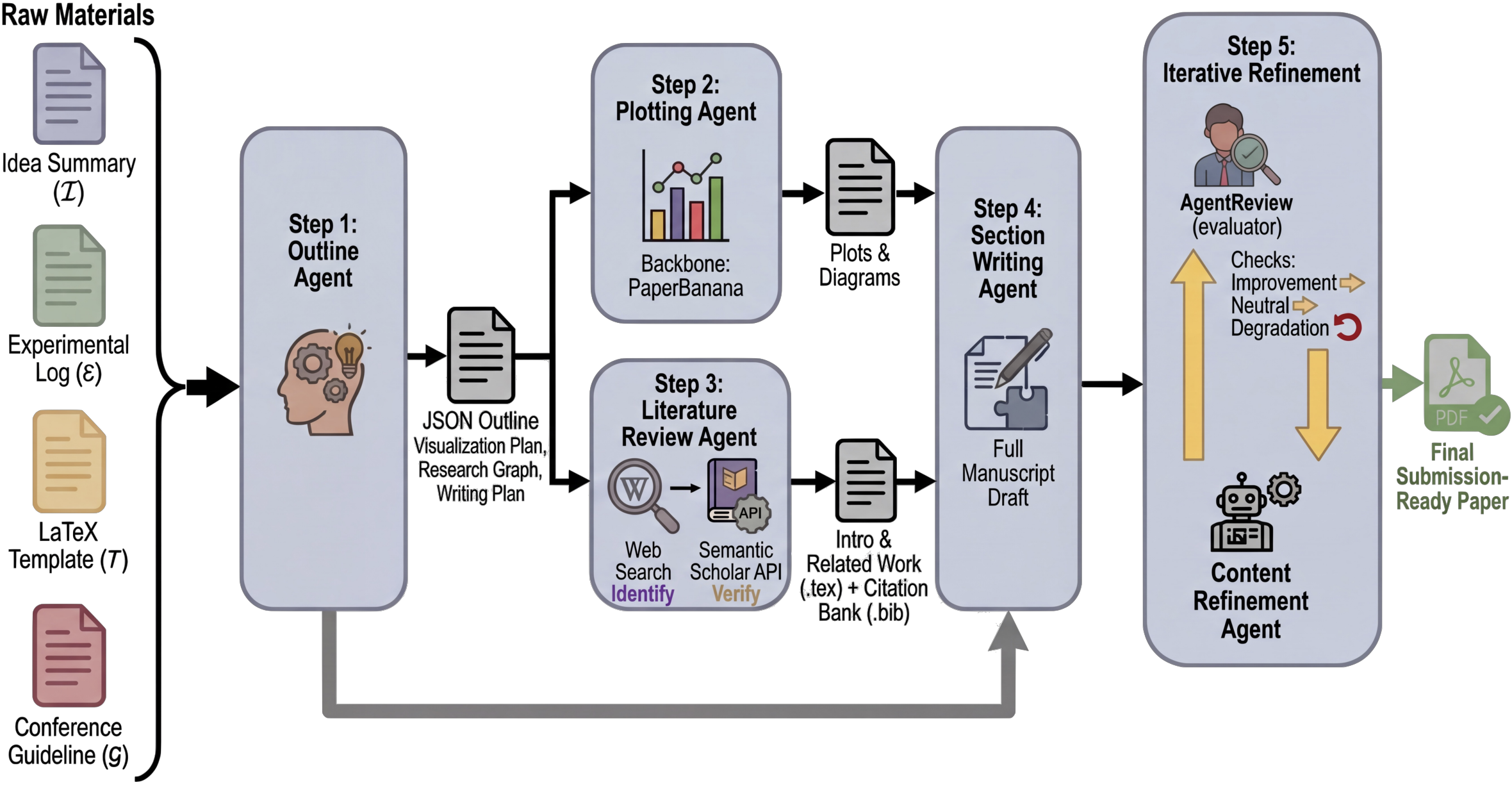}
    \caption[Overview of the \ourmethod{} framework]{Overview of the \ourmethod{} framework. Specialized agents systematically parse unstructured inputs, synthesize plots and literature, compile a full draft, and iteratively refine the manuscript into a submission-ready PDF. (This figure was generated using PaperBanana~\citep{zhu2026paperbanana}.)}
    \label{fig:overview}
\end{figure}

Figure~\ref{fig:overview} illustrates \ourmethod, a multi-agent framework (App.~\ref{appendix:paper_writing_prompts}) that autonomously transforms pre-writing materials, including Idea Summary ($\mathcal{I}$), Experimental Log ($\mathcal{E}$), \mylatex{} Template ($\mathcal{T}$), and Conference Guidelines ($\mathcal{G}$), into submission-ready manuscripts. The pipeline executes five steps, with Step 2 and Step 3 operating in parallel:

\paragraph{Step 1: Outline Generation.} The \textit{Outline Agent} synthesizes pre-writing materials into a JSON outline comprising: (a) a visualization plan dictating specific plot types, data sources, and aspect ratios; (b) a targeted literature search strategy establishing both macro-level context and micro-level methodology clusters to guide the Literature Review Agent in conducting searches and constructing a citation map, which is subsequently leveraged to draft the \textit{Introduction} and \textit{Related Work} sections; and (c) a section-level writing plan including high-level content bullets and a comprehensive list of citation hints for all core external dependencies, such as baselines, datasets, and metrics used in the paper.

\paragraph{Step 2: Plot Generation.} The \textit{Plotting Agent} executes the visualization plan to generate conceptual diagrams and statistical plots. We use \textit{PaperBanana}~\citep{zhu2026paperbanana} as the default module, which employs a closed-loop refinement system where a VLM critic evaluates rendered images against design objectives, iteratively revising text descriptions and regenerating images to resolve visual artifacts, and synthesizing context-aware captions.

\paragraph{Step 3: Literature Review.} Executing the search strategy outlined in Step 1, the \textit{Literature Review Agent} drives a concurrent, hybrid discovery pipeline. It employs an LLM equipped with web search to identify candidate papers, then uses the Semantic Scholar API to authenticate their existence (App.~\ref{appendix:citation_verification}). Upon a successful match, the agent fetches the abstract and metadata while enforcing temporal cutoffs (App.~\ref{appendix:model_details}); candidates are discarded if they exceed the cutoff date or lack a verified mapping. Following deduplication via Semantic Scholar IDs, the agent compiles a citation registry and auto-generates a BibTeX (\texttt{.bib}) file. Finally, it uses the verified citations to draft the \textit{Introduction} and \textit{Related Work} sections.

\paragraph{Step 4: Section Writing.} The \textit{Section Writing Agent} drafts the remaining core sections using the outputs from previous stages. Building upon the partially filled \mylatex{} file, it extracts numeric values from the experimental log to construct tables. Guided by the section-level outline and the established citation bank, the agent authors the abstract, methodology, experiments, and conclusion sections, seamlessly integrating the generated figures to produce a complete \mylatex{} manuscript.

\paragraph{Step 5: Iterative Content Refinement.} The \textit{Content Refinement Agent} iteratively optimizes the manuscript using simulated peer-review feedback. Here we use \textit{AgentReview}~\citep{jin2024agentreview} as the default system in this module. After modifying the \mylatex{} source to address weaknesses, revisions are accepted if the overall score increases, or if it ties while net sub-axis gains are non-negative. The agent immediately reverts to the previous version and halts upon any overall score decrease, negative tie-breaker, or reaching the iteration limit. The resulting \mylatex{} document and compiled PDF represent the final output of \ourmethod.

\section{Experiments}

\subsection{Baselines}

We benchmark \ourmethod{} against two pipelines (App.~\ref{appendix:baseline_selection}): (1) a \textbf{Single Agent} baseline that processes all raw materials ($\mathcal{I}, \mathcal{E}, \mathcal{T}, \mathcal{G}, \mathcal{F}$) and executes end-to-end drafting and bibliography construction in a single LLM call, validating the necessity of a multi-agent system; and (2) \textbf{AI Scientist-v2}~\citep{yamada2025aiscientistv2}, a state-of-the-art system featuring multi-round citation gathering, VLM-guided plot refinement, and iterative self-reflection. To ensure fair comparison against baselines lacking diagram generation, we standardize visual inputs by restricting allowable figures to those extracted from ground-truth papers ($\mathcal{F} = \mathcal{F}_{GT}$). Furthermore, we adopt the \textit{sparse idea} as our standard setting to simulate realistic scenarios where researchers provide preliminary notes rather than mature formulations. To prevent pre-training data memorization, all evaluated writing pipelines employ a universal anti-leakage prompt (App.~\ref{appendix:info_leakage}).

\begin{figure}[tbp]
    \centering
    \includegraphics[width=\linewidth]{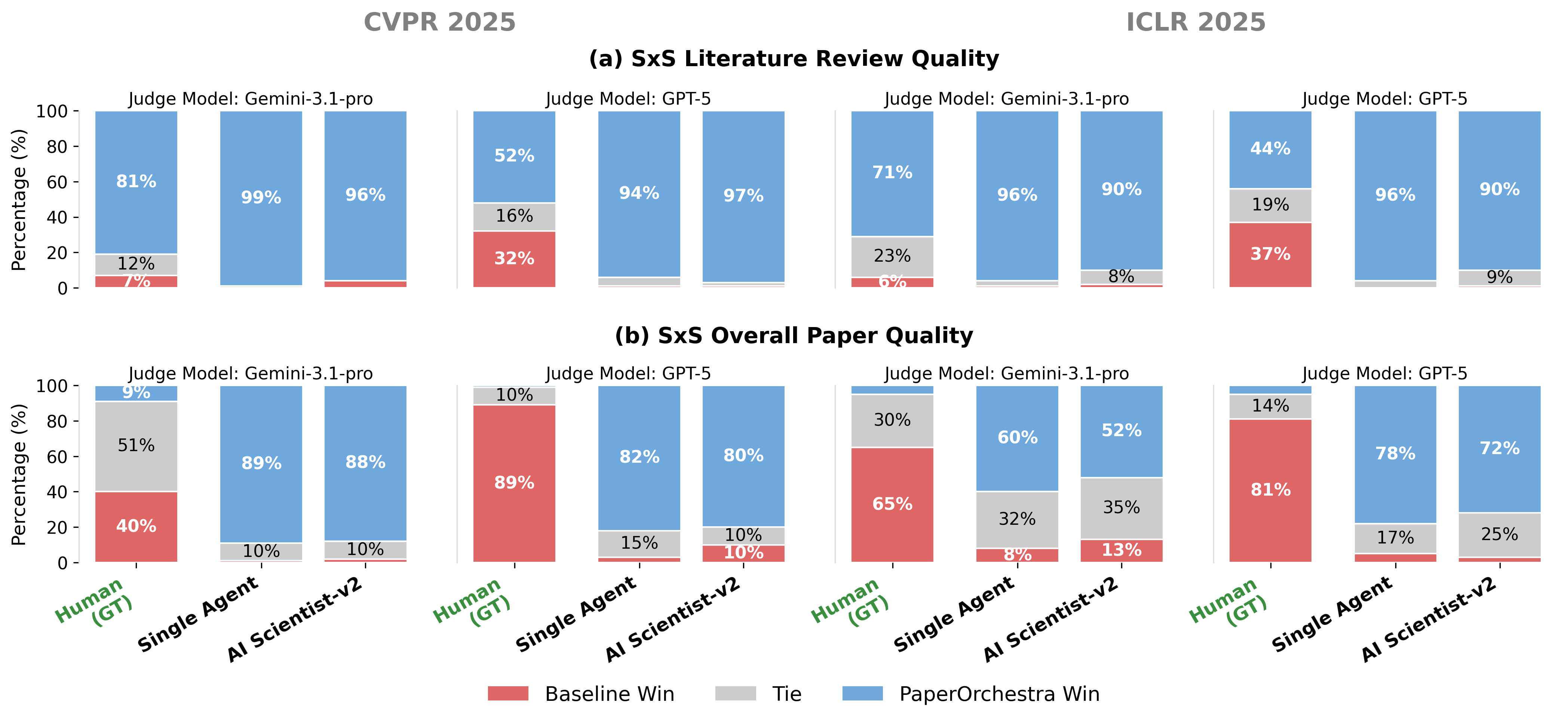} 
    \caption{Automated SxS evaluation of \ourmethod{} against baselines on the CVPR and ICLR 2025 datasets under the sparse idea setting. The \textbf{Human (GT)} baseline serves as an upper-bound reference. For both \textbf{(a)} literature review and \textbf{(b)} overall paper quality, \ourmethod{} significantly outperforms the AI baselines (Single Agent and AI Scientist-v2) across both judge models.}
    \label{fig:sxs_comparison}
\end{figure}

\begin{table}[tbp]
    \centering
    \resizebox{\textwidth}{!}{
    \begin{tabular}{llccccccccc}
        \toprule
        \textbf{Venue} & \textbf{Pipeline} & \begin{tabular}{@{}c@{}}\textbf{Originality} \\ \textbf{(1-4)}\end{tabular} & \begin{tabular}{@{}c@{}}\textbf{Quality} \\ \textbf{(1-4)}\end{tabular} & \begin{tabular}{@{}c@{}}\textbf{Clarity} \\ \textbf{(1-4)}\end{tabular} & \begin{tabular}{@{}c@{}}\textbf{Significance} \\ \textbf{(1-4)}\end{tabular} & \begin{tabular}{@{}c@{}}\textbf{Soundness} \\ \textbf{(1-4)}\end{tabular} & \begin{tabular}{@{}c@{}}\textbf{Presentation} \\ \textbf{(1-4)}\end{tabular} & \begin{tabular}{@{}c@{}}\textbf{Contribution} \\ \textbf{(1-4)}\end{tabular} & \begin{tabular}{@{}c@{}}\textbf{Overall} \\ \textbf{(1-10)}\end{tabular} & \begin{tabular}{@{}c@{}}\textbf{Acceptance} \\ \textbf{Rate (\%)}\end{tabular} \\
        \midrule
        \multicolumn{11}{c}{\cellcolor{gray!15}\textbf{AI Scientist-v2 Reviewer~\citep{yamada2025aiscientistv2}}} \\
        \midrule
        \multirow{4}{*}{CVPR 2025} 
        & Original (GT)   & 2.96 & \textbf{2.81} & \textbf{3.44} & \textbf{3.06} & \textbf{2.74} & \textbf{3.26} & \textbf{2.91} & \textbf{5.95} & \textbf{71.00} \\
        \cmidrule{2-11}
        & SingleAgent      & \cellcolor{best_color}\textbf{3.01} & 2.27 & 2.78 & \cellcolor{best_color}3.03 & 2.26 & 2.79 & \cellcolor{best_color}2.74 & 4.60 & 33.00 \\
        & AI Scientist-v2  & 2.93 & 2.11 & 2.80 & 2.85 & 2.15 & 2.82 & 2.61 & 4.22 & 22.00 \\
        & \ourmethod{}~(PlotOff)     & 2.95 & \cellcolor{best_color}2.43 & \cellcolor{best_color}3.40 & 2.97 & \cellcolor{best_color}2.42 & \cellcolor{best_color}3.22 & \cellcolor{best_color}2.74 & \cellcolor{best_color}5.12 & \cellcolor{best_color}48.00 \\
        \midrule
        \multirow{4}{*}{ICLR 2025} 
        & Original (GT)   & \textbf{2.99} & \textbf{2.79} & \textbf{3.41} & \textbf{3.03} & \textbf{2.76} & \textbf{3.30} & \textbf{2.94} & \textbf{5.81} & \textbf{63.00} \\
        \cmidrule{2-11}
        & SingleAgent     & \cellcolor{best_color}2.70 & 1.67 & 2.20 & 2.44 & 1.76 & 2.19 & 2.09 & 3.22 & 4.00 \\
        & AI Scientist-v2 & 2.58 & 1.76 & 2.27 & 2.46 & 1.82 & 2.29 & 2.16 & 3.42 & 11.00 \\
        & \ourmethod{}~(PlotOff)     & 2.68 & \cellcolor{best_color}2.10 & \cellcolor{best_color}2.82 & \cellcolor{best_color}2.57 & \cellcolor{best_color}2.12 & \cellcolor{best_color}2.79 & \cellcolor{best_color}2.37 & \cellcolor{best_color}4.10 & \cellcolor{best_color}22.00 \\
        \midrule
        \multicolumn{11}{c}{\cellcolor{gray!15}\textbf{ScholarPeer Reviewer~\citep{goyal2026scholarpeer}}} \\
        \midrule
        \multirow{4}{*}{CVPR 2025} 
        & Original (GT)   & \textbf{3.22} & \textbf{3.02} & 3.54 & \textbf{3.63} & 2.87 & 3.39 & \textbf{3.42} & \textbf{7.07} & \textbf{86.00} \\
        \cmidrule{2-11}
        & SingleAgent     & \cellcolor{best_color}\textbf{3.22} & 2.79 & 3.46 & 3.48 & 2.72 & 3.34 & 3.29 & 6.35 & 71.00 \\
        & AI Scientist-v2 & 3.11 & 2.71 & 3.44 & 3.39 & 2.64 & 3.37 & 3.22 & 6.30 & 70.00 \\
        & \ourmethod{}~(PlotOff)  & 3.13 & \cellcolor{best_color}2.95 & \cellcolor{best_color}\textbf{3.75} & \cellcolor{best_color}3.56 & \cellcolor{best_color}\textbf{2.91} & \cellcolor{best_color}\textbf{3.64} & \cellcolor{best_color}3.34 & \cellcolor{best_color}6.93 & \cellcolor{best_color}84.00 \\
        \midrule
        \multirow{4}{*}{ICLR 2025} 
        & Original (GT)   & \textbf{3.33} & \textbf{3.14} & \textbf{3.76} & \textbf{3.74} & \textbf{3.05} & \textbf{3.59} & \textbf{3.57} & \textbf{7.48} & \textbf{94.00} \\
        \cmidrule{2-11}
        & SingleAgent     & 3.26 & 2.73 & 3.27 & 3.48 & 2.71 & 3.20 & 3.25 & 6.35 & 72.00 \\
        & AI Scientist-v2 & 3.23 & 2.62 & 3.28 & 3.41 & 2.60 & 3.23 & 3.11 & 6.04 & 64.00 \\
        & \ourmethod{}~(PlotOff)     & \cellcolor{best_color}3.35 & \cellcolor{best_color}2.96 & \cellcolor{best_color}3.65 &\cellcolor{best_color}3.59 & \cellcolor{best_color}2.94 & \cellcolor{best_color}3.48 & \cellcolor{best_color}3.37 & \cellcolor{best_color}7.03 & \cellcolor{best_color}81.00 \\
        \bottomrule
    \end{tabular}
    }
  \caption{Holistic technical quality evaluation via the AI Scientist-v2~\citep{yamada2025aiscientistv2} and ScholarPeer~\citep{goyal2026scholarpeer} frameworks (Sparse Idea Setting). Higher scores indicate better manuscript quality and higher simulated acceptance rates. \textbf{Bold} indicates best scores; while \colorbox{best_color}{green} highlights the highest score achieved by an AI pipeline.}
  \label{tab:combined_reviewer_quality}
\end{table}
\subsection{Autoraters}

\paragraph{Citation F1.} To quantitatively assess citation coverage, we compare the generated reference list against the ground-truth (GT) paper. Using an LLM provided with the paper context (see App.~\ref{appendix:autorater_prompts}), we partition the GT references into two categories: (1) \textbf{P0 (Must-Cite)} comprises core citations strictly necessary for contextualizing the work, including direct experimental baselines, utilized datasets, metrics, and foundational methodologies upon which the paper directly builds; (2) \textbf{P1 (Good-to-Cite)} includes all remaining GT citations, providing valuable but non-essential background context such as orthogonal work or supplementary information. To ensure accurate matching, we extract reference lists from both the GT and generated papers and resolve them to unique entity IDs using the Semantic Scholar API. We then compute Precision, Recall, and F1 scores independently for the P0 set, the P1 set, and the combined overall reference set.

\paragraph{Literature Review Quality.} To qualitatively evaluate the generated \textit{Introduction} and \textit{Related Work} sections, we employ an LLM evaluator (App.~\ref{appendix:autorater_prompts}) to score the generated text from 0--100 across six axes: coverage and completeness, relevance and focus, critical analysis and synthesis, positioning and novelty, organization, and citation rigor. To mitigate standard LLM score inflation, the evaluation is grounded against venue-specific average citation counts. The system enforces strict score caps and penalties for purely descriptive summaries, unsupported novelty claims, or citation padding, outputting an overall quality score for each paper.

\paragraph{Overall Quality.} To evaluate the holistic technical quality of the generated papers, we employ two AI-based peer review frameworks simulating expert assessment: (1) the \textbf{AI Scientist-v2 Reviewer}~\citep{yamada2025aiscientistv2}, an automated module for structured manuscript evaluation; and (2) \textbf{ScholarPeer}~\citep{goyal2026scholarpeer}, a search-enabled multi-agent system mimicking expert workflows via iterative retrieval and evidence checking. Both systems yield multi-axis scores, an overall rating, and a simulated acceptance decision.

\paragraph{Side-by-Side (SxS) Comparison.} To directly compare manuscripts generated from the same research idea across different pipelines, we implement two automated SxS LLM evaluators. \textbf{(1) SxS Literature Review Quality} extracts the \textit{Introduction} and \textit{Related Work} to assess problem framing, prior work coverage, organization and synthesis, contribution positioning, and readability. \textbf{(2) SxS Paper Quality} holistically compares the full manuscript (including visual layout) across six axes: scientific depth, technical execution, logical flow, writing clarity, presentation of evidence, and academic style. To mitigate LLM positional bias, we evaluate each pair of manuscripts in both orderings. The final aggregated outcome is recorded as a \textit{win} (two wins, or one win and one tie), a \textit{tie} (one win and one loss, or two ties), or a \textit{loss}.

\subsection{Results}
\paragraph{Side-by-Side Comparison.} As shown in \Cref{fig:sxs_comparison}, \ourmethod{} consistently outperforms both the Single Agent baseline and AI Scientist-v2 in SxS evaluations. In \textit{Literature Review Quality}, our framework dominates the autonomous baselines, achieving absolute win margins of \textbf{88\%--99\%}. For \textit{Overall Paper Quality}, although Human (GT) remains the upper bound, \ourmethod{} substantially surpasses all tested AI competitors. It strongly outperforms AI Scientist-v2 and the Single Agent by margins of \textbf{39\%--86\%} and \textbf{52\%--88\%}, respectively, across all settings, confirming that our multi-agent architecture significantly enhances overall manuscript quality (cost analysis in App.~\ref{appendix:cost}).

\paragraph{Technical Quality.} As detailed in \Cref{tab:combined_reviewer_quality}, we evaluate holistic manuscript quality using the AI Scientist-v2 and ScholarPeer frameworks. Under ScholarPeer, \ourmethod{} achieves high simulated acceptance rates of \textbf{84\%} (CVPR) and \textbf{81\%} (ICLR), closely tracking Human (GT) baselines (\textbf{86\%} and \textbf{94\%}, respectively). It significantly outperforms existing AI pipelines, delivering absolute acceptance gains of \textbf{13\%} (CVPR) and \textbf{9\%} (ICLR) over the strongest autonomous baseline. Beyond acceptance rates, \ourmethod{} dominates across critical sub-axes (notably \textit{Clarity}, \textit{Presentation}, and \textit{Soundness}), demonstrating that our multi-agent system yields structurally coherent and technically sound manuscripts.

\begin{table}[tbp]
    \centering
    \resizebox{\textwidth}{!}{%
    \begin{tabular}{ll c c ccc ccc}
        \toprule
        \multirow{2}{*}{\textbf{Venue}} & \multirow{2}{*}{\textbf{Pipeline}} & \multirow{2}{*}{\begin{tabular}{@{}c@{}}\textbf{GT Avg} \\ \textbf{\# Cites}\end{tabular}} & \multirow{2}{*}{\begin{tabular}{@{}c@{}}\textbf{Avg} \\ \textbf{\# Cites}\end{tabular}} & \multicolumn{3}{c}{\textbf{Gemini-3.1-Pro}} & \multicolumn{3}{c}{\textbf{GPT5}} \\
        \cmidrule(lr){5-7} \cmidrule(lr){8-10}
        & & & & \begin{tabular}{@{}c@{}}\textbf{Overall} \\ \textbf{F1}\end{tabular} & \begin{tabular}{@{}c@{}}\textbf{P0} \\ \textbf{Recall}\end{tabular} & \begin{tabular}{@{}c@{}}\textbf{P1} \\ \textbf{Recall}\end{tabular} & \begin{tabular}{@{}c@{}}\textbf{Overall} \\ \textbf{F1}\end{tabular} & \begin{tabular}{@{}c@{}}\textbf{P0} \\ \textbf{Recall }\end{tabular} & \begin{tabular}{@{}c@{}}\textbf{P1} \\ \textbf{Recall}\end{tabular} \\
        \midrule
        \multirow{3}{*}{CVPR 2025} 
        & SingleAgent      & \multirow{3}{*}{58.52} & 11.46 & 28.03 & 60.84 & 2.77 & 28.16 & 63.04 & 3.27 \\
        & AI Scientist-v2  & & 14.18 & 17.15 & 36.12 & 3.26 & 17.26 & 37.46 &  3.30 \\
        & \ourmethod{} (PlotOff)     & & \cellcolor{best_color}47.98 & \cellcolor{best_color}\textbf{29.65} & \cellcolor{best_color}\textbf{63.58} & \cellcolor{best_color}\textbf{15.85} & \cellcolor{best_color}\textbf{29.92} & \cellcolor{best_color}\textbf{65.17} & \cellcolor{best_color}\textbf{16.43} \\
        \midrule
        \multirow{3}{*}{ICLR 2025} 
        & SingleAgent      & \multirow{3}{*}{59.18} & 9.75 & 23.16 & 48.41 & 3.09 & 23.29 & 50.75 & 3.23 \\
        & AI Scientist-v2  & & 13.71 & 15.80 & 33.66 & 3.07 & 15.94 & 34.73 & 3.36  \\
        & \ourmethod{} (PlotOff)     & & \cellcolor{best_color}45.73 & \cellcolor{best_color}\textbf{27.76} & \cellcolor{best_color}\textbf{54.48} & \cellcolor{best_color}\textbf{16.43} & \cellcolor{best_color}\textbf{27.97} & \cellcolor{best_color}\textbf{55.70} & \cellcolor{best_color}\textbf{17.11} \\
        \bottomrule
    \end{tabular}%
    } 
    \caption{Citation F1 evaluation (Sparse Idea Setting). P0 and P1 denote must-cite and good-to-cite references, respectively. F1 and Recall are percentages.}
    \label{tab:citation_f1}
\end{table}
\begin{table*}[tbp]
    \centering
    \resizebox{\textwidth}{!}{
    \begin{tabular}{llccccccc}
        \toprule
        \multirow{2}{*}{\textbf{Venue}} & \multirow{2}{*}{\textbf{Pipeline}} & \textbf{Citation} & \textbf{Coverage \&} & \textbf{Critical Analysis} & \textbf{Organization} & \textbf{Positioning} & \textbf{Relevance} & \textbf{Overall} \\
        & & \textbf{Practices} & \textbf{Completeness} & \textbf{\& Synthesis} & \textbf{\& Writing} & \textbf{\& Novelty} & \textbf{\& Focus} & \textbf{Score} \\
        \midrule
        \multicolumn{9}{c}{{\cellcolor{gray!15}\textit{Judge Model: Gemini-3.1-Pro}}} \\
        \midrule
        \multirow{4}{*}{CVPR 2025} 
        & Human (GT)   & 69.88 & \textbf{71.55} & 65.80 & 74.31 & 75.55 & 78.25 & 71.08 \\
        \cmidrule{2-9}
        & SingleAgent     & 36.83 & 30.10 & 52.93 & 77.03 & 66.45 & 71.40 & 43.04 \\
        & AI Scientist-v2 & 39.24 & 34.25 & 54.83 & 77.21 & 65.16 & 69.12 & 45.43 \\
        & \ourmethod{} (PlotOff)     & \cellcolor{best_color}\textbf{75.36} & \cellcolor{best_color}70.15 & \cellcolor{best_color}\textbf{79.01} & \cellcolor{best_color}\textbf{83.74} & \cellcolor{best_color}\textbf{82.68} & \cellcolor{best_color}\textbf{81.15} & \cellcolor{best_color}\textbf{78.30} \\
        \midrule
        \multirow{4}{*}{ICLR 2025} 
        & Human (GT)   & \textbf{71.66} & \textbf{68.30} & 74.60 & 80.54 & \textbf{82.22} & \textbf{80.67} & 75.18 \\
        \cmidrule{2-9}
        & SingleAgent     & 33.03 & 26.35 & 49.80 & 75.38 & 61.14 & 69.70 & 38.91 \\
        & AI Scientist-v2 & 37.60 &  32.27 & 51.84 & 75.74 & 62.96 & 67.20 & 42.98 \\
        & \ourmethod{} (PlotOff)     & \cellcolor{best_color}71.15 & \cellcolor{best_color}65.71 & \cellcolor{best_color}\textbf{77.34} & \cellcolor{best_color}\textbf{82.02} & \cellcolor{best_color}81.94 & \cellcolor{best_color}80.08 & \cellcolor{best_color}\textbf{76.23} \\
        \midrule
        \multicolumn{9}{c}{{\cellcolor{gray!15}\textit{Judge Model: GPT5}}} \\
        \midrule
        \multirow{4}{*}{CVPR 2025} 
        & Human (GT)   & 63.47 & 63.86 & 57.36 & 76.16 & \textbf{61.34} & \textbf{73.23} & 53.66 \\
        \cmidrule{2-9}
        & SingleAgent     & 49.69 & 47.01 & 54.49 & 76.55 & 56.62 & 70.60 & 44.33 \\
        & AI Scientist-v2 & 49.18 & 47.59 & 53.71 & 75.71 & 55.80 & 69.00 & 43.57 \\
        & \ourmethod{} (PlotOff)     & \cellcolor{best_color}\textbf{63.88} & \cellcolor{best_color}\textbf{64.13} & \cellcolor{best_color}\textbf{60.04} & \cellcolor{best_color}\textbf{77.46} & \cellcolor{best_color}59.18 & \cellcolor{best_color}72.65 & \cellcolor{best_color}\textbf{53.99} \\
        \midrule
        \multirow{4}{*}{ICLR 2025} 
        & Human (GT)   & 61.61 & 60.91 & 58.75 & 77.19 & \textbf{62.91} & 73.11 & \textbf{55.53} \\
        \cmidrule{2-9}
        & SingleAgent     & 49.34 & 45.97 & 54.10 & 77.79 & 56.47 & 71.52 & 44.30 \\
        & AI Scientist-v2 & 49.55 & 48.26 & 54.18 & 76.62 & 56.08 & 69.86 & 44.29 \\
        & \ourmethod{} (PlotOff)     & \cellcolor{best_color}\textbf{62.54} & \cellcolor{best_color}\textbf{62.12} & \cellcolor{best_color}\textbf{60.58} & \cellcolor{best_color}\textbf{78.70} & \cellcolor{best_color}60.21& \cellcolor{best_color}\textbf{73.58} & \cellcolor{best_color}54.15 \\
        \bottomrule
    \end{tabular}
    }
    \caption{Literature review quality assessed by LLM-as-a-Judge (0--100 scale, Sparse Idea Setting). All numbers are percentages.}
    \label{tab:lit_review_quality}
\end{table*}

\paragraph{Citation Coverage.} \Cref{tab:citation_f1} demonstrates \ourmethod{} successfully mitigates under-citation issues. While autonomous baselines achieve competitive Overall F1 scores, this is a mathematical artifact of their extremely low citation counts (averaging 9--14). They focus on fetching obvious P0 papers, which inflates their F1 scores but leaves P1 Recall near zero. In contrast, our framework generates \textbf{45.73--47.98} citations, closely mirroring Human (GT) writeups ($\sim$59). Alongside improving P0 Recall by absolute margins of \textbf{2.13\%--6.07\%}, \ourmethod{} significantly increases P1 Recall by \textbf{12.59\%--13.75\%} over the strongest baselines. This proves our method actively explores the broader academic landscape rather than relying on shallow keyword matching.

\paragraph{Literature Review Quality.} Citation F1 alone cannot fully evaluate literature synthesis; a review must also identify gaps and motivate the proposed method. From \Cref{tab:lit_review_quality}, \ourmethod{} achieves absolute Overall Score gains of \textbf{32.87\%--33.25\%} (Gemini-3.1-Pro) and \textbf{9.66\%--9.85\%} (GPT-5) over the strongest AI baseline, remaining highly comparable to Human baselines. Dominating in \textit{Citation Practices} and \textit{Critical Analysis}, our framework synthesizes analytical, well-grounded narratives rather than generic LLM summaries.

\subsection{Human Evaluation}

\begin{figure}[tbp]
    \centering
    
    \begin{subfigure}[b]{0.58\textwidth}
        \centering
        \includegraphics[width=\linewidth]{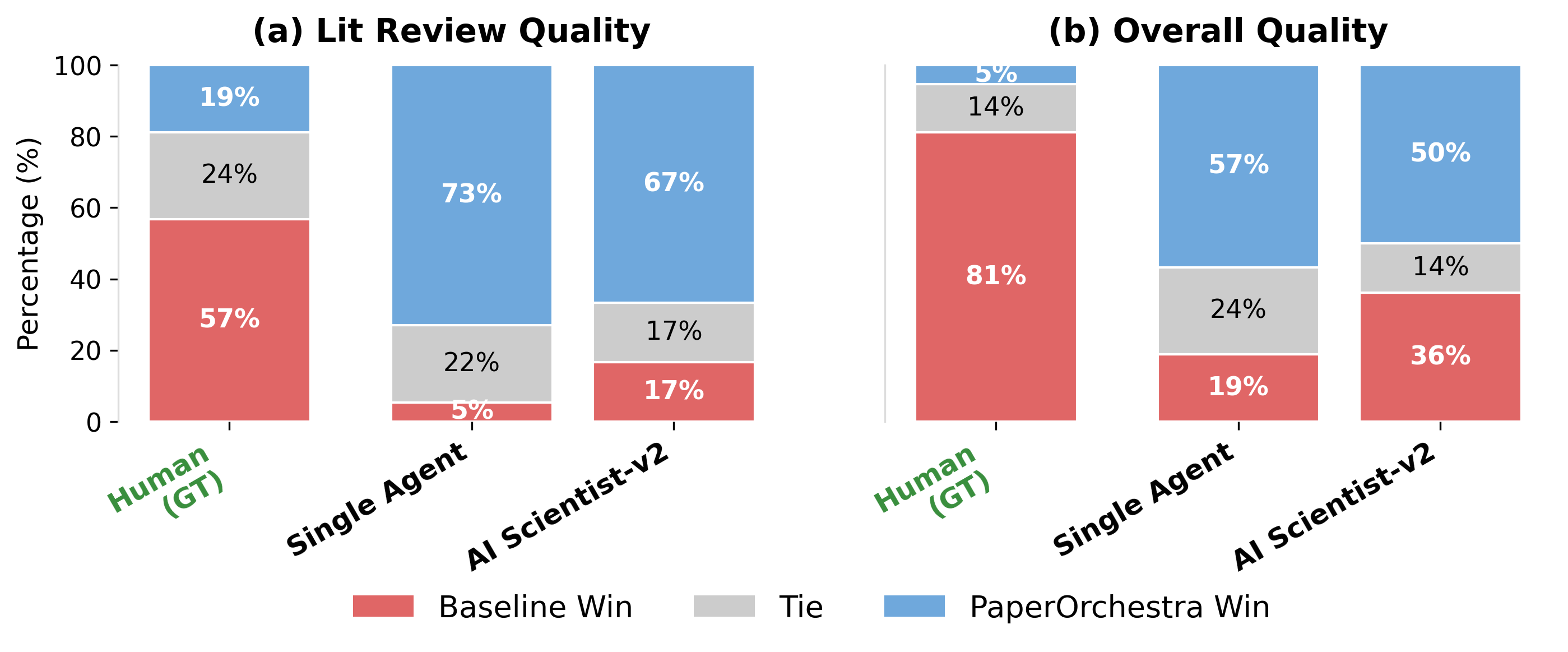}
        \caption{Human SxS Evaluation Results}
        \label{fig:human_sxs_a}
    \end{subfigure}
    \hfill
    \begin{subfigure}[b]{0.4\textwidth}
        \centering
        \resizebox{\linewidth}{!}{%
        \begin{tabular}{llcc}
            \toprule
            \textbf{Metric} & \textbf{Judge(s)} & \textbf{Pearson} & \textbf{Spearman} \\
            \midrule
            \multicolumn{4}{c}{\cellcolor{gray!15}\textit{Human vs. Autorater Correlation}} \\
            \midrule
            \multirow{2}{*}{Lit Review Quality} 
            & Gemini-3.1-Pro & 0.0868 & 0.0715 \\
            & GPT-5 & \textbf{0.2764} & \textbf{0.2300} \\
            \midrule
            \multirow{2}{*}{Overall Quality} 
            & Gemini-3.1-Pro & 0.5672 & 0.5727 \\
            & GPT-5 & \textbf{0.6458} & \textbf{0.6355} \\
            \midrule
            \multicolumn{4}{c}{\cellcolor{gray!15}\textit{Inter-Autorater Correlation (Gemini-3.1-Pro vs. GPT-5)}} \\
            \midrule
            Lit Review Quality & \multicolumn{1}{c}{--} & 0.5697 & 0.5615 \\
            Overall Quality & \multicolumn{1}{c}{--} & \textbf{0.7395} & \textbf{0.7454} \\
            \bottomrule
        \end{tabular}%
        }
        \vspace{0.5em} 
        \caption{Autorater Validation Metrics}
        \label{fig:autorater_correlation_b}
    \end{subfigure}
    
    \caption{Human \& Autorater Evaluation (Sparse Idea Setting). \textbf{(a)} Human preferences for \ourmethod{} vs.\ baselines. \textbf{(b)} GPT-5 strongly correlates with human scores on overall quality, alongside high inter-autorater consistency between distinct LLM judges.}
    \label{fig:human_eval_combined}
\end{figure}

To validate automated SxS metrics and assess human-perceived manuscript quality, we recruited 11 AI researchers to evaluate 40 randomly sampled papers (20 per venue) from \ourdataset{}. For each paper, \ourmethod{} was compared against all baselines (Human GT, Single Agent, and AI Scientist-v2), yielding 180 paired evaluations (App.~\ref{appendix:human_eval}).

Human preferences correlate strongly with our GPT-5 evaluator for Overall Quality (Pearson $r=0.6458$, Spearman $\rho=0.6355$; Figure~\ref{fig:autorater_correlation_b}). Literature review correlation is lower due to inherent LLM self-bias. Manual inspection reveals LLMs tend to act as structural graders, rewarding rigid formatting such as explicit "Problem-Gap-Solution" paragraphs or bulleted thematic groups. In contrast, human experts prioritize dense, pragmatic factuality and a nuanced, narrative-driven flow, which the autorater often mistakenly penalizes for lacking explicit formatting cues. Nevertheless, relative rankings remain strictly consistent. The stability of our automated SxS metrics is further validated by high inter-autorater agreement across different LLM judges (Figure~\ref{fig:autorater_correlation_b}). Human SxS evaluations (Figure~\ref{fig:human_sxs_a}) confirm \ourmethod{} strongly outperforms AI baselines by absolute margins of \textbf{50\%--68\%} (Literature Review) and \textbf{14\%--38\%} (Overall Quality). It also achieves a highly competitive \textbf{43\%} tie/win rate against human GT in literature synthesis.

\subsection{Ablation Studies}

\begin{table}[tbp]
    \centering
    \resizebox{0.8\linewidth}{!}{%
    \begin{tabular}{ll ccc c ccc}
        \toprule
        \multirow{2}{*}{\textbf{Venue}} & \multirow{2}{*}{\textbf{Metric}} & \multicolumn{3}{c}{\textbf{Gemini-3.1-Pro}} && \multicolumn{3}{c}{\textbf{GPT-5}} \\
        \cmidrule{3-5} \cmidrule{7-9}
        & & \textbf{Sparse Win} & \textbf{Tie} & \textbf{Dense Win} && \textbf{Sparse Win} & \textbf{Tie} & \textbf{Dense Win} \\
        \midrule
        \multirow{2}{*}{CVPR 2025} 
        & Paper Quality & 18\% & 26\% & \cellcolor{best_color}56\% && 24\% & 32\% & \cellcolor{best_color}44\% \\
        & Lit Review    & 37\% & 24\% & \cellcolor{best_color}39\% && \cellcolor{best_color}40\% & 22\% & 38\% \\
        \midrule
        \multirow{2}{*}{ICLR 2025} 
        & Paper Quality & 24\% & 31\% & \cellcolor{best_color}45\% && 20\% & 37\% & \cellcolor{best_color}43\% \\
        & Lit Review    & 33\% & 33\% & \cellcolor{best_color}34\% && \cellcolor{best_color}32\% & 40\% & 28\% \\
        \bottomrule
    \end{tabular}%
    }
    \caption{Automated SxS evaluation of Sparse vs.\ Dense idea settings.}
    \label{tab:ablation_sparse_dense}
\end{table}
\begin{table}[tbp]
    \centering
    \resizebox{0.8\linewidth}{!}{%
    \begin{tabular}{l ccc c ccc}
        \toprule
        \multirow{2}{*}{\textbf{Venue}} & \multicolumn{3}{c}{\textbf{Gemini-3.1-Pro}} && \multicolumn{3}{c}{\textbf{GPT-5}} \\
        \cmidrule{2-4} \cmidrule{6-8}
        & \textbf{PlotOff Win} & \textbf{Tie} & \textbf{PlotOn Win} && \textbf{PlotOff Win} & \textbf{Tie} & \textbf{PlotOn Win} \\
        \midrule
        CVPR 2025 & \cellcolor{best_color}34\% & 32\% & 34\% && \cellcolor{best_color}49\% & 19\% & 32\% \\
        ICLR 2025 & \cellcolor{best_color}35\% & 37\% & 28\% && \cellcolor{best_color}36\% & 38\% & 26\% \\
        \bottomrule
    \end{tabular}%
    }
    \caption{Automated SxS Paper Quality evaluating the Plotting Agent. Despite lacking access to data represented by GT visuals, \textit{PlotOn} achieves highly competitive performance against \textit{PlotOff} that use human-authored figures.}
    \label{tab:ablation_plotting}
\end{table}

\paragraph{Sparse vs.\ Dense Idea}
\Cref{tab:ablation_sparse_dense} evaluates the impact of input density. For \textit{Overall Paper Quality}, the Dense idea setting dominates win rates (\textbf{43\%--56\%} vs.\ \textbf{18\%--24\%}) by facilitating more rigorous methodology generation. Meanwhile, \textit{Literature Review Quality} exhibits near-parity, with the Sparse setting securing \textbf{32\%--40\%} against Dense's \textbf{28\%--39\%}. This demonstrates the exceptional robustness of \ourmethod: our \textit{Literature Review Agent} autonomously executes targeted searches and identifies research gaps without relying heavily on the dense input drafted by human. Additional qualitative comparisons are provided in App.~\ref{appendix:paper_viz_sparse_dense}.

\paragraph{Autonomous Visual Generation (PlotOff vs.\ PlotOn)}
\Cref{tab:ablation_plotting} evaluates our \textit{Plotting Agent} by comparing papers using human-authored GT figures (\textit{PlotOff}) vs. autonomously generated visuals (\textit{PlotOn}). Although \textit{PlotOff} benefits from an inherent information advantage (GT figures often embed supplementary results absent from raw input logs), \textit{PlotOn} remains highly competitive, securing ties or wins in \textbf{51\%--66\%} of SxS matchups across all four setups. As shown in App.~\ref{appendix:paper_viz_sxs}, \ourmethod{} synthesizes coherent visuals from scratch, effectively augmenting the manuscript and reinforcing the scientific narrative.

\begin{figure}[tbp]
    \centering    \includegraphics[width=\linewidth]{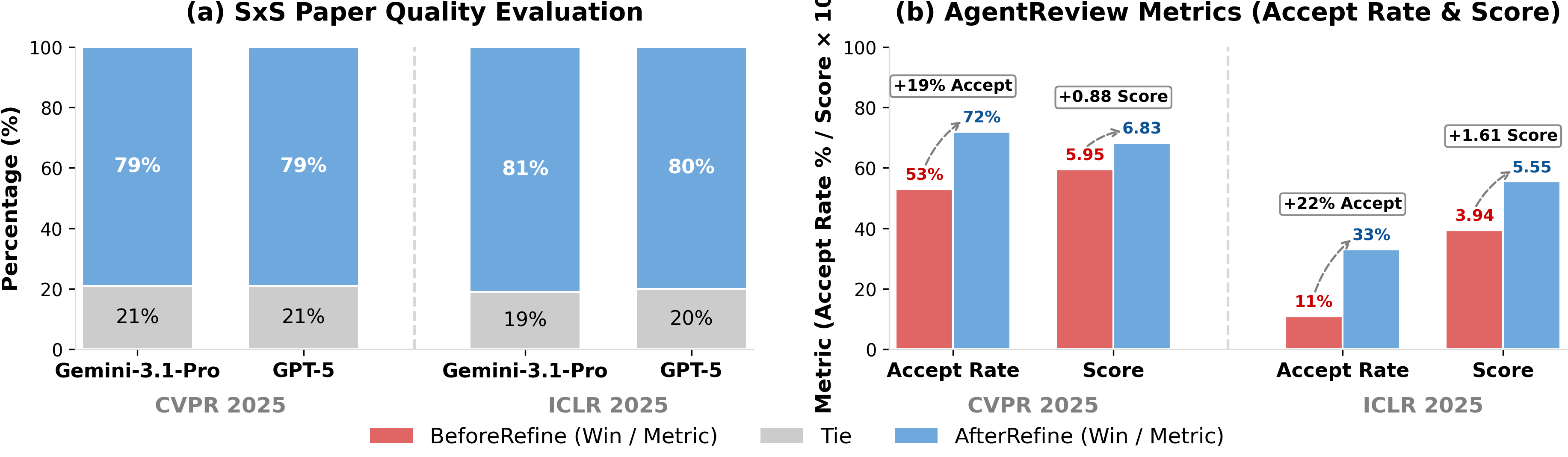}
    \caption{\textbf{Impact of the Content Refinement Agent.} (a) In automated SxS paper quality evaluation, refined manuscripts dominates. (b) This improvement is reflected through absolute gains in acceptance rates simulated by AgentReview~\citep{jin2024agentreview}.}
    \label{fig:ablation_refinement}
\end{figure}

\paragraph{Importance of the Content Refinement Agent}
\Cref{fig:ablation_refinement} isolates the impact of our iterative evaluation-feedback loop. In automated SxS comparisons (\ref{fig:ablation_refinement}(a)), refined manuscripts (\textit{AfterRefine}) dominate unrefined drafts (\textit{BeforeRefine}) with \textbf{79\%--81\%} win rates and \textbf{0\%} losses. AgentReview metrics (\ref{fig:ablation_refinement}(b)) reflect this trend, showing substantial increases in simulated acceptance rates (\textbf{+19\%} CVPR, \textbf{+22\%} ICLR) and overall scores (\textbf{+0.88} and \textbf{+1.61}). Driven by targeted clarity and presentation corrections, these results prove our content refinement loop is critical for elevating raw drafts into rigorous, submission-ready manuscripts.
\section{Conclusion}
\label{sec:conclusion}
In this work, we introduced \ourmethod{} and \ourdataset{} to transform unstructured, preliminary AI research materials into submission-ready manuscripts. Experiments demonstrate that our multi-agent framework can generate high-quality research papers with competitive runtime (App.~\ref{appendix:cost}) while synthesizing deep, context-aware literature reviews. Future research can focus on expanding the framework's capacity to ingest richer research artifacts and transitioning these autonomous agents into interactive, dynamic writing environments to enable seamless human-AI scientific collaboration (more limitations discussed in App.~\ref{appendix:limitations}).
\section*{Ethics Statement}
\label{sec:ethics}

We position our system as an advanced assistive tool designed to accelerate the drafting process of AI research papers, rather than an independent entity capable of claiming authorship. Human researchers must retain full accountability for the factual accuracy, originality, and validity of the claims presented in any generated manuscript. While \ourmethod{} incorporates robust programmatic safeguards (such as API-grounded citation validation) to minimize hallucinations and ensure academic rigor, users are responsible for verifying the outputs to prevent the propagation of LLM-derived biases or misinformation.

\bibliography{main}

\clearpage
\appendix
\section{Limitations and Future Work}
\label{appendix:limitations}

While \ourmethod{} advances the capabilities of autonomous manuscript generation, several areas remain for future exploration. First, relying on external frameworks like PaperBanana~\citep{zhu2026paperbanana} for visual generation limits our direct control over figure hallucinations. Although mitigating these errors falls outside the primary scope of our pipeline, future systems could benefit from integrating targeted Vision-Language Models and dedicated human evaluations to systematically verify that generated visual content is factually sound and optimally placed within the \mylatex{} layout. 

Second, although our current refinement agent effectively improves paper quality using structured, LLM-generated feedback, transitioning this framework toward an interactive, human-in-the-loop (HITL) system would enable researchers to iteratively steer drafts via natural language critiques. This evolution would solidify the framework's intended role as an advanced assistive tool rather than a fully independent writing entity. 

Finally, evaluating foundation models on \ourdataset{} carries an inherent risk of pretraining data contamination. To mitigate this risk, we rigorously de-contextualized and anonymized all input materials. Additionally, since all evaluated baselines share the same underlying language models, any potential knowledge leakage applies uniformly, ensuring a strictly fair relative comparison. To fully isolate generation from memorization, future benchmarks could leverage unpublished research or autonomously generated raw materials.
\section{Computational Cost}
\label{appendix:cost}

\begin{table}[ht]
\centering
\begin{tabular}{lccm{5cm}}
\toprule
\textbf{Method} & \textbf{\#LLM Calls} & \textbf{Latency (Mean)} & \textbf{System Description} \\
\midrule
Single Agent & 1 & 3.3 mins & Single monolithic prompt and generation. \\
\addlinespace
AI Scientist-v2 & $\sim$40--45 & 35.1 mins & Multi-step pipeline (20-round citation scan, drafting, $3\times$ reflections, and vision critique). \\
\addlinespace
\ourmethod{} & $\sim$60--70 & 39.6 mins & Multi-agent orchestration (outline generation, parallel literature search, sequential citation verification, writing, plotting with critique, and a $3\times$ content refinement loop). \\
\bottomrule
\end{tabular}
\caption{Comparison of computational cost and single-paper latency. This table outlines the estimated number of LLM API calls, mean execution time, and core architectural components for each pipeline.}
\label{tab:runtime_efficiency}
\end{table}

Table~\ref{tab:runtime_efficiency} compares the computational cost and single-paper latency of \ourmethod{} against the Single Agent and AI Scientist-v2 baselines. Although the Single Agent executes rapidly, it fails to produce the rigorous citations and data-grounded visuals generated by our system. Despite requiring more LLM calls ($\sim$60--70) than AI Scientist-v2 ($\sim$40--45), \ourmethod{} maintains a highly competitive mean processing time of 39.6 minutes (compared to 35.1 minutes for AI Scientist-v2). To calculate these mean latency values, we evaluated all methods on a random subset of 10 papers (5 from CVPR and 5 from ICLR) using a single worker. For each method, we removed the highest and lowest execution times to mitigate the impact of API network anomalies before computing the final average.

The efficiency of \ourmethod{} is achieved by strategically decoupling the paper discovery and verification pipeline. To optimize throughput, we execute retrieval in two distinct stages: (1) \textbf{Parallel Candidate Discovery}, which leverages 10 concurrent workers for search-grounded LLM calls to rapidly pool candidate papers; and (2) \textbf{Sequential Citation Verification}, which safely processes the pooled candidates through Semantic Scholar at the maximum allowable rate (1 query per second). This architecture successfully combines the high-concurrency tolerance of the LLM API with the strict throughput limits of the Semantic Scholar API to prevent quota-induced latency.

The LLM calls are distributed across our multi-agent system as follows:
\begin{itemize}
    \item \textbf{Outline Agent} (1 call): Initial manuscript structuring and planning.
    \item \textbf{Hybrid Literature Agent} ($\sim$20--30 calls): Execution of the decoupled paper discovery and citation verification pipeline described above.
    \item \textbf{Plotting Agent} ($\sim$20--30 calls): Few-shot retrieval, visual planning, image generation, VLM-guided critique and redraw cycles, and context-aware captioning.
    \item \textbf{Section Writing Agent} (1 call): A single, comprehensive multimodal call to draft and compile the complete \mylatex{} manuscript.
    \item \textbf{Content Refinement Agent} ($\sim$5--7 calls): Score-driven, iterative reflection and manuscript revision.
\end{itemize}

Ultimately, this computational investment is well justified. \ourmethod{} delivers substantial improvements in overall manuscript quality and citation depth, while adding only minimal latency compared to the AI Scientist-v2 baseline.
\section{Dataset Details}
\label{appendix:dataset_details}

\subsection{Data Distribution}
\label{appendix:data_distribution}

\begin{figure}[htbp]
    \centering
    \begin{subfigure}[b]{0.32\textwidth}
        \centering
        \includegraphics[width=\linewidth]{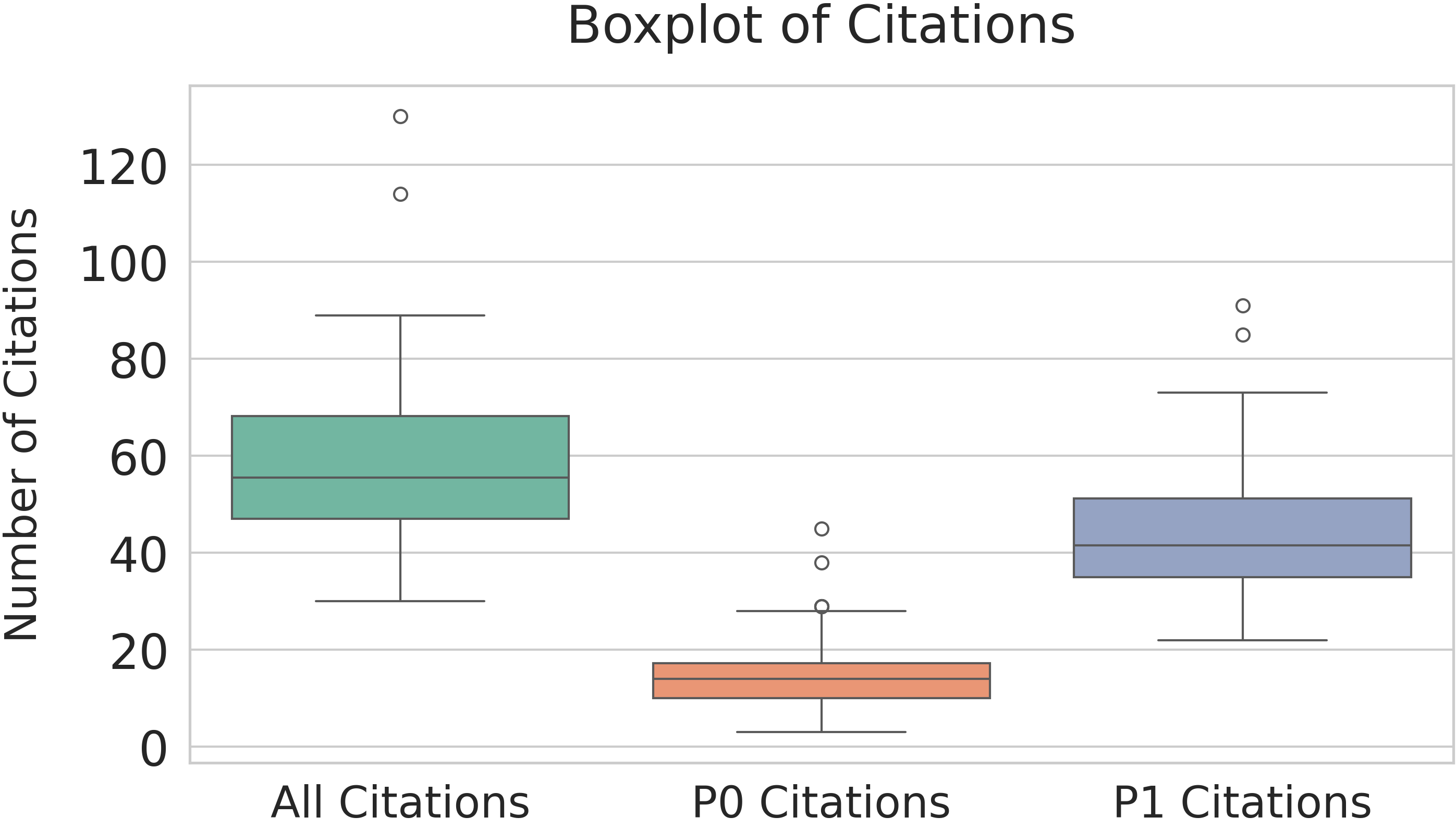}
        \caption{Citation Count Distribution}
    \end{subfigure}
    \hfill
    \begin{subfigure}[b]{0.32\textwidth}
        \centering
        \includegraphics[width=\linewidth]{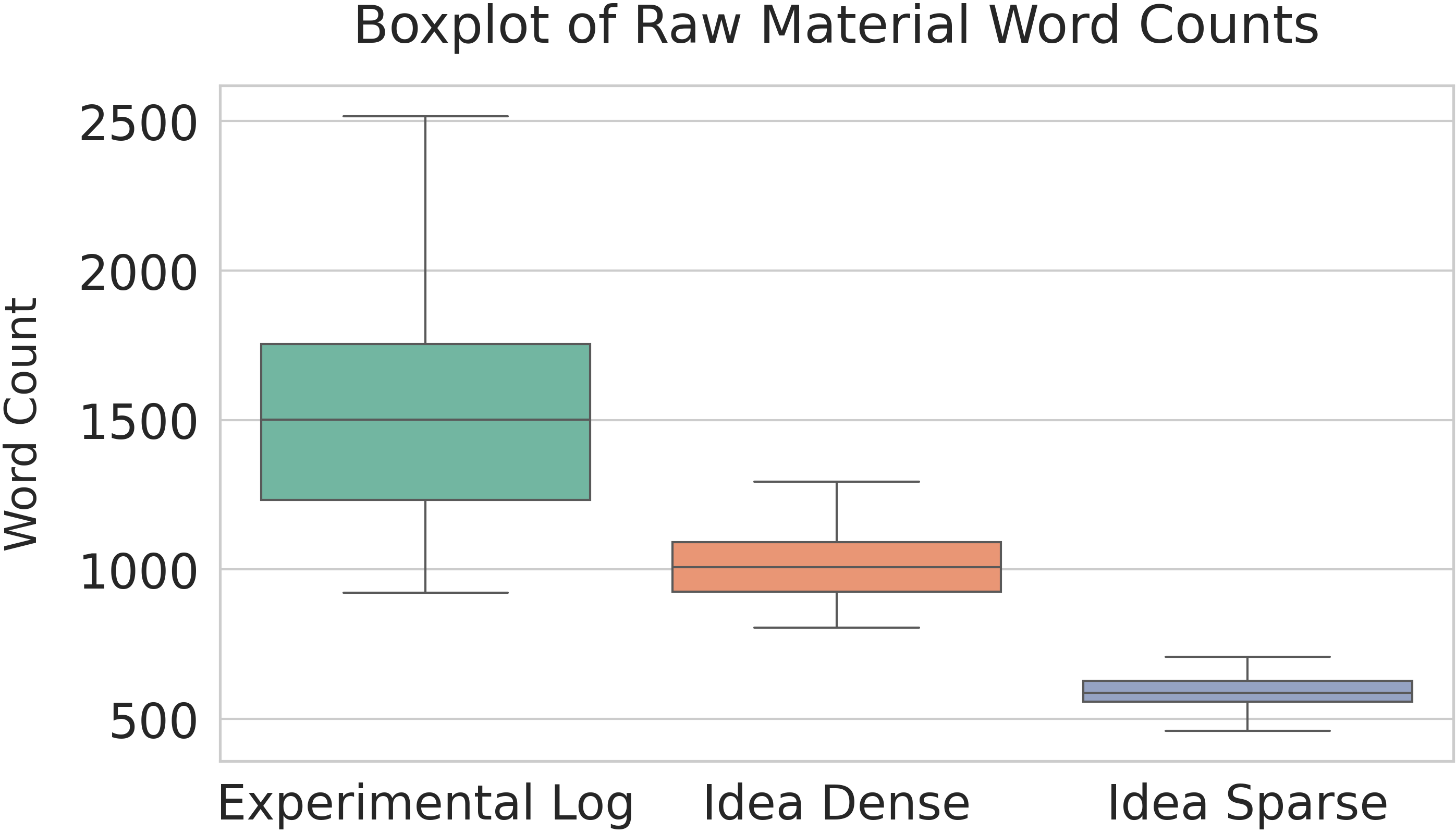}
        \caption{Material Length Distribution}
    \end{subfigure}
    \hfill
    \begin{subfigure}[b]{0.32\textwidth}
        \centering
        \includegraphics[width=\linewidth]{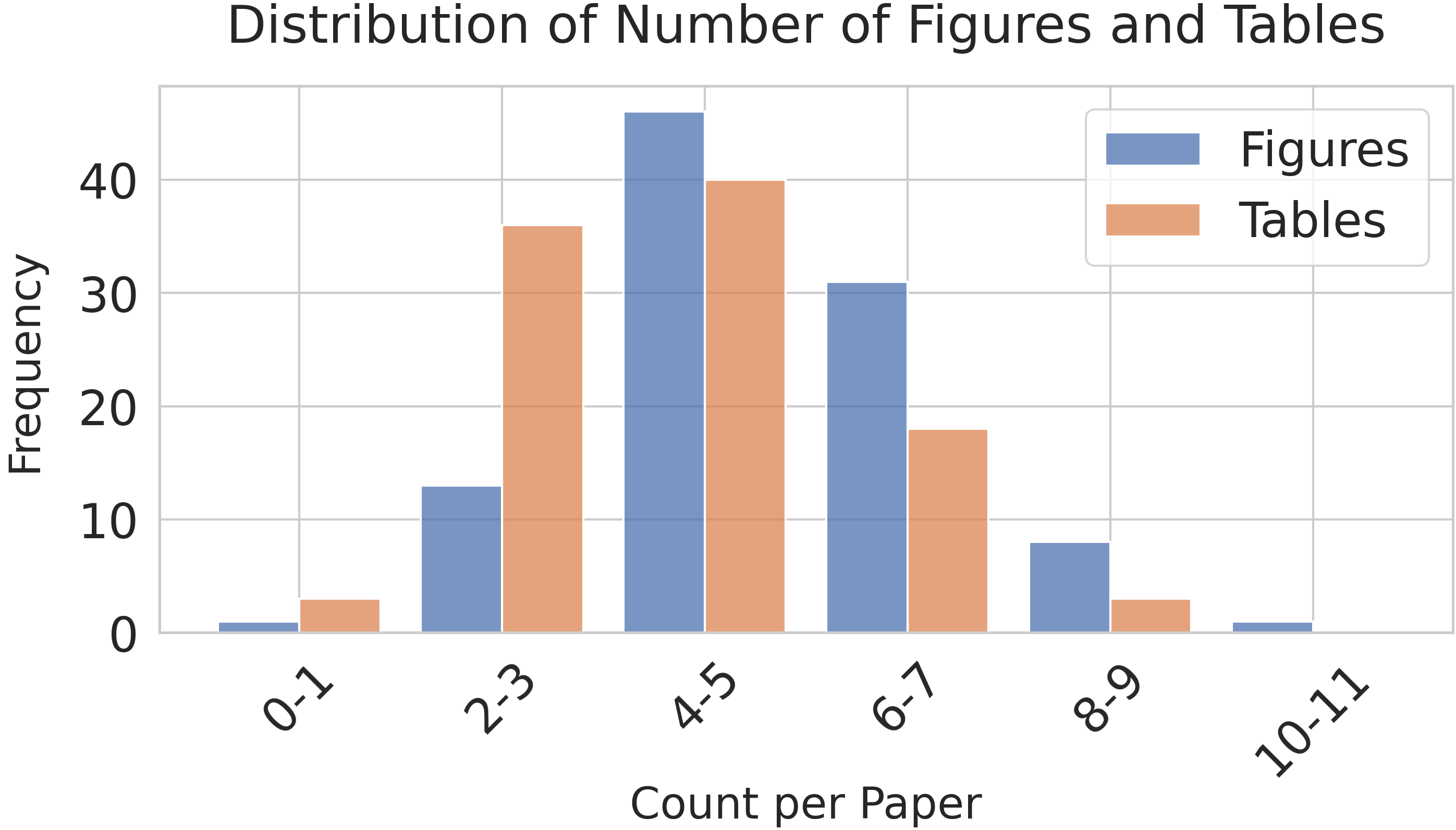}
        \caption{Tables/Figures Distribution}
    \end{subfigure}
    
    \caption{CVPR 2025 Dataset Statistics}
    \label{fig:cvpr2025_distribution}
\end{figure}

\begin{figure}[htbp]
    \centering
    \begin{subfigure}[b]{0.32\textwidth}
        \centering
        \includegraphics[width=\linewidth]{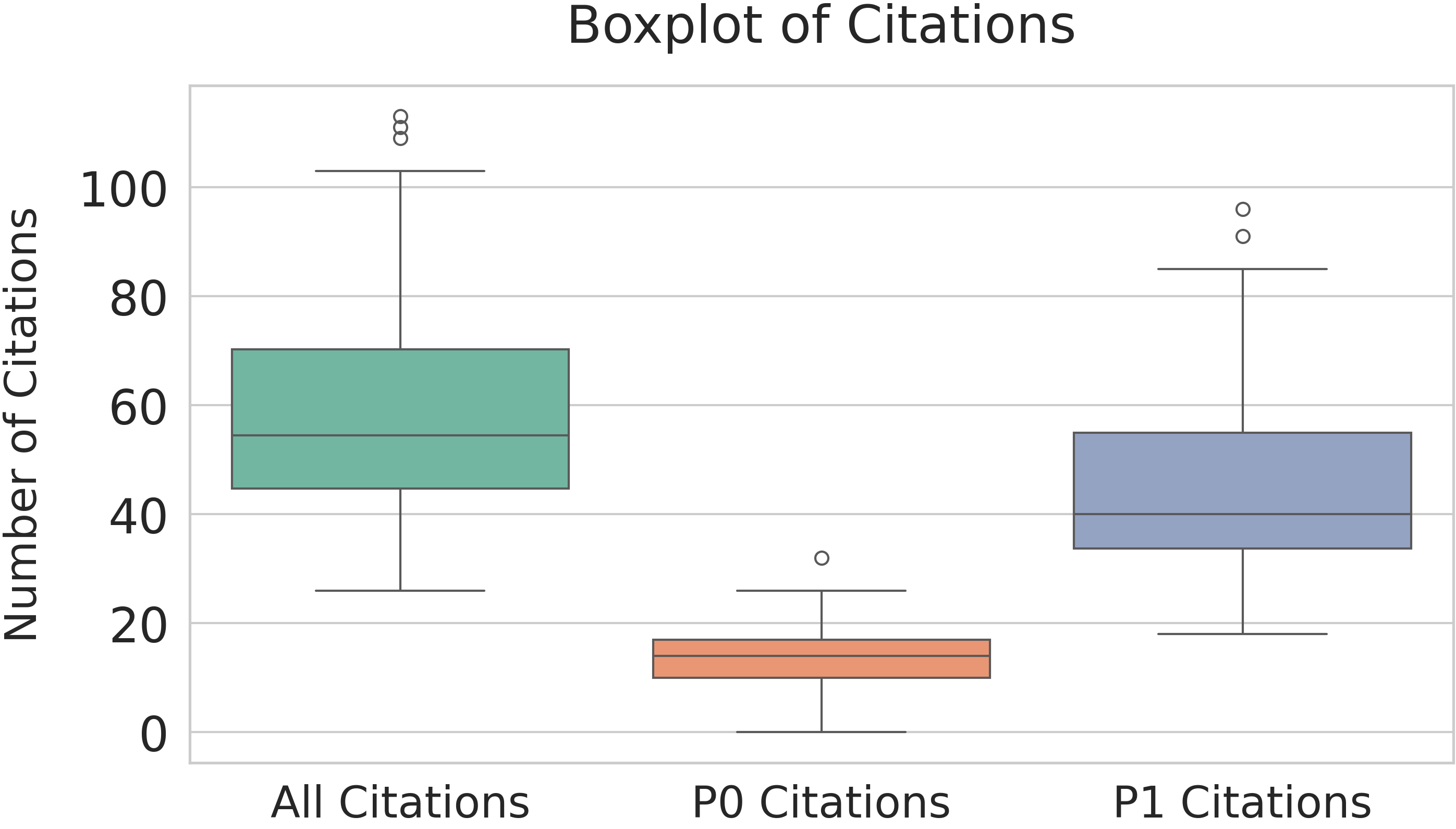}
        \caption{Citation Count Distribution}
    \end{subfigure}
    \hfill
    \begin{subfigure}[b]{0.32\textwidth}
        \centering
        \includegraphics[width=\linewidth]{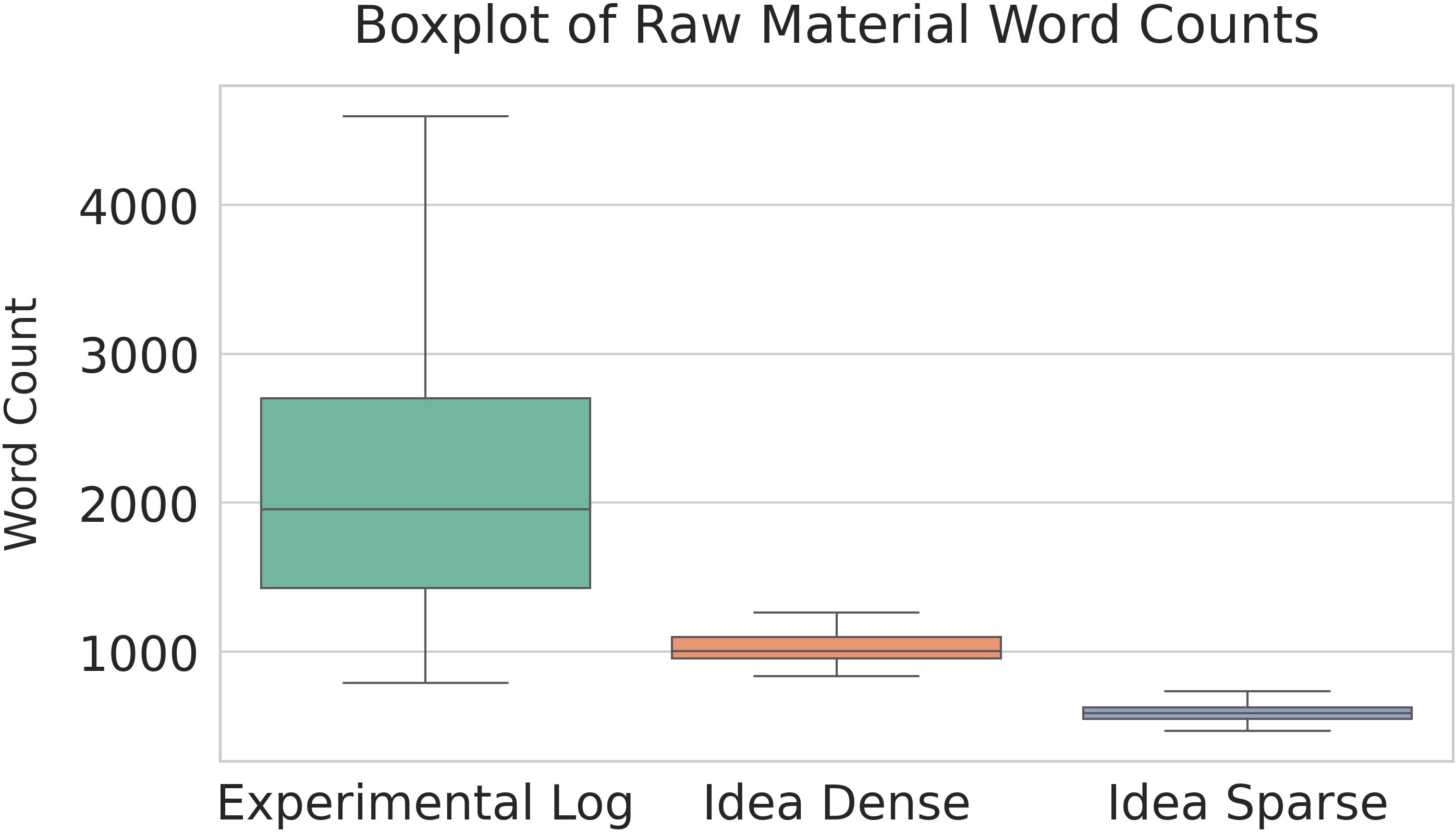}
        \caption{Material Length Distribution}
    \end{subfigure}
    \hfill
    \begin{subfigure}[b]{0.32\textwidth}
        \centering
        \includegraphics[width=\linewidth]{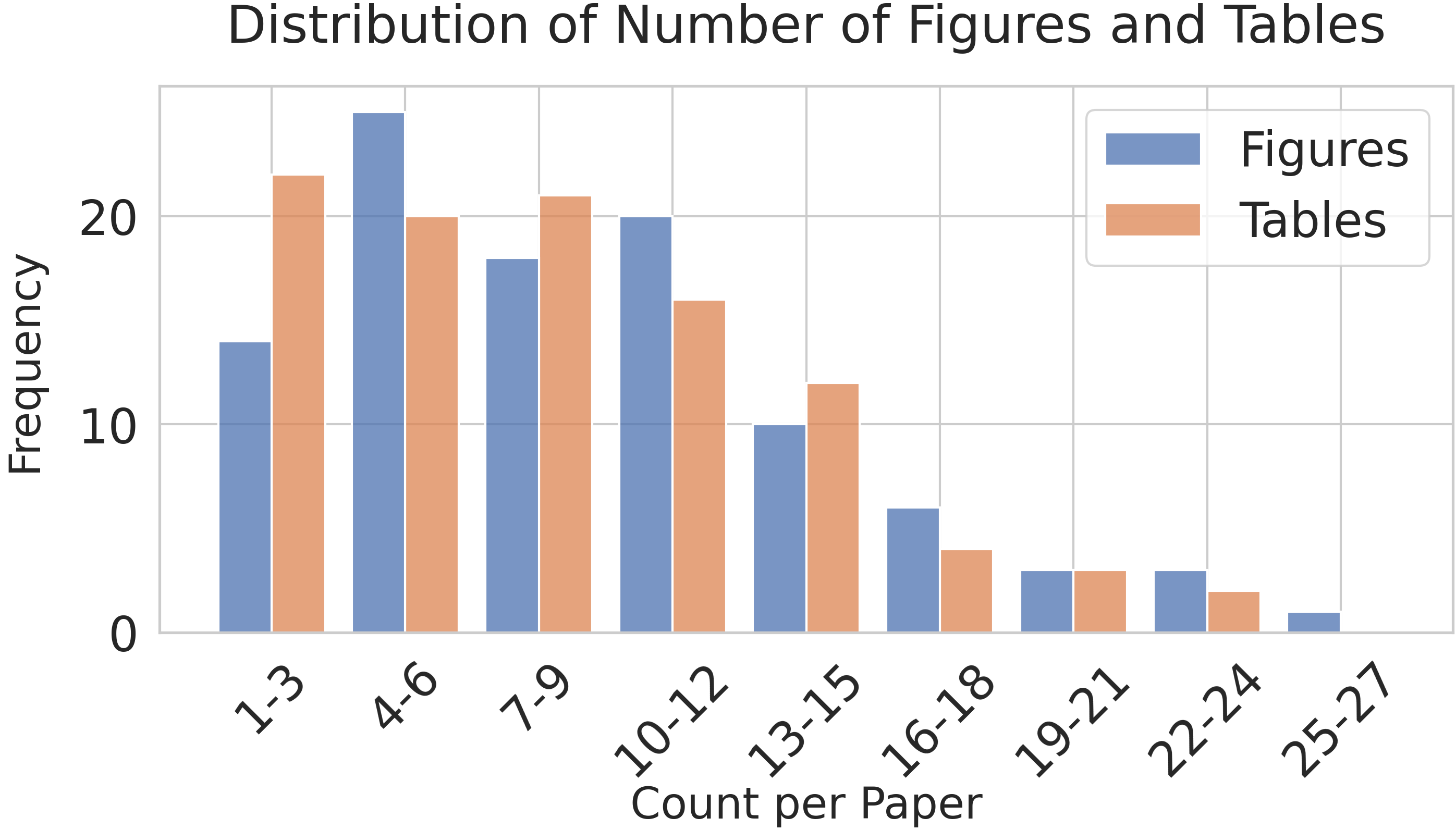}
        \caption{Tables/Figures Distribution}
    \end{subfigure}
    \caption{ICLR 2025 Dataset Statistics}
    \label{fig:iclr2025_distribution}
\end{figure}

\begin{table}[h]
\centering
\resizebox{0.75\linewidth}{!}{
\begin{tabular}{lcc}
\toprule
\textbf{Metric} & \textbf{CVPR 2025} & \textbf{ICLR 2025} \\
\midrule
\textit{Visual Elements} & & \\
Number of Figures & 5.20 $\pm$ 1.73 & 9.19 $\pm$ 5.39 \\
Number of Tables & 4.20 $\pm$ 1.65 & 8.13 $\pm$ 5.19 \\
\midrule
\textit{Citations} & & \\
Total Citations & 58.52 $\pm$ 17.55 & 59.18 $\pm$ 20.01 \\
P0 Citations (must-cite) & 14.86 $\pm$ 6.72 & 13.65 $\pm$ 5.57 \\
P1 Citations (good-to-cite) & 43.66 $\pm$ 13.43 & 45.53 $\pm$ 17.36 \\
\midrule
\textit{Raw Material Word Counts} & & \\
Experimental Log ($\mathcal{E}$) & 1529.91 $\pm$ 364.87 & 2386.71 $\pm$ 1484.56 \\
Dense Idea Summary ($\mathcal{I}_{\text{Dense}}$) & 1082.82 $\pm$ 263.45 & 1056.78 $\pm$ 165.62 \\
Sparse Idea Summary ($\mathcal{I}_{\text{Sparse}}$) & 591.42 $\pm$ 52.68 & 586.96 $\pm$ 53.81 \\
\bottomrule
\end{tabular}
}
\caption{Detailed statistical breakdown of the \ourdataset{} dataset across CVPR 2025 and ICLR 2025. Values are reported as mean $\pm$ standard deviation.}
\label{tab:appendix_dataset_stats}
\end{table}

In this section, we detail the data distribution of \ourdataset{}. As shown in Figure~\ref{fig:cvpr2025_distribution}, Figure~\ref{fig:iclr2025_distribution} and Table~\ref{tab:appendix_dataset_stats}, ICLR 2025 papers exhibit higher visual and analytical density than CVPR 2025, averaging roughly twice as many figures (9.19 vs.\ 5.20) and tables (8.13 vs.\ 4.20). Consequently, ICLR experimental logs are substantially longer (2,387 vs.\ 1,530 words). To ensure controlled evaluations, reverse-engineered input materials remain strictly consistent across both venues: Dense idea files average 1,083 (CVPR) and 1,057 (ICLR) words, while Sparse idea files are constrained to 591 (CVPR) and 587 (ICLR) words. Both datasets maintain comparable citation distributions, averaging ${\sim}59$ total references per manuscript.

\subsection{Raw Material Extraction}
\label{appendix:raw_material_extraction}

The raw materials for \ourdataset{} are extracted using Gemini-3.1-Pro (prompts detailed in App.~\ref{appendix:material_gen_prompts}). To systematically control the granularity of the provided concepts while ensuring the empirical data is extracted with high fidelity and minimal leakage, our pipeline employs two key mechanisms:

\begin{itemize}
    \item \textbf{Controlled Idea Density (Dense vs.\ Sparse):} The \textit{Dense} variant is prompted to explicitly preserve all mathematical formulations, loss functions, and \mylatex{} variables from the source text. Conversely, the \textit{Sparse} variant abstracts away mathematical notation, describing formal architectures through high-level conceptual narratives.

    \item \textbf{Structured Context Injection:} Naive text-only extraction often corrupts tabular and mathematical data. To ensure the \textit{Experimental Log} maintains strict fidelity to the original paper's content, we first parse the source PDF into markdown using MinerU~\citep{wang2024mineruopensourcesolutionprecise}. We independently extract visual assets and their original captions using PDFFigures~2.0~\citep{pdffigures2}. Rather than relying solely on unstructured markdown, we inject the extracted images of all tables and figures as multi-modal context blocks into Gemini-3.1-Pro. Sourced with this direct visual context during extraction, the model is then requested to flatten all explicit document references (e.g., stripping ``as shown in Figure 2'') and translate visual information into standalone factual observations, forcing downstream agents to autonomously reconstruct the empirical narrative.
\end{itemize}

\subsection{Raw Material Example}
\label{appendix:raw_material_data_example}

In this section, we present a complete data sample comprising the sparse idea ($\mathcal{I}_{\text{sparse}}$), dense idea ($\mathcal{I}_{\text{dense}}$), and experimental log ($\mathcal{E}$) extracted from a randomly selected paper~\citep{11093026} in the CVPR 2025 split.

\begin{rawmaterialbox}[Sparse Idea ($\mathcal{I}_{sparse}$)]
\noindent\textbf{Problem Statement}

The Segment Anything Model (SAM) has established a new baseline for static image segmentation; however, it is structurally ill-equipped for Referring Audio-Visual Segmentation (Ref-AVS). Current foundation models like SAM suffer from two critical limitations in this context:

\begin{enumerate}
    \item \textbf{Lack of Temporal Awareness:} SAM processes inputs as isolated static frames, failing to capture the temporal consistency and dynamic context necessary for video segmentation.
    \item \textbf{Reliance on Explicit Interaction:} SAM depends on manual user prompts (points, boxes, or masks) to identify targets. It lacks the native ability to interpret implicit ``multimodal prompts''---such as identifying an object described by a specific sound or textual description---without human intervention.
\end{enumerate}

\vspace{1em}
\noindent\textbf{Core Hypothesis}

We hypothesize that we can adapt the frozen, pre-trained knowledge of SAM for dynamic audio-visual scenes without heavy retraining by introducing two novel architectural components. First, a parallel \textbf{temporal modeling branch} can inject time-series context into the static encoder. Second, we can replace the manual ``user-in-the-loop'' prompting mechanism with \textbf{data-driven multimodal prompts}. By translating aligned audio and text features into the sparse and dense prompt formats that SAM natively understands, we believe the model can learn to ``prompt itself'' to segment temporally evolving objects described by complex semantic cues.

\vspace{1em}
\noindent\textbf{Proposed Methodology (High-Level Technical Approach)}

We propose \textbf{TSAM}, a modified architecture that wraps the standard SAM backbone. Our approach focuses on minimal trainable additions while keeping the heavy image encoder frozen.

\vspace{0.5em}
\noindent\textbf{1. Temporal Modeling Branch (Context Injection)}

Instead of retraining the heavyweight image encoder, we will introduce a lightweight auxiliary branch running in parallel.
\begin{itemize}
    \item \textbf{Sequential Processing:} As the frozen encoder processes a frame, this parallel branch will accept features from current and previous steps.
    \item \textbf{Cached Memory \& Adapters:} We will utilize a memory mechanism to store a running summary of previous frame representations. An adapter module will fuse these historical visual states with current audio-text embeddings. This allows the model to maintain object permanence and consistency across time.
\end{itemize}

\vspace{0.5em}
\noindent\textbf{2. Automated Multimodal Prompting}

We aim to synthesize the ``prompts'' SAM expects (points and masks) using audio-visual-text correlations rather than manual clicks.
\begin{itemize}
    \item \textbf{Sparse Prompting Module (Simulating Points):} We will design a query selection mechanism. By analyzing the correlation between the reference text and the audio stream, the system will select the most relevant ``audio cues.'' These cues will be projected and treated as sparse queries (analogous to point prompts) to guide the mask decoder toward the sound source.
    \item \textbf{Dense Prompting Module (Simulating Masks):} We will implement a deep fusion module that uses cross-attention to mix visual features with aligned audio and text embeddings. This will generate a dense, spatial feature map that acts as a coarse ``mask prompt,'' providing global context to the decoder.
\end{itemize}

\vspace{0.5em}
\noindent\textbf{3. Decoding}

The final segmentation will be generated by the standard SAM mask decoder, which will be queried by our synthetic sparse and dense prompts, refined by the temporally aware features from our auxiliary branch.

\vspace{1em}
\noindent\textbf{Expected Contribution}

\begin{itemize}
    \item \textbf{Architectural Novelty:} A framework for extending static, unimodal foundation models (like SAM) into the temporal and multimodal domains without compromising their zero-shot capabilities.
    \item \textbf{Methodological Advancement:} A demonstration of ``latent prompting,'' where semantic cues (audio/text) are mathematically translated into the geometric prompts (points/masks) required by segmentation models.
    \item \textbf{Theoretical Impact:} Establishing that lightweight temporal adapters and cross-modal attention mechanisms are sufficient to unlock video understanding in static image backbones.
\end{itemize}
\end{rawmaterialbox}

\begin{rawmaterialbox}[Dense Idea ($\mathcal{I}_{dense}$)]
\noindent\textbf{Problem Statement}

Referring Audio-Visual Segmentation (Ref-AVS) presents a complex challenge: segmenting objects in dynamic video scenes based on multimodal cues (audio, visual, and textual). Existing approaches, such as Video Object Segmentation (VOS) or Audio-Visual Segmentation (AVS), are limited by their reliance on single-modality cues or labor-intensive frame-level annotations. While Large Visual Foundation Models, specifically the Segment Anything Model (SAM), have revolutionized static image segmentation, they are fundamentally ill-suited for Ref-AVS in their current form. SAM lacks temporal awareness necessary for processing video frames and relies heavily on explicit user-interactive prompts (points, boxes) rather than interpreting the nuanced semantic interactions between audio, text, and visual data. There is currently no architecture that effectively adapts SAM's zero-shot capabilities to the temporal and multimodal requirements of Ref-AVS without requiring explicit user guidance.

\vspace{1em}
\noindent\textbf{Core Hypothesis}

We propose \textbf{TSAM (Temporal SAM)}, a novel architecture designed to extend SAM for dynamic audio-visual scenes. We hypothesize that by freezing the core SAM image encoder and augmenting it with a parallel \textbf{Temporal Modeling Branch}, we can inject spatio-temporal awareness into the model while preserving pre-trained knowledge. Furthermore, we propose replacing SAM's manual prompting mechanism with \textbf{data-driven multimodal prompts}. By synthesizing sparse queries (from audio-text alignment) and dense embeddings (from cross-modal attention), we can automate the prompting process, effectively guiding the mask decoder to segment targets defined by complex multimodal expressions.

\vspace{1em}
\noindent\textbf{Proposed Methodology (Detailed Technical Approach)}

\vspace{0.5em}
\noindent\textbf{1. Input Formulation and Feature Extraction}

We will process a sequence of inputs consisting of video frames, aligned audio, and reference text.
\begin{itemize}
    \item \textbf{Visual Input:} $n$ frames sampled at 1-second intervals, where each frame has a resolution of $1024 \times 1024$.
    \item \textbf{Audio Input:} Encoded offline using a pre-trained VGGish model to produce representations $\boldsymbol{F}_{a} \in \mathbb{R}^{n \times 128}$.
    \item \textbf{Text Input:} Extracted using a pre-trained RoBERTa model to generate representation $F_{t} \in \mathbb{R}^{l \times 768}$, where $l$ is the sequence length.
\end{itemize}

To align these modalities with SAM, we will project both audio and text representations into the SAM visual embedding space with dimensionality $d_{emb}$. This yields audio cues $F_{a} \in \mathbb{R}^{n \times 1 \times d_{emb}}$ and text cues $F_{t} \in \mathbb{R}^{n \times l \times d_{emb}}$ (temporally expanded along $n$).

\vspace{0.5em}
\noindent\textbf{2. Temporal Modality Fusion and Cached Memory}

We will implement a \textbf{Temporal Modality Fusion Layer (TMFL)} to synthesize expression-related multimodal cues. We define the fused cues as:
\[ F_{a}^{\prime}, F_{t}^{\prime} = \mathrm{SA}(\mathrm{Concat}(F_{a}, F_{t})) \]
where $\operatorname{SA}(\cdot)$ applies temporal self-attention.

We will also utilize a \textbf{Cached Memory (CM)} mechanism. This will store $\hat{F}_{a}$, an accumulated summary of $F_{a}^{\prime}$ over time, capturing the mean modality cues and evolving audio context. Text cues are enriched via summation: $\hat{F}_{t} = F_{t} + F_{t}^{\prime}$.

\vspace{0.5em}
\noindent\textbf{3. Temporal Modeling Branch}

To address SAM's lack of temporal dynamics, we will introduce a trainable temporal modeling branch parallel to the final $M$ blocks of the frozen SAM image encoder.
\begin{itemize}
    \item \textbf{Structure:} The branch consists of blocks initialized with SAM's pre-trained weights.
    \item \textbf{Data Flow:} Each $m$-th temporal block ($m = 1, \dots, M$) integrates the output of the previous temporal block $y_{m-1}$, the output of the corresponding frozen SAM block $y_{(N-M)+(m-1)}^{\mathrm{SAM}}$, and the enriched audio cues $\hat{F}_{a}$ via an Adapter Module (AM).
    \item \textbf{Formulation:}
    \[ x_{m} = y_{m-1} + \mathbf{CM}_{m}(y_{(N-M)+(m-1)}^{\mathrm{SAM}}) + \mathbf{AM}_{m}(\hat{F}_{a}) \]
    Here, $\mathbf{CM}_{m}$ is a cached memory applied to the SAM block output, and $\mathbf{AM}_{m}$ is a bottleneck adapter block facilitating deep interaction between audio, visual, and text modalities.
\end{itemize}

The final video cues $F_{v}$ are derived by summing the output of the temporal branch ($Y_{M}$) and the final SAM encoder output ($Y_{N}^{SAM}$):
\[ F_{v} = Y_{M} + Y_{N}^{SAM} \]

\vspace{0.5em}
\noindent\textbf{4. Multimodal Prompting Modules}

We will replace SAM's manual prompts with two specific modules designed to query the mask decoder:

\vspace{0.25em}
\noindent\textbf{A. Sparse Prompting Module (SPM)}

This module acts as a global context guide. We will employ a language-guided query selection mechanism to identify the $k$ most relevant audio cues associated with the text expression. The indices of these cues, $A_{k}$, are selected via:
\[ A_{k} = \mathrm{Top}_{k} \left( \operatorname*{max} \mathrm{f} (\hat{F}_{a} \hat{F}_{t}^{\top}) \right) \]
where $\mathrm{Top}_{k}$ selects the top $k$ indices and $\operatorname{maxf}$ computes the maximum along the cue dimension. These selected cues are fused with visual cues via cross-attention and mapped to sparse prompt embeddings.

\vspace{0.25em}
\noindent\textbf{B. Dense Prompting Module (DPM)}

This module operates at a fine granularity. It generates spatially comprehensive embeddings through a sequence of cross-attention mechanisms:
\begin{enumerate}
    \item \textbf{Audio Cross-Attention:} Aligns audio cues with visual cues.
    \item \textbf{Text Cross-Attention:} Refines the result with text cues to focus on the specific target object.
    \item \textbf{Refinement:} A feed-forward network processes the output to create dense embeddings for the SAM mask decoder.
\end{enumerate}

\vspace{0.5em}
\noindent\textbf{5. Training Objective}

The model will be trained end-to-end using a weighted sum of Binary Cross-Entropy ($\mathcal{L}_{\mathrm{BCE}}$) and Intersection over Union ($\mathcal{L}_{\mathrm{IoU}}$) losses to compare predicted masks against ground truth:
\[ \mathcal{L}_{\mathrm{total}} = \mathcal{L}_{\mathrm{BCE}} + \lambda \cdot \mathcal{L}_{\mathrm{IoU}} \]
We will set $\lambda = 1.0$ to balance the contributions of both loss components.

\vspace{1em}
\noindent\textbf{Expected Contribution}

\begin{enumerate}
    \item \textbf{Architecture:} A novel end-to-end framework (TSAM) that repurposes the Segment Anything Model for Referring Audio-Visual Segmentation without requiring extensive retraining of the image backbone.
    \item \textbf{Temporal Adaptation:} The introduction of a lightweight Temporal Modeling Branch that enables SAM to capture intricate spatio-temporal interactions across video frames, overcoming its static-image limitation.
    \item \textbf{Automated Multimodal Prompting:} A theoretical framework for converting audio-visual-text correlations into the sparse and dense prompts required by SAM, effectively replacing human interaction with data-driven multimodal guidance.
\end{enumerate}
\end{rawmaterialbox}

\begin{rawmaterialbox}[Experimental Log ($\mathcal{E}$)]
\noindent\textbf{1. Experimental Setup}

We conducted a comprehensive evaluation of the proposed TSAM method for the Referring Audio-Visual Segmentation (Ref-AVS) task.

\begin{itemize}
    \item \textbf{Datasets:}
    \begin{itemize}
        \item \textbf{Ref-AVS Dataset:} We utilized the Ref-AVS dataset containing 20,000 text expressions and pixel-level annotations across 4,000 10-second videos.
        \item \textbf{Object Categories:} The dataset included audible objects (20 musical instruments, 8 animals, 15 machines, 5 humans) and 3 categories of static, inaudible objects.
        \item \textbf{Splits:}
        \begin{itemize}
            \item \textit{Training Set:} 2,908 videos.
            \item \textit{Validation Set:} 276 videos.
            \item \textit{Test Set:} 818 videos total, subdivided into:
            \begin{itemize}
                \item \textit{Seen:} 292 videos (categories present in training).
                \item \textit{Unseen:} 269 videos (categories not seen during training; tests generalization).
                \item \textit{Null:} 257 videos (text refers to non-existent/not visible objects; empty true masks).
            \end{itemize}
        \end{itemize}
    \end{itemize}
    
    \item \textbf{Evaluation Metrics:}
    \begin{itemize}
        \item \textbf{Standard Metrics:} We employed the Jaccard Index ($\mathcal{J}$) and F-score ($\mathcal{F}$) for the Seen and Unseen subsets.
        \item \textbf{Null Subset Metric:} We employed the metric $\mathcal{S}$, which measures the ratio of the predicted mask area to the background area ($\mathcal{S} = \sqrt{\text{predicted mask area} / \text{background area}}$). A lower $\mathcal{S}$ value indicates better performance (less incorrect segmentation).
    \end{itemize}
    
    \item \textbf{Baselines Compared:}
    \begin{itemize}
        \item \textbf{AVS Methods (Augmented with Text):} AVSBench (PVT-v2 backbone), AVSegFormer (PVT-v2 backbone), GAVS (SAM backbone), SAMA (SAM backbone). \textit{Note: We re-evaluated SAMA by integrating text with audio/visual inputs.}
        \item \textbf{Ref-VOS Methods (Augmented with Audio):} ReferFormer (Video-Swin backbone), R2-VOS (Video-Swin backbone).
        \item \textbf{Ref-AVS Method:} EEMC (Mask2Former backbone; previous state-of-the-art).
    \end{itemize}
    
    \item \textbf{Implementation Details:}
    \begin{itemize}
        \item \textbf{Model Initialization:} We initialized TSAM using the pre-trained ViT-B variant of SAM ($N=12$, embedding dimension $d_{emb}=256$).
        \item \textbf{Architecture Settings:} The temporal branch consisted of $M=4$ blocks. The number of selected audio queries was set to $k=5$.
        \item \textbf{Hardware:} Training was performed on eight AMD GPUs in a distributed setup.
        \item \textbf{Optimizer:} AdamW optimizer was used.
        \item \textbf{Hyperparameters:} Initial learning rate of $1 \cdot 10^{-4}$, batch size of 1.
        \item \textbf{Training Duration:} The model was trained for 15 epochs with periodic evaluations.
        \item \textbf{Model Selection:} We selected the best-performing model on the validation subset for testing.
    \end{itemize}
\end{itemize}

\vspace{1em}
\noindent\textbf{2. Raw Numeric Data}

\vspace{0.5em}
\noindent\textbf{Table 1: Performance comparison on the Ref-AVS dataset}\\
\textit{Note: Baselines marked with \dag{} had text integration added; baselines marked with \ddag{} had audio integration added; * marks our re-implementation of SAMA with text added.}

\begin{center}
\resizebox{\linewidth}{!}{
\begin{tabular}{lllccccc}
\toprule
\textbf{Method} & \textbf{Task} & \textbf{Visual Backbone} & \textbf{Seen J(\%)} & \textbf{Seen F} & \textbf{Unseen J(\%)} & \textbf{Unseen F} & \textbf{Null S($\downarrow$)} \\
\midrule
AVSBench & AVS\dag & PVT-v2 & 23.20 & 0.511 & 32.36 & 0.547 & 0.208 \\
AVSegFormer & AVS\dag & PVT-v2 & 33.47 & 0.470 & 36.05 & 0.501 & 0.171 \\
GAVS & AVS\dag & SAM & 28.93 & 0.498 & 29.82 & 0.497 & 0.190 \\
SAMA & AVS* & SAM & 39.22 & 0.562 & 47.50 & 0.566 & 0.130 \\
ReferFormer & Ref-VOS\ddag & V-Swin & 31.31 & 0.501 & 30.40 & 0.488 & 0.176 \\
R2-VOS & Ref-VOS\ddag & V-Swin & 25.01 & 0.410 & 27.93 & 0.498 & 0.183 \\
EEMC & Ref-AVS & M2F & 34.20 & 0.513 & 49.54 & 0.648 & \textbf{0.007} \\
\textbf{TSAM (Ours)} & Ref-AVS & SAM & \textbf{43.43} & \textbf{0.568} & \textbf{54.58} & \textbf{0.664} & 0.017 \\
\bottomrule
\end{tabular}
}
\end{center}

\vspace{1em}
\noindent\textbf{Table 2: Ablation study on TSAM components and IoU loss}\\
\textit{Legend: TB=Temporal Branch, TMFL=Temporal Modality Fusion Layer, DPM=Dense Prompting Module, SPM=Sparse Prompting Module, CM=Cached Memory, AM=Adapter Module. Mix(S+U) is the average of Seen and Unseen.}

\begin{center}
\resizebox{\linewidth}{!}{
\begin{tabular}{lccccccc}
\toprule
\textbf{Setting} & \textbf{Seen J(\%)} & \textbf{Seen F} & \textbf{Unseen J(\%)} & \textbf{Unseen F} & \textbf{Mix(S+U) J(\%)} & \textbf{Mix(S+U) F} & \textbf{Null S($\downarrow$)} \\
\midrule
(1) TSAM (Full) & \textbf{43.43} & 0.568 & \textbf{54.58} & \textbf{0.664} & \textbf{49.01} & \textbf{0.616} & 0.017 \\
(2) - TB & 33.05 & 0.507 & 50.48 & 0.657 & 41.77 & 0.582 & 0.505 \\
(3) - TMFL & 40.35 & 0.579 & 45.54 & 0.627 & 42.95 & 0.603 & 0.018 \\
(4) - DPM & 42.72 & 0.580 & 49.10 & 0.647 & 45.91 & 0.614 & 0.018 \\
(5) - SPM & 43.04 & 0.580 & 49.75 & 0.652 & 46.40 & \textbf{0.616} & 0.018 \\
(6) - SPM+DPM & 42.60 & 0.602 & 40.58 & 0.604 & 41.59 & 0.603 & 0.018 \\
(7) - CM(a+v) & 42.07 & 0.544 & 49.11 & 0.659 & 45.59 & 0.602 & 0.018 \\
(8) - CM(v) & 42.75 & 0.549 & 51.18 & 0.660 & 46.97 & 0.605 & 0.018 \\
(9) - AM & 43.13 & \textbf{0.600} & 40.79 & 0.617 & 41.96 & 0.609 & 0.017 \\
(10) - $\mathcal{L}_{\mathrm{IoU}}$ & 38.29 & 0.564 & 42.15 & 0.631 & 40.22 & 0.598 & \textbf{0.008} \\
\bottomrule
\end{tabular}
}
\end{center}

\vspace{1em}
\noindent\textbf{Table 3: Effect of audio queries ($k$) and temporal branch depth ($M$)}

\begin{center}
\resizebox{\linewidth}{!}{
\begin{tabular}{lccccccc}
\toprule
\textbf{Variation} & \textbf{Seen J(\%)} & \textbf{Seen F} & \textbf{Unseen J(\%)} & \textbf{Unseen F} & \textbf{Mix(S+U) J(\%)} & \textbf{Mix(S+U) F} & \textbf{Null S($\downarrow$)} \\
\midrule
\textbf{$k=3$} & 43.58 & \textbf{0.579} & 50.43 & 0.655 & 47.01 & \textbf{0.617} & 0.018 \\
\textbf{$k=5$} & 43.43 & 0.568 & \textbf{54.58} & \textbf{0.664} & \textbf{49.01} & 0.616 & \textbf{0.017} \\
\textbf{$k=7$} & 43.24 & 0.573 & 46.95 & 0.630 & 45.10 & 0.601 & 0.018 \\
\textbf{$M=2$} & 42.04 & 0.566 & 53.51 & 0.651 & 47.76 & 0.609 & 0.023 \\
\textbf{$M=4$} & \textbf{43.43} & 0.568 & \textbf{54.58} & \textbf{0.664} & \textbf{49.01} & \textbf{0.616} & \textbf{0.017} \\
\textbf{$M=6$} & 43.27 & \textbf{0.575} & 49.86 & 0.640 & 46.57 & 0.608 & 0.020 \\
\bottomrule
\end{tabular}
}
\end{center}

\vspace{1em}
\noindent\textbf{3. Qualitative Observations}

\vspace{0.5em}
\noindent\textbf{Comparisons with State-of-the-Art:}
\begin{itemize}
    \item \textbf{Backbone Analysis:} We observed that methods utilizing prior segmentation visual backbones (SAM and Mask2Former) generally outperformed those based on PVT-v2 and V-Swin backbones.
    \item \textbf{SAM-Based Baseline Limitations:} Although GAVS and SAMA are SAM-based, they performed worse than EEMC. We noted that SAMA failed to leverage SAM's flexible, promptable nature, and GAVS lacked cohesive multimodal fusion.
    \item \textbf{TSAM Performance:} TSAM achieved the highest performance on both Seen and Unseen test sets. Specifically, TSAM improved over EEMC by 9.23\% in Jaccard Index on the Seen set and 5.04\% on the Unseen set.
    \item \textbf{Null Set Performance:} TSAM fell slightly behind EEMC on the Null test set (S value 0.017 vs 0.007). We attribute this to SAM's inherent limitation of always attempting to produce a segmentation mask even when no target object is present.
\end{itemize}

\vspace{0.5em}
\noindent\textbf{Ablation Observations:}
\begin{itemize}
    \item \textbf{Temporal Branch (TB):} Removing the temporal branch caused the most significant performance degradation (Seen Jaccard dropped from 43.43\% to 33.05\%). This confirmed the critical role of temporal modeling for generalizing across video frames.
    \item \textbf{Prompting Modules (SPM/DPM):} We found that the Sparse Prompting Module (SPM) and Dense Prompting Module (DPM) play complementary roles. Removing both simultaneously resulted in a notable decrease in segmentation performance.
    \item \textbf{Cached Memory (CM):} Omitting cached memory, especially the visual-only memory, degraded performance, highlighting the importance of shared memory for temporal alignment.
    \item \textbf{Adapter Module (AM):} The omission of the adapter module caused a significant performance drop, particularly on the Unseen test set, validating its role in facilitating deep multimodal interaction.
    \item \textbf{Loss Function:} Including the IoU loss ($\mathcal{L}_{\mathrm{IoU}}$) improved performance on Seen and Unseen sets but slightly degraded the Null score.
\end{itemize}

\vspace{0.5em}
\noindent\textbf{Hyperparameter Sensitivity:}
\begin{itemize}
    \item \textbf{Audio Queries ($k$):} We found that $k=5$ yielded optimal results. A lower value ($k=3$) performed well on Seen data but worse on Unseen, while a higher value ($k=7$) likely introduced irrelevant queries, overloading the decoder.
    \item \textbf{Temporal Depth ($M$):} $M=4$ blocks provided the best balance. Shallower setups ($M=2$) lacked sufficient temporal depth, while deeper setups ($M=6$) appeared to introduce excessive complexity that impaired generalization.
\end{itemize}

\vspace{0.5em}
\noindent\textbf{Visual Qualitative Analysis:}
\begin{itemize}
    \item \textbf{Seen Test Set:} We observed that TSAM consistently produced high-quality masks for targeted objects. It successfully segmented inaudible objects when guided by textual cues (e.g., ``The object behind the sounding women''), demonstrating effective processing of complex multimodal instructions.
    \item \textbf{Unseen Test Set:} TSAM demonstrated a remarkable capacity to generalize to novel objects. It effectively aligned audio-visual and textual inputs in novel scenes (e.g., segmenting a ``truck moving'' or ``tuba being played'' that were not in training categories).
    \item \textbf{Generalization:} The visual results underscored TSAM's ability to preserve SAM's pre-trained knowledge while using the temporal branch to understand dynamic scenes.
\end{itemize}
\end{rawmaterialbox}
\section{Experiment Details}
\label{appendix:exp_details}

\subsection{Models and Configuration}
\label{appendix:model_details}

\paragraph{API Access} We access all models from the Gemini-3 family (\texttt{gemini-3.1-pro-preview}, \texttt{gemini-\\3-flash-preview}, and \texttt{gemini-3-pro-image-preview}) via the Google Cloud Vertex AI platform. The GPT model (\texttt{gpt-5-2025-08-07}) is accessed via the official OpenAI API. Paper searches for verification and metadata fetching are conducted using the Semantic Scholar API (\url{https://api.semanticscholar.org/graph/v1/paper/search}).

\paragraph{Model Usage} The primary manuscript writing backbone for \ourmethod{} and all baselines is strictly fixed to Gemini-3.1-Pro. For literature discovery, \ourmethod{} utilizes Gemini-3-Flash with Google Search grounding. To ensure a fair comparison, we also use Gemini-3-Flash for the citation gathering process in AI Scientist-v2.

\paragraph{Evaluation Settings} For automated technical quality evaluations (AgentReview~\citep{jin2024agentreview}, AI Scientist-v2 Reviewer~\citep{yamada2025aiscientistv2}, and ScholarPeer~\citep{goyal2026scholarpeer}), we apply a universal temperature of 0.75 (Gemini-3.1-Pro) to balance instruction adherence with creative synthesis. For automated side-by-side evaluations, Gemini-3.1-Pro is set to a temperature of 0.0 to guarantee reproducibility. Since GPT-5 does not currently support temperature adjustment, we operate it at its fixed default temperature of 1.0 across all evaluations. During ScholarPeer~\citep{goyal2026scholarpeer} evaluation, we disable the baseline scouting agent for all pipelines to isolate presentation quality from fixed experimental constraints.

\paragraph{Research Cutoff} We align the research cutoff dates with the official submission deadlines of each venue: November 2024 for CVPR 2025 papers and October 2024 for ICLR 2025 papers.

\subsection{Baseline Selection}
\label{appendix:baseline_selection}

We explain why several existing systems are not included as baselines:

\begin{itemize}
    \item \textbf{OmniScientist}~\citep{shao2025omniscientist}: Lacks a publicly available, reproducible codebase.
    \item \textbf{PaperRobot}~\citep{wang2019paperrobot} and \textbf{Data-to-Paper}~\citep{technion2024datatopaper}: Do not support end-to-end generation of submission-ready \mylatex\ manuscripts.
    \item \textbf{AI-Researcher}~\citep{tang2025ai}: Operates as a full research pipeline requiring highly structured intermediate states (e.g., pre-written sections, curated literature summaries) rather than unconstrained raw materials. Adapting it to our setting would require reconstructing these abstractions, bypassing the core synthesis challenge our benchmark evaluates.
    \item \textbf{CycleResearcher}~\citep{weng2024cycleresearcher}: Requires a structured BibTeX reference list as input—an artifact rarely available during early drafting. The model also fails on unstructured inputs outside its specific training format, preventing it from directly processing raw materials.
    \item \textbf{Generic RAG pipelines} (\citep{wu2025autosurvey2}, \citep{go2025lira}): Designed primarily for literature surveys, these lack the specialized mechanisms required to autonomously draft, structure, and format full submission-ready research papers.
\end{itemize}

\subsection{Citation Verification}
\label{appendix:citation_verification}

To guarantee the factual accuracy and relevance of the generated literature review, our system employs a two-phase retrieval and verification pipeline for citation gathering. First, we use Gemini-3-Flash with Google Search grounding to rapidly discover candidate papers based on the generated outline. Following this discovery phase, each candidate paper undergoes strict sequential verification via the Semantic Scholar API to extract abstracts and metadata.

During verification, each candidate must resolve to a valid Semantic Scholar entity via a fuzzy title match (Levenshtein distance ratio $> 70$~\citep{Levenshtein1965BinaryCC}), augmented by a point-bonus for exact year alignment. To enter the final context pool, the entity must possess a retrievable abstract and strictly predate the research cutoff (when specified down to the month, the system defaults to the first day of that month as the strict boundary). Finally, gathered citations are deduplicated using unique paper ID keys.

Once the papers are verified and enriched with metadata (including authors, venue, year, citation count, and abstract), the compiled context is passed to the primary writing agent. To enforce high citation density and eliminate hallucinated references, the system prompt strictly constrains the model to cite only the provided verified papers, explicitly mandating that at least 90\% of the gathered literature pool must be actively integrated and cited when synthesizing the \textit{Introduction} and \textit{Related Work} sections.



\subsection{Information Leakage Prevention}
\label{appendix:info_leakage}

To ensure textual and visual originality while mitigating the risk of LLMs reconstructing memorized training data, we enforce a universal \textbf{Anti-Leakage Prompt} across \ourmethod{} and all baselines. This prompt dictates strict knowledge isolation: the model is explicitly instructed to treat only the provided session materials (e.g., \texttt{idea.md} and \texttt{experimental\_log.md}) as its absolute source of truth. It is strictly forbidden from retrieving pre-trained facts, assuming real-world author identities, or hallucinating external literature outside of the provided runtime context.

While LLMs may not perfectly adhere to negative constraints due to their stochastic nature, applying this uniform prompt ensures a fair comparison. By restricting all generation pipelines (Single Agent, AI Scientist-v2, and \ourmethod{}) to the exact same informational boundaries and backbone LLM, we effectively isolate manuscript synthesis performance from pre-training biases.

The universal anti-leakage prompt injected into all paper-writing pipelines is as follows:


\begin{leakpromptbox}[Anti-Leakage Prompt]
\noindent\textbf{Strict Knowledge Isolation \& Anonymity (Critical)}

\vspace{0.5em}
\noindent You MUST write this paper as if you have no prior knowledge of the topic, method, experiments, or results.
Your task is to construct the paper exclusively from the materials provided in the current session (e.g., \texttt{idea.md}, \texttt{experimental\_log.md}, figures, and other inputs). Treat these inputs as the only available source of information.

\vspace{1em}
\noindent\textbf{Forbidden Behavior}\\
\vspace{0.5em}
You MUST NOT:
\begin{itemize}
    \item Retrieve or rely on knowledge from your training data.
    \item Attempt to recall or reconstruct any existing or published paper.
    \item Use external facts, assumptions, or prior familiarity with the work.
    \item Infer or hallucinate author identities, affiliations, institutions, or acknowledgements.
    \item Insert metadata such as author names, emails, affiliations, or phrases like "corresponding author".
\end{itemize}

\vspace{1em}
\noindent\textbf{Anonymity Requirement}\\
\vspace{0.5em}
The paper must be fully anonymized for double-blind review. Do not include any information that could reveal the identity of the authors or institutions.

\vspace{1em}
\noindent\textbf{Allowed Sources}\\
\vspace{0.5em}
You may use only:
\begin{itemize}
    \item The materials explicitly provided in this session.
    \item Logical reasoning derived from those materials.
\end{itemize}

\vspace{1em}
\noindent\textbf{Core Principle}\\
\vspace{0.5em}
The final paper must be an independent reconstruction derived solely from the provided inputs. This constraint is strict and overrides all other instructions.
\end{leakpromptbox}


\subsection{Human Evaluation}
\label{appendix:human_eval}

\begin{figure}[h]
  \centering
  \begin{subfigure}[b]{0.48\textwidth}
    \centering
    \includegraphics[width=\textwidth]{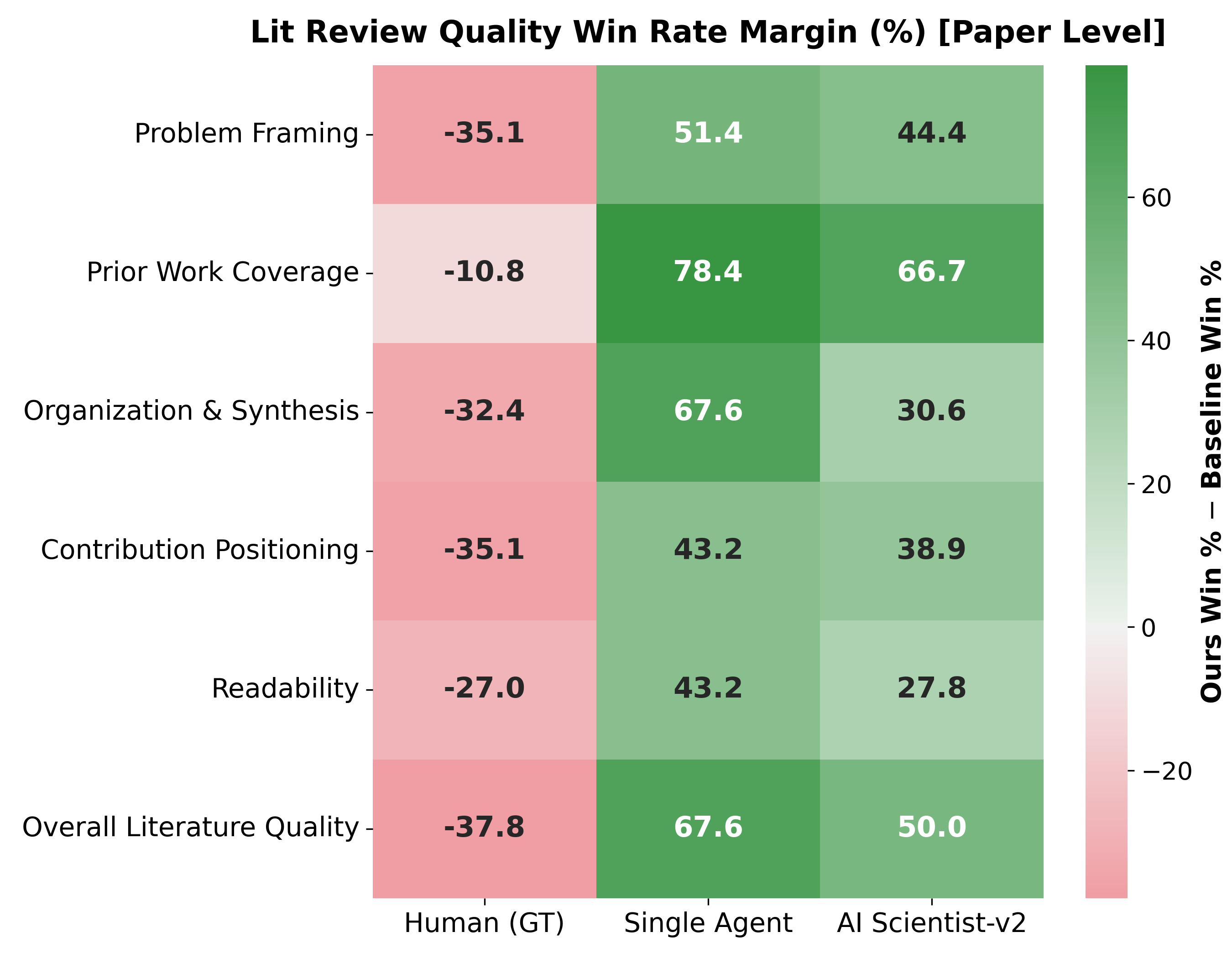}
    \caption{SxS Results: Literature Review Quality}
    \label{fig:human_eval_heatmap_lit}
  \end{subfigure}
  \hfill
  \begin{subfigure}[b]{0.48\textwidth}
    \centering
    \includegraphics[width=\textwidth]{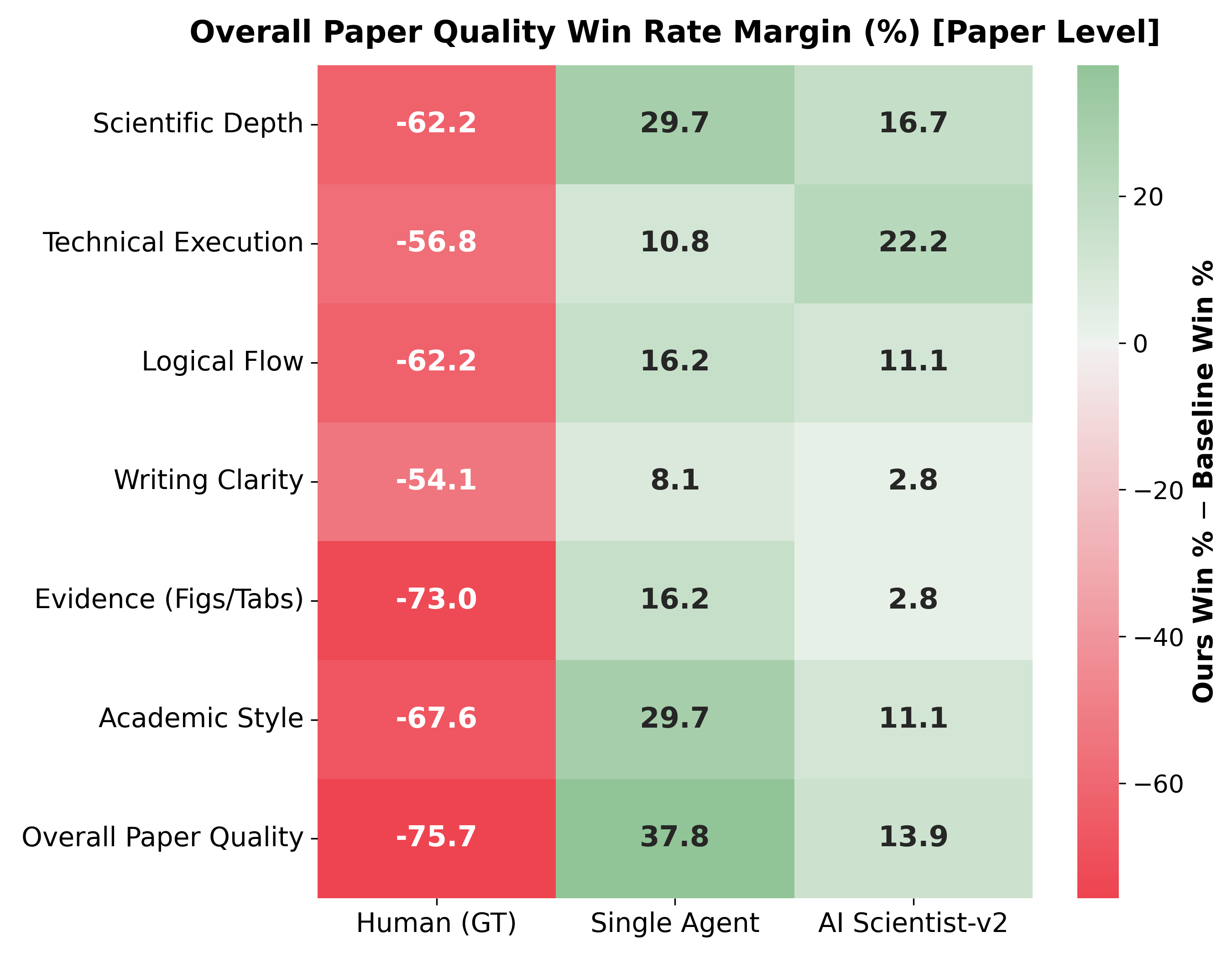}
    \caption{SxS Results: Overall Paper Quality}
    \label{fig:human_eval_heatmap_overall}
  \end{subfigure}
  \caption{\textbf{Human Side-by-Side (SxS) Evaluation Results.} Heatmaps display the win rate margin (\%) of \ourmethod{} against baselines across various evaluation dimensions. Positive values (green) favor our approach. \ourmethod{} consistently outperforms both AI baselines (Single Agent and AI Scientist-v2), though a quality gap remains compared to the human-written ground truth (GT).}
  \label{fig:human_eval_heatmaps_appendix}
\end{figure}

\begin{figure}[htbp]
  \centering
  \begin{subfigure}[b]{\textwidth}
    \centering
    \includegraphics[width=0.8\linewidth]{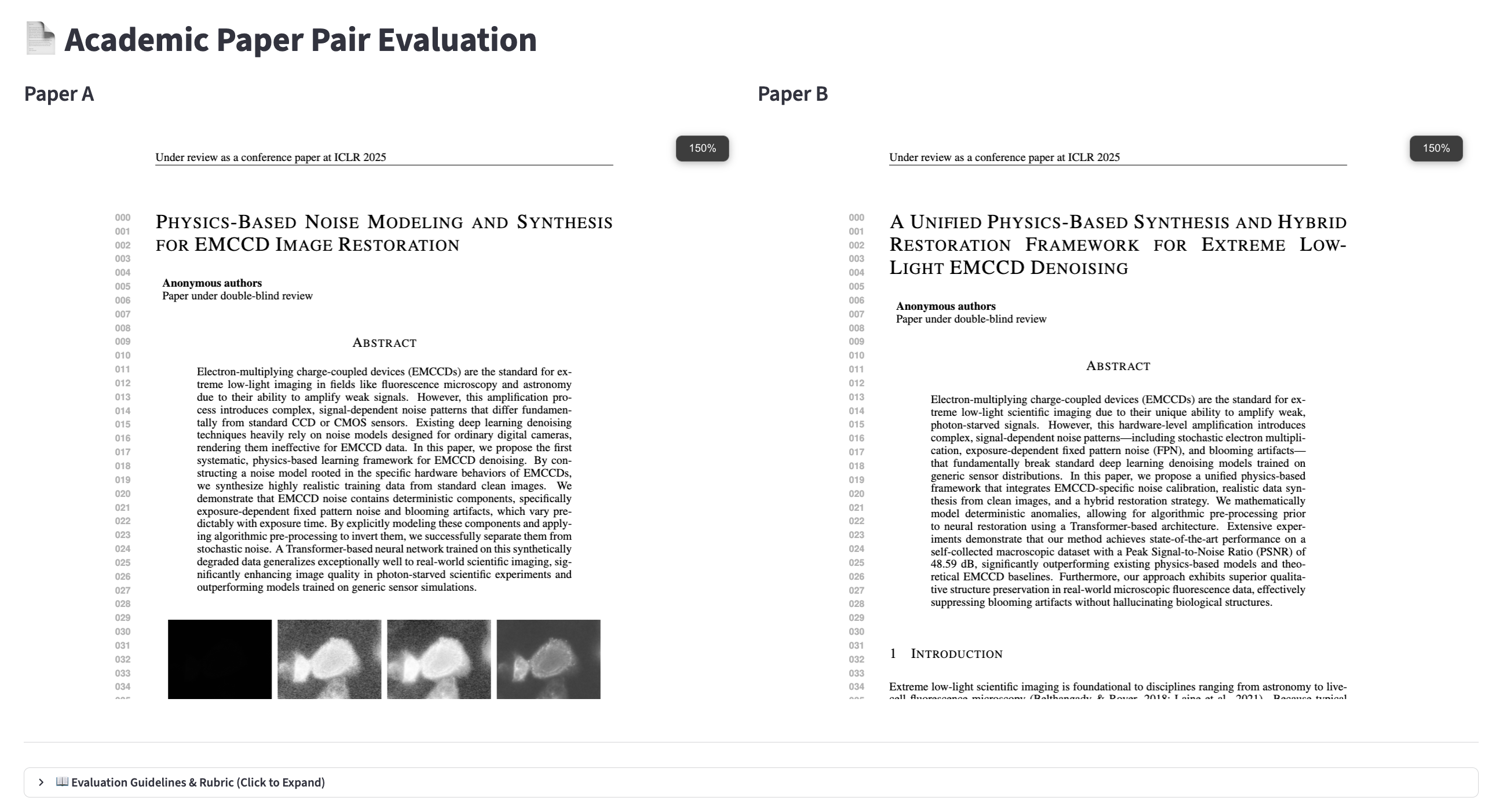}
    \caption{Side-by-Side Document View}
    \label{fig:human_eval_ui_top}
  \end{subfigure}
  
  \vspace{1em} 
  
  \begin{subfigure}[b]{\textwidth}
    \centering
    \includegraphics[width=0.75\linewidth]{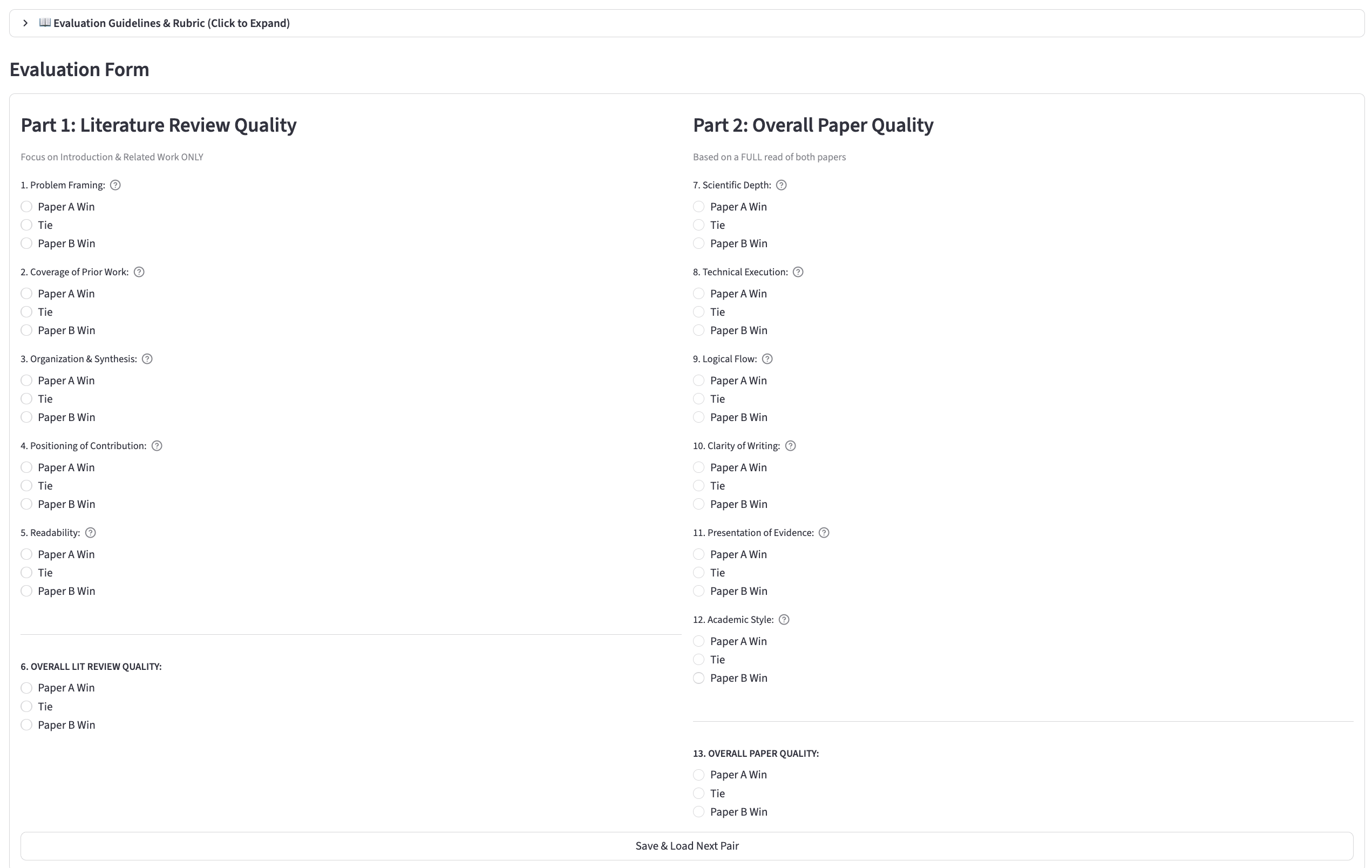}
    \caption{Evaluation Questionnaire}
    \label{fig:human_eval_ui_bottom}
  \end{subfigure}
  \caption{\textbf{Streamlit-based Human Evaluation Interface.} The custom UI displays two manuscripts side-by-side (a) and provides a fine-grained questionnaire alongside a final preference selection (b).}
  \label{fig:human_eval_interface}
\end{figure}

We developed a Streamlit interface (Figure~\ref{fig:human_eval_interface}) for blind SxS comparison of generated manuscripts. Annotators evaluated paper pairs on a 3-point scale (Win, Tie, Loss) across two aspects: (1) \textit{Literature Review Quality} (evaluating the \textit{Introduction} and \textit{Related Work}) based on problem framing, prior work coverage, synthesis, positioning, and readability; and (2) \textit{Overall Paper Quality} (evaluating the full manuscript) based on scientific depth, technical execution, logical flow, writing clarity, evidence presentation, and academic style. 

Eleven AI researchers completed 180 randomized, paired evaluations across 40 sampled papers from \ourdataset{} (120 unique paper pairs). For each evaluation, annotators answered 12 fine-grained diagnostic questions before making a final holistic judgment. As shown in Figure~\ref{fig:human_eval_heatmaps_appendix}, our approach, \ourmethod{}, systematically outperforms all AI baselines across every evaluation metric. While a quality gap remains when compared to the human-written papers (GT), \ourmethod{} achieves substantial win margins over all evaluated AI baselines in both literature review synthesis and overall paper quality.
\vspace{1em}

The detailed annotation guidelines and rubric provided to the human annotators are listed below:

\begin{humanevalbox}[Human Annotation Guidelines \& Rubric]
\textbf{General Rules}
\vspace{0.5em}
\begin{itemize}
    \item \textbf{Reading Scope:} 
    \begin{itemize}
        \item \textbf{Literature Review Quality} must be judged by reading the \textbf{Introduction and Related Work} sections \textit{only}.
        \item \textbf{Overall Paper Quality} must be judged based on a \textbf{full read} of both papers entirely.
    \end{itemize}
    \item \textbf{Ignore Templates and Metadata:} Some papers may include conference templates (e.g., CVF headers), watermarks, or text indicating prior publication. \textbf{Additionally, some papers may list author names while others are anonymized for review.} Please completely ignore these artifacts. Evaluations must be based purely on the intrinsic quality of the text and research, not the templates, perceived prestige, or author identities.
    \item Evaluate each paper independently before comparing them.
    \item \textbf{Do not} base your decision solely on paper length or verbosity.
\end{itemize}

\vspace{1em}
\noindent\textbf{Literature Review Quality (Focus on Introduction \& Related Work)}
\vspace{0.5em}
\begin{itemize}
    \item \textbf{Motivation:} Does it clearly explain the problem, why it matters, and the gap in existing work?
    \item \textbf{Coverage:} Is the overview of prior research relevant and complete?
    \item \textbf{Synthesis:} Does it organize and group related work logically, rather than just blindly listing papers?
    \item \textbf{Positioning:} Does it clearly explain how \textit{this} paper differs from existing methods?
    \item \textbf{Readability:} Is the text concise, clear, and easy to follow?
\end{itemize}

\vspace{1em}
\noindent\textbf{Overall Paper Quality (Holistic Review)}
\vspace{0.5em}
\begin{itemize}
    \item \textbf{Scientific Depth:} Are the theoretical foundations, justifications, and experimental setups rigorous?
    \item \textbf{Execution:} Is the methodology implemented innovatively and effectively?
    \item \textbf{Logical Flow:} Does the paper transition smoothly from Abstract to Conclusion?
    \item \textbf{Clarity:} Is the writing precise and free of repetitive fluff or ambiguity?
    \item \textbf{Evidence:} Are figures, tables, and results cleanly integrated and referenced in the text?
    \item \textbf{Style:} Does it maintain a polished, consistent, professional academic tone?
\end{itemize}
\end{humanevalbox}

\section{Paper Visualization}
\label{appendix:paper_viz}

\subsection{Side-by-Side (SxS) Paper Visualizations}
\label{appendix:paper_viz_sxs}
In this section, we present full manuscript generation results for two samples (one from CVPR~\citep{bahari2025certified} and one from ICLR~\citep{drakulic2025goalgeneralistcombinatorialoptimization}) from \ourdataset{}. For each sample, we display the manuscripts generated from the original papers' raw materials under the sparse idea setting. We compare the final outputs produced by Single Agent, AI Scientist-v2, \ourmethod{} (PlotOff), and \ourmethod{} (PlotOn).


\begin{figure}[htbp]
  \centering
  \includegraphics[width=\textwidth, height=0.72\textheight, keepaspectratio]{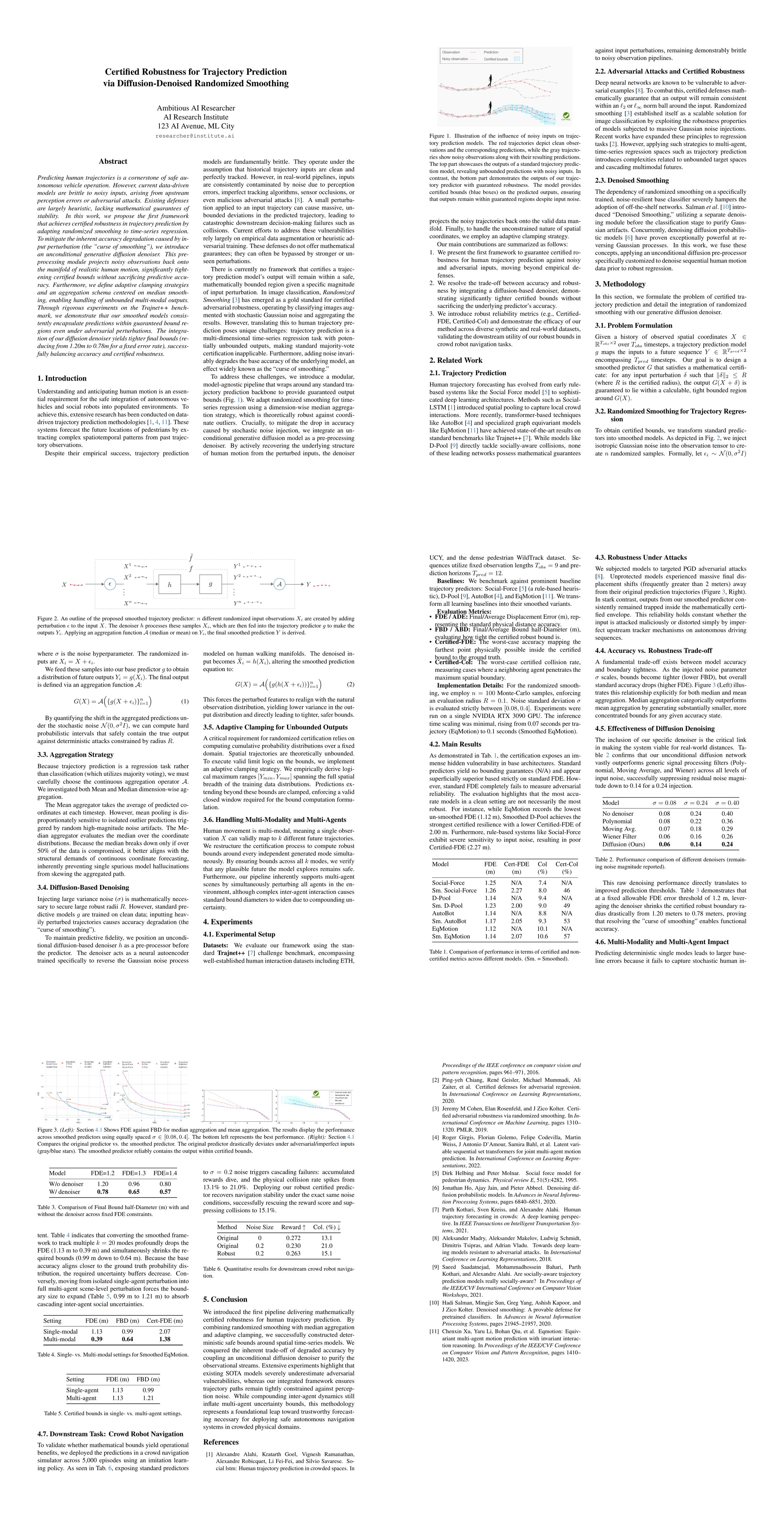}
  \caption{\textbf{CVPR Sample (Single Agent).} Manuscript generated by the Single Agent baseline from raw materials under the sparse idea setting.}
  \label{fig:sxs_cvpr_single_agent}
\end{figure}

\clearpage

\begin{figure}[p]
  \centering
  \includegraphics[width=\textwidth, height=0.85\textheight, keepaspectratio]{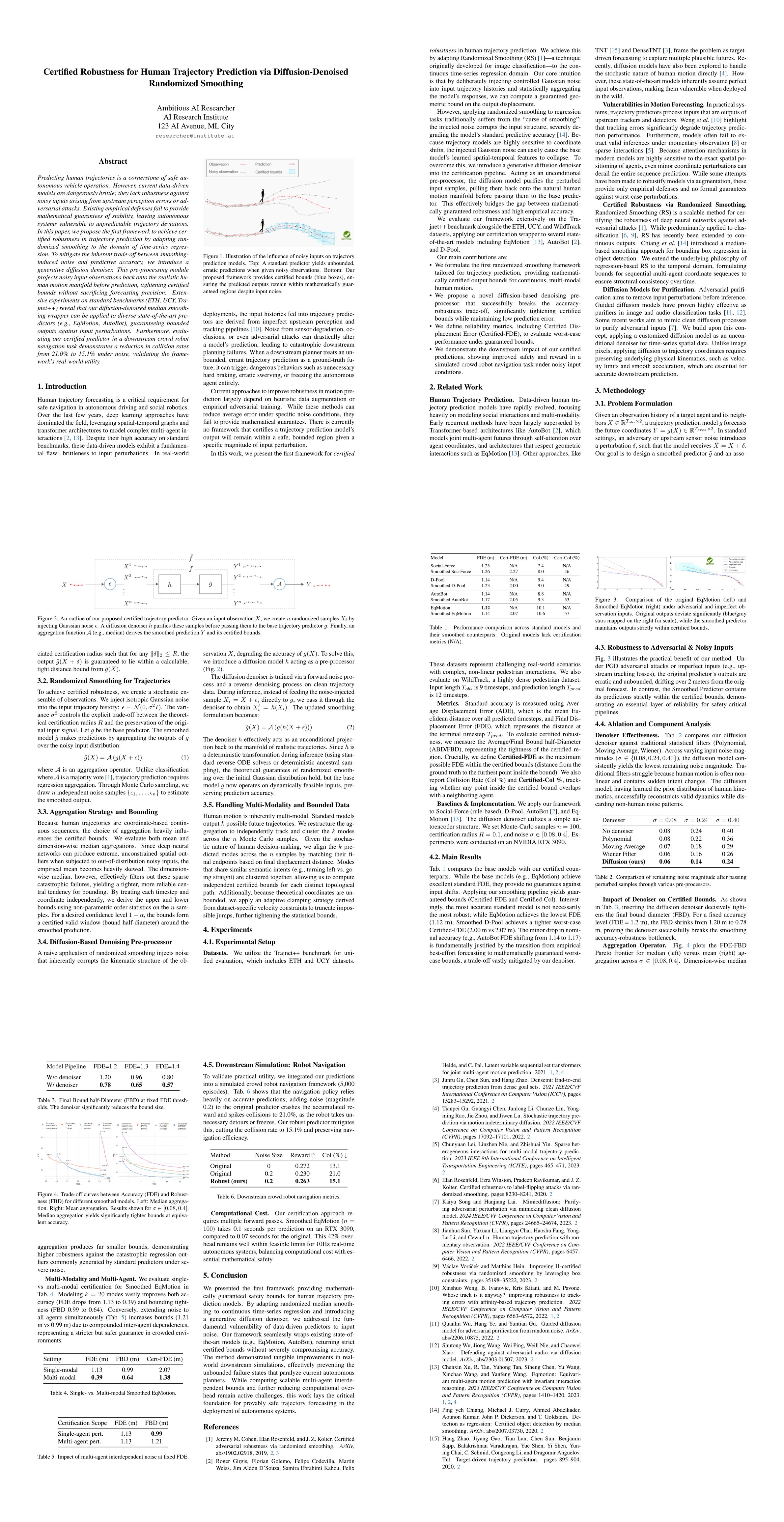}
  \caption{\textbf{CVPR Sample (AI Scientist-v2).} Manuscript generated by the AI Scientist-v2 baseline from raw materials under the sparse idea setting.}
  \label{fig:sxs_cvpr_ai_scientist}
\end{figure}

\clearpage

\begin{figure}[p]
  \centering
  \includegraphics[width=\textwidth, height=0.85\textheight, keepaspectratio]{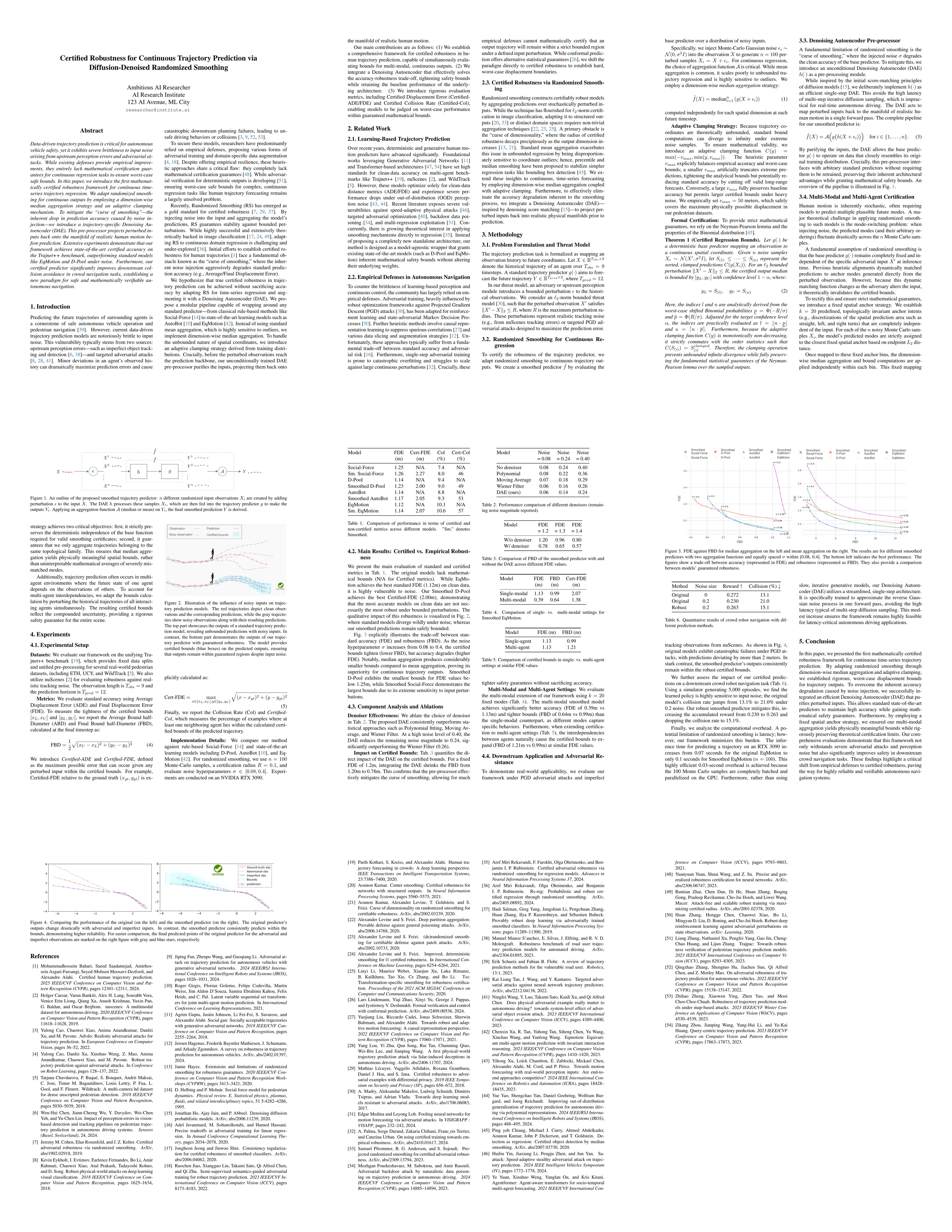}
  \caption{\textbf{CVPR Sample (\ourmethod{} - PlotOff).} Manuscript generated by \ourmethod{} (PlotOff) from raw materials under the sparse idea setting.}
  \label{fig:sxs_cvpr_ours_plotoff}
\end{figure}

\clearpage

\begin{figure}[p]
  \centering
  \includegraphics[width=\textwidth, height=0.85\textheight, keepaspectratio]{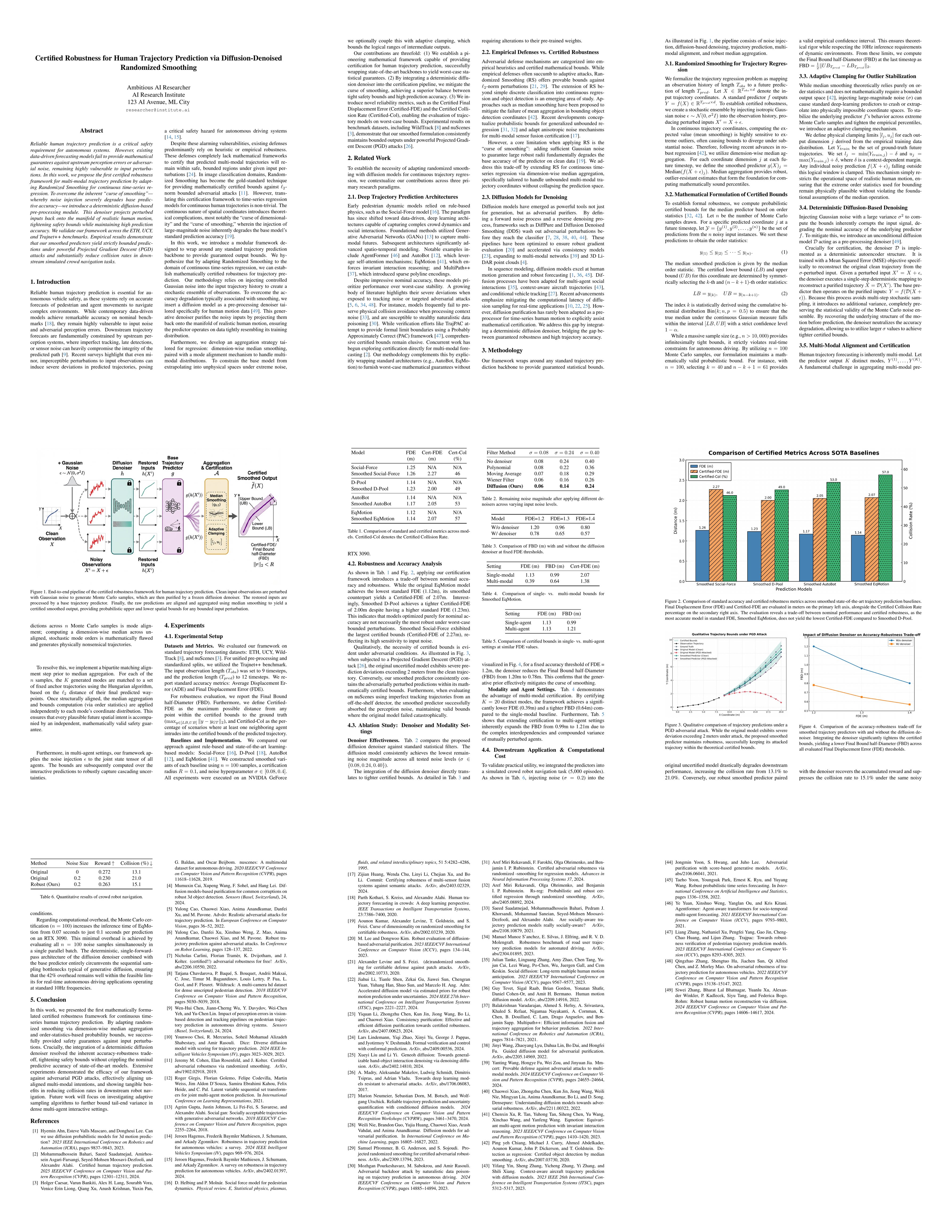}
  \caption{\textbf{CVPR Sample (\ourmethod{} - PlotOn).} Manuscript generated by \ourmethod{} (PlotOn) from raw materials under the sparse idea setting.}
  \label{fig:sxs_cvpr_ours_ploton}
\end{figure}

\clearpage


\begin{figure}[p]
  \centering
  \includegraphics[width=\textwidth, height=0.85\textheight, keepaspectratio]{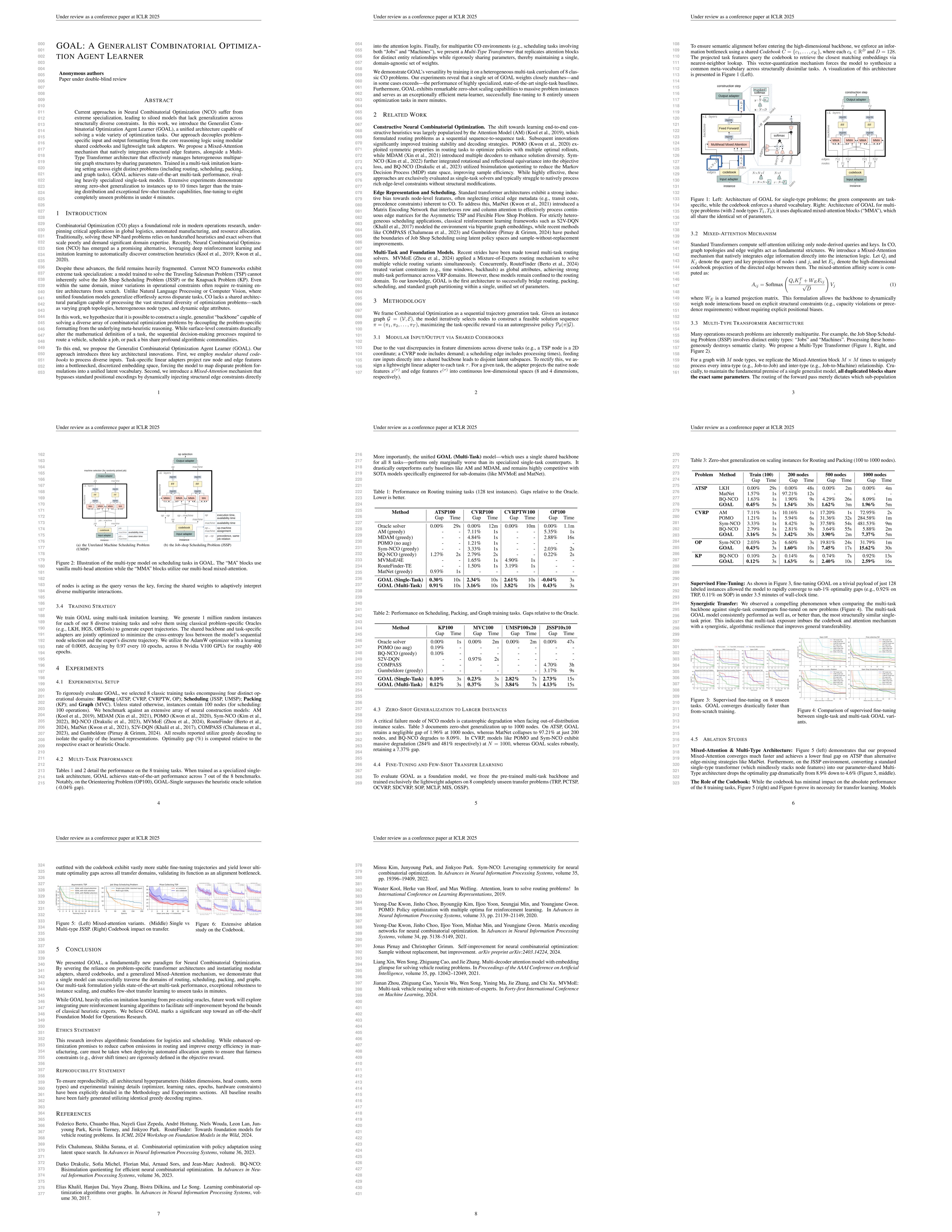}
  \caption{\textbf{ICLR Sample (Single Agent).} Manuscript generated by the Single Agent baseline from raw materials under the sparse idea setting.}
  \label{fig:sxs_iclr_single_agent}
\end{figure}

\clearpage

\begin{figure}[p]
  \centering
  \includegraphics[width=\textwidth, height=0.85\textheight, keepaspectratio]{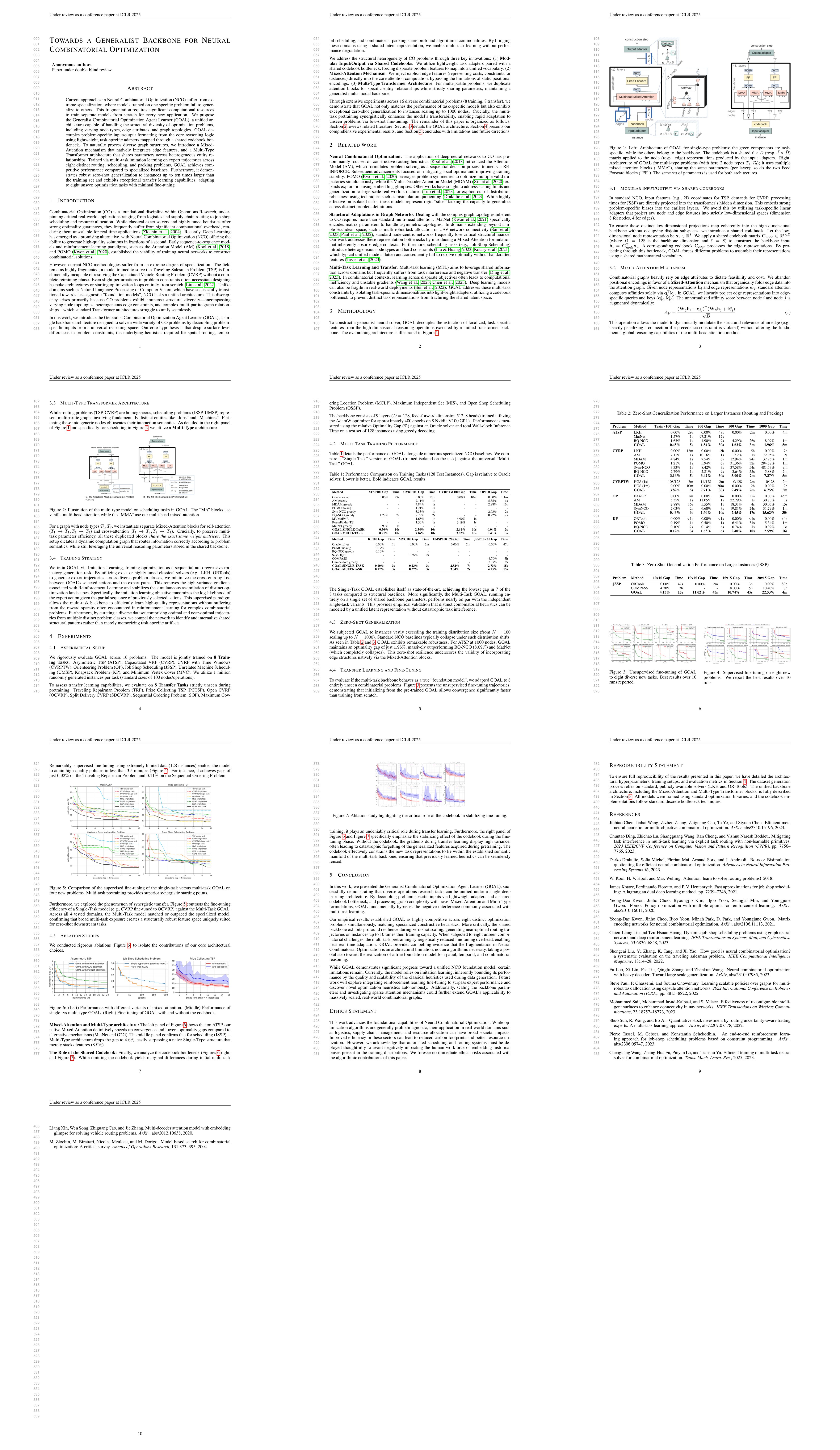}
  \caption{\textbf{ICLR Sample (AI Scientist-v2).} Manuscript generated by the AI Scientist-v2 baseline from raw materials under the sparse idea setting.}
  \label{fig:sxs_iclr_ai_scientist}
\end{figure}

\clearpage

\begin{figure}[p]
  \centering
  \includegraphics[width=\textwidth, height=0.85\textheight, keepaspectratio]{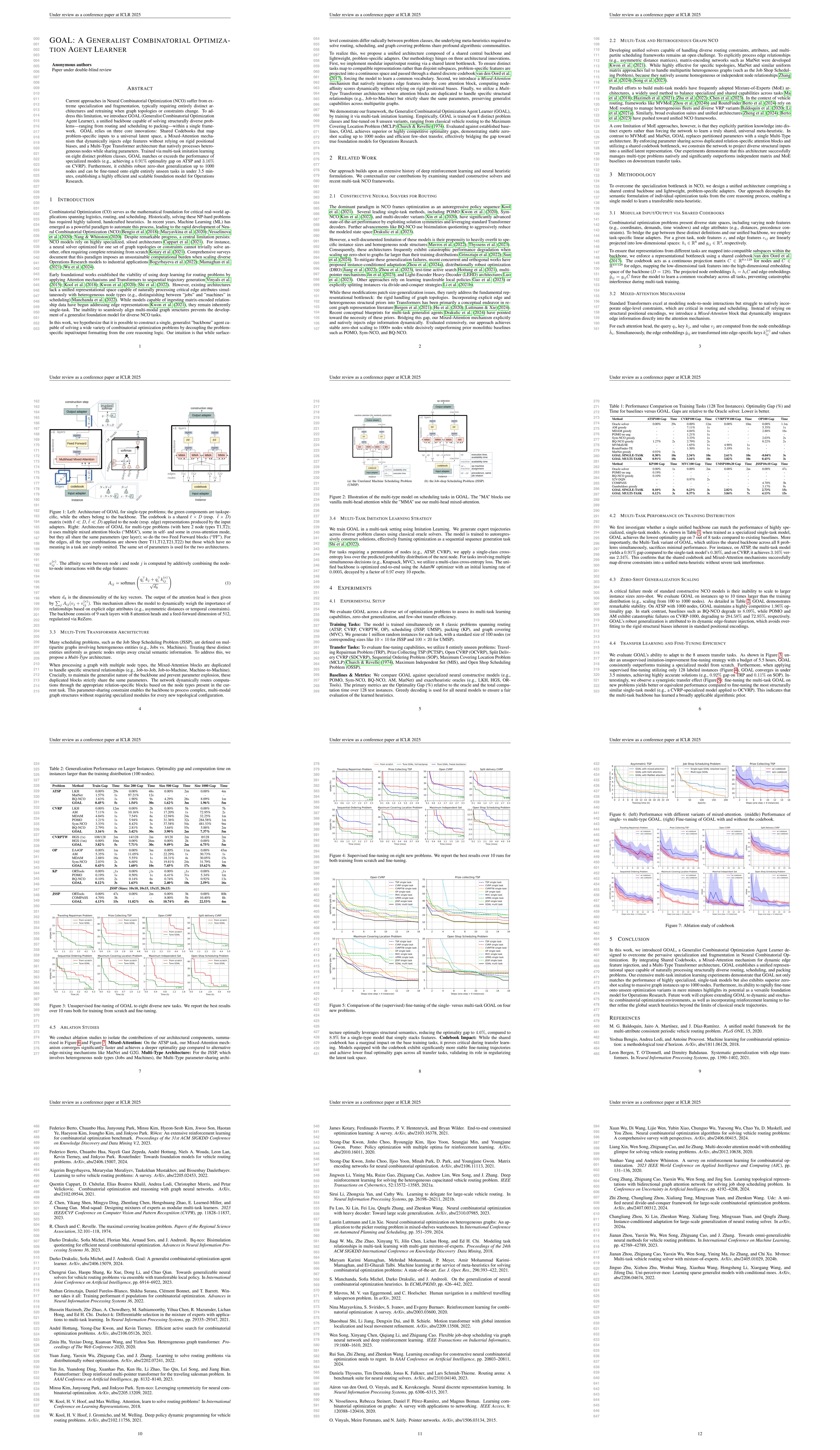}
  \caption{\textbf{ICLR Sample (\ourmethod{} - PlotOff).} Manuscript generated by \ourmethod{} (PlotOff) from raw materials under the sparse idea setting.}
  \label{fig:sxs_iclr_ours_plotoff}
\end{figure}

\clearpage

\begin{figure}[p]
  \centering
  \includegraphics[width=\textwidth, height=0.85\textheight, keepaspectratio]{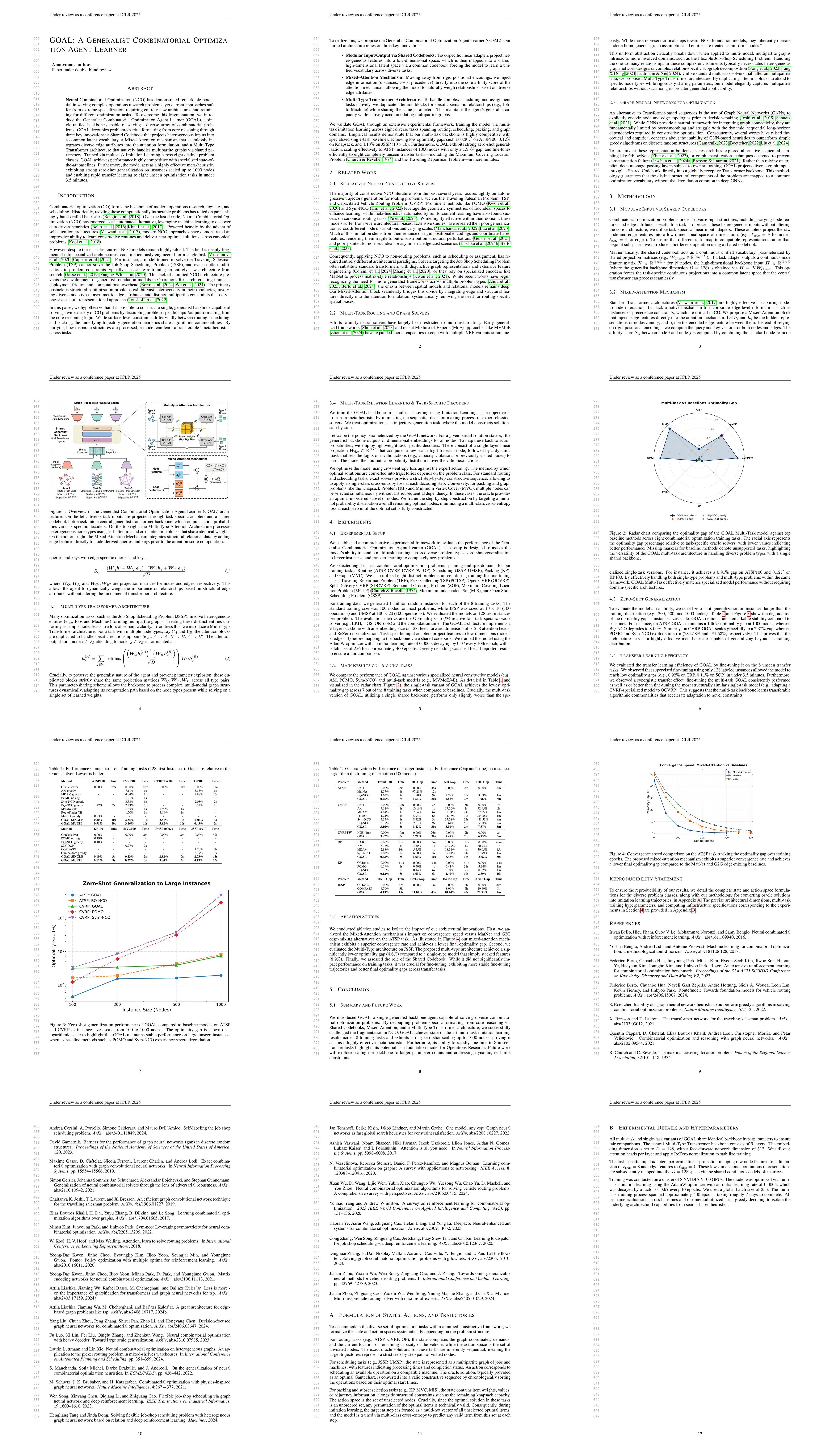}
  \caption{\textbf{ICLR Sample (\ourmethod{} - PlotOn).} Manuscript generated by \ourmethod{} (PlotOn) from raw materials under the sparse idea setting.}
  \label{fig:sxs_iclr_ours_ploton}
\end{figure}

\clearpage

\subsection{Sparse vs. Dense Idea Settings}
\label{appendix:paper_viz_sparse_dense}

To illustrate how input granularity influences technical depth, we contrast the methodology text synthesized by \ourmethod{} under the Sparse and Dense settings using a sample from the ICLR split~\citep{drakulic2025goalgeneralistcombinatorialoptimization}. While the Sparse setting produces a conceptual, high-level overview of the architecture, the Dense setting grounds the methodology in formal mathematical notation, explicitly defining the state space (e.g., BQ-MDPs) and the underlying operations.

\begin{methodologybox}[Generated Methodology (Sparse Idea)]
\noindent To overcome the specialization bottleneck in Neural Combinatorial Optimization (NCO), we design a unified architecture comprising a shared central backbone and lightweight, problem-specific adapters. Our approach decouples the semantic formulation of individual optimization tasks from the core reasoning process, enabling a single model to learn a transferable meta-heuristic. 

\vspace{1em}
\noindent\textbf{Modular Input/Output via Shared Codebooks}
\vspace{0.5em}

\noindent Combinatorial optimization problems present diverse state spaces, including varying node features (e.g., coordinates, demands, time windows) and edge attributes (e.g., distances, precedence constraints). To bridge the gap between these distinct definitions and our unified backbone, we employ task-specific linear adapters. For a given task, node features $x_i$ and edge features $e_{ij}$ are linearly projected into low-dimensional spaces: $h_i \in \mathbb{R}^8$ and $g_{ij} \in \mathbb{R}^4$, respectively. 

\vspace{0.5em}
\noindent To ensure that representations from different tasks are mapped into compatible subspaces within the backbone, we enforce a representational bottleneck using a shared codebook. The codebook acts as a continuous projection matrix $C \in \mathbb{R}^{8 \times 128}$ for nodes and $\bar{C} \in \mathbb{R}^{4 \times 128}$ for edges, mapping the low-dimensional task features into the high-dimensional embedding space of the backbone ($D=128$). The projected node embeddings $\hat{h}_i = h_i C$ and edge embeddings $\hat{g}_{ij} = g_{ij} \bar{C}$ force the model to learn a common vocabulary across all tasks, preventing catastrophic interference during multi-task training.

\vspace{1em}
\noindent\textbf{Mixed-Attention Mechanism}
\vspace{0.5em}

\noindent Standard Transformers excel at modeling node-to-node interactions but struggle to natively incorporate edge-level constraints, which are critical in routing and scheduling. Instead of relying on structural positional encodings, we introduce a \textit{Mixed-Attention} block that dynamically integrates edge information directly into the attention mechanism. 

\vspace{0.5em}
\noindent For each attention head, the query $q_i$, key $k_j$, and value $v_j$ are computed from the node embeddings $\hat{h}_i$. Simultaneously, the edge embeddings $\hat{g}_{ij}$ are transformed into edge-specific keys $k^{(e)}_{ij}$ and values $v^{(e)}_{ij}$. The affinity score between node $i$ and node $j$ is computed by additively combining the node-to-node interactions with the edge features:
\[
    A_{ij} = \text{softmax} \left( \frac{q_i^\top k_j + q_i^\top k^{(e)}_{ij}}{\sqrt{d_k}} \right)
\]
where $d_k$ is the dimensionality of the key vectors. The output of the attention head is then given by $\sum_j A_{ij} (v_j + v^{(e)}_{ij})$. This mechanism allows the model to dynamically weigh the importance of relationships based on explicit edge attributes (e.g., asymmetric distances or temporal constraints). The backbone consists of 9 such layers with 8 attention heads and a feed-forward dimension of 512, regularized via ReZero.

\vspace{1em}
\noindent\textbf{Multi-Type Transformer Architecture}
\vspace{0.5em}

\noindent Many scheduling problems, such as the Job Shop Scheduling Problem (JSSP), are defined on multipartite graphs involving heterogeneous entities (e.g., jobs vs. machines). Treating these distinct entities uniformly as generic nodes strips away crucial semantic information. To address this, we propose a \textit{Multi-Type} architecture. 

\vspace{0.5em}
\noindent When processing a graph with multiple node types, the Mixed-Attention blocks are duplicated to handle specific structural relationships (e.g., job-to-job, job-to-machine, machine-to-machine). Crucially, to maintain the generalist nature of the backbone and prevent parameter explosion, these duplicated blocks strictly share the same parameters. The network dynamically routes computations through the appropriate relation-specific blocks based on the node types present in the current task. This parameter-sharing constraint enables the backbone to process complex, multi-modal graph structures without requiring specialized modules for every new topological configuration.

\vspace{1em}
\noindent\textbf{Multi-Task Imitation Learning Strategy}
\vspace{0.5em}

\noindent We train GOAL in a multi-task setting using Imitation Learning. We generate expert trajectories across diverse problem classes using classical oracle solvers. The model is trained to autoregressively construct solutions, effectively framing optimization as a sequential generation task. 

\vspace{0.5em}
\noindent For tasks requiring a permutation of nodes (e.g., ATSP, CVRP), we apply a standard cross-entropy loss over the predicted probability distribution of the next node. For tasks involving multiple simultaneous decisions (e.g., Knapsack, MVC), we utilize a multi-class cross-entropy loss. The unified backbone is optimized end-to-end using the AdamW optimizer with an initial learning rate of $0.0005$, decayed by a factor of $0.97$ every 10 epochs.
\end{methodologybox}

\begin{methodologybox}[Generated Methodology (Dense Idea)]
\noindent We formulate the solution to multi-task combinatorial optimization as a constructive sequential decision-making process. The multi-task environment is defined as a disjoint union of Bisimulation Quotient Markov Decision Processes (BQ-MDPs), where states correspond to problem instances and actions correspond to construction steps (e.g., sequential node selections). To process highly variable environments within a single framework, GOAL decouples task-specific feature mapping from a shared reasoning backbone.

\vspace{1em}
\noindent\textbf{Task Representation and Input Adapters}
\vspace{0.5em}

\noindent For any given task $t$, we define an instance formally as a tuple $(x, \bar{x})$. Here, $x \in \mathbb{R}^{N \times F_t}$ represents the feature vectors of $N$ nodes with task-specific dimension $F_t$, and $\bar{x} \in \mathbb{R}^{N \times N \times \bar{F}_t}$ represents the feature vectors of edges with dimension $\bar{F}_t$. 

\vspace{0.5em}
\noindent Because different problems exhibit varying initial feature dimensions, we map these variables to a shared backbone embedding dimension $D$ using Task-Specific Adapters combined with a Shared Linear Projection. To prevent isolated subspace specialization, we enforce low-rank projections:

\vspace{0.5em}
\noindent 1. \textbf{Input Adapter:} A task-specific linear layer maps the raw inputs down to a low-dimensional space $\ell \ll D$.

\vspace{0.5em}
\noindent 2. \textbf{Shared Linear Projection:} A universally learnable matrix of size $\ell \times D$ projects this compressed representation into the shared backbone space. (Note: this component is labeled as a `codebook' in figures for visual brevity, though it acts purely as a continuous linear projection and not a discrete quantization dictionary).

\vspace{1em}
\noindent\textbf{Multi-Head Mixed-Attention Mechanism}
\vspace{0.5em}

\noindent The core reasoning engine of GOAL is a Transformer stack utilizing a Multi-Head Mixed-Attention mechanism. Unlike standard transformers that rely solely on node features, our mechanism natively ingests edge features $\bar{x}$ directly into the attention score computation. 

\vspace{0.5em}
\noindent For a given attention head $h$, let $M$ be the number of query nodes and $N$ be the number of key/value nodes. The standard projections for node queries ($Q$), keys ($K$), and values ($V$) are defined as:
\[
Q_n^{(h)} = Q_n \mathbf{W}_Q^{(h)}, \quad K_m^{(h)} = K_m \mathbf{W}_K^{(h)}, \quad V_m^{(h)} = V_m \mathbf{W}_V^{(h)}
\]
Simultaneously, we define projections for the edge features $E_{mn}$ (derived from $\bar{x}$) into the query and key spaces:
\[
Q_{mn}^{\prime (h)} = E_{mn} \mathbf{W}_Q^{\prime (h)}, \quad K_{mn}^{\prime (h)} = E_{mn} \mathbf{W}_K^{\prime (h)}
\]
The attention scores $S_{mn}^{(h)}$ are then computed by adding the edge-derived vectors to the node-derived vectors \textit{before} the scalar product:
\[
S_{mn}^{(h)} = \langle K_m^{(h)} + K_{mn}^{\prime (h)} \mid Q_n^{(h)} + Q_{mn}^{\prime (h)} \rangle
\]
This formulation structurally integrates relational data without incurring computational overhead when such features are absent, as the edge terms simply evaluate to zero. 

\vspace{0.5em}
\noindent While our mechanism shares mathematical foundations with relative positional encodings and other edge-augmented models, it fundamentally differs in application. Instead of learning 1D scalar biases based on graph distance, our Mixed-Attention dynamically projects dense, multi-dimensional asymmetric edge features (such as explicit transition costs or capacity constraints) directly into the high-dimensional query and key vector spaces before the scalar product. 

\vspace{1em}
\noindent\textbf{Multi-Type Transformer Architecture}
\vspace{0.5em}

\noindent To handle combinatorial problems defined on heterogeneous graphs (e.g., bipartite connections mapping ``Jobs'' to ``Machines'' in scheduling tasks), we extend the backbone into a Multi-Type Architecture. Rather than padding disparate types into a single homogeneous tensor, we dynamically compose layer operations based on the inherent task structure.

\vspace{0.5em}
\noindent We instantiate separate self-attention blocks for each distinct node type, and cross-attention blocks for valid type pairings. Crucially, to maintain the generalist nature of the model, all attention blocks within a specific layer share the \textit{exact same parameters} ($\mathbf{W}_Q, \mathbf{W}_K, \mathbf{W}_V$, etc.), regardless of whether they are performing self-attention or cross-attention. This forces disparate graph typologies to map into a unified algorithmic reasoning space.

\vspace{0.5em}
\noindent One might consider an alternative approach: concatenating all nodes into a single sequence and utilizing a block-sparse attention mask. However, standard transformers with sparse masks often still allocate memory-intensive full-sequence tensors ($N \times N$) before applying the mask. Our Multi-Type architecture avoids this by explicitly instantiating structured attention blocks \textit{only} for valid topological pairings, bypassing the quadratic computational waste associated with invalid interactions. 

\vspace{1em}
\noindent\textbf{Training Protocol and Decoding}
\vspace{0.5em}

\noindent GOAL is trained in a multi-task setting via Imitation Learning. To train the model, we require step-by-step constructive trajectories derived from optimal or highly optimized heuristic oracles. 

\vspace{0.5em}
\noindent \textbf{Trajectory Canonicalization:} For sequential routing tasks with bidirectional symmetries (e.g., the TSP, where a forward and backward tour represent the same cycle), we canonicalize the expert trajectories to prevent conflicting teaching signals. We enforce a fixed traversal direction by always starting at the depot and breaking initial directional ties by visiting the lowest-indexed node first.

\vspace{0.5em}
\noindent For unordered tasks such as the Knapsack Problem (KP) and Minimum Vertex Cover (MVC), expert solvers output static sets of selected items. To convert these into sequential trajectories without masking optimal items, we take the exact optimal subset $S^*$ provided by the oracle, and strictly order \textit{only} the elements within $S^*$ by a defined heuristic (e.g., value-to-weight ratio for KP, residual degree for MVC). This translates the unordered solution into a deterministic, valid trajectory that the policy can effectively imitate.

\vspace{0.5em}
\noindent We minimize the cross-entropy loss between the model's predicted action probabilities and these optimal trajectories. A task-specific linear projection maps the final backbone output to a score matrix representing the log-logits for valid node selection. A softmax operation is then applied over the dynamically updated valid action mask to yield the final policy distribution.

\vspace{0.5em}
\noindent \textbf{Autoregressive Decoding and Efficiency:} To maintain fast inference times scaling up to 1000-node instances, GOAL does \textit{not} re-evaluate the entire Transformer backbone at every autoregressive step. Instead, the heavy 9-layer backbone processes the static graph features (e.g., node coordinates, distance matrices) exactly once, caching the high-dimensional node and edge embeddings. At each constructive step, a lightweight Output Adapter takes the cached embeddings along with the rapidly changing dynamic state features (e.g., current location, remaining vehicle capacity) to compute the next valid action. This cached decoding strategy prevents quadratic recalculations at each step.
\end{methodologybox}
\section{Prompts}
\label{appendix:prompts}


\subsection{Paper Writing}
\label{appendix:paper_writing_prompts}
In this section, we detail the prompts used for each agent involved in the paper writing pipeline.


\begin{paperpromptbox}[Outline Agent]
You are a senior AI researcher drafting a paper for a top-tier conference (e.g., NeurIPS, ICML, CVPR, ICLR). 
Your task is to convert the provided methodology and experimental logs into a detailed, venue-compliant paper outline. You must output a single JSON object.

\vspace{1em}
\noindent\textbf{Your inputs are:}
\begin{enumerate}
    \item \texttt{idea.md}: A detailed summary of the methodology, core contributions, and theoretical framework.
    \item \texttt{experimental\_log.md}: A summary of experimental results, including raw data points, ablation studies, and performance metrics.
    \item \texttt{template.tex}: The target structure. You must use the section commands (e.g., \textbackslash{}section\{...\}) found here as your primary skeleton.
    \item \texttt{conference\_guidelines.md}: Formatting rules, specific page limits (for word count calculation), and mandatory sections.
\end{enumerate}

\vspace{0.5em}
\noindent\textbf{Processing Directives}
\vspace{0.5em}

\noindent\textbf{Global Instruction:} Do not analyze inputs in isolation. You must synthesize information across all provided documents for every step.

\vspace{1em}
\noindent\textbf{Directive 1: Plotting \& Visualization Plan}\\
\vspace{0.5em}
Synthesize \texttt{experimental\_log.md} and \texttt{idea.md} to identify the most compelling evidence.
\begin{itemize}
    \item Determine which figures are essential to visually prove the hypothesis (e.g., convergence rates, qualitative visual comparisons).
    \item The \texttt{plot\_type} MUST be exactly "plot" or "diagram". If it is a plot, specify the specific chart type (e.g., Radar Chart) inside the \texttt{objective}.
    \item The \texttt{data\_source} MUST be exactly "idea.md", "experimental\_log.md", or "both".
    \item Determine the ideal \texttt{aspect\_ratio} for each figure. The \texttt{aspect\_ratio} MUST be exactly one of: "1:1", "1:4", "2:3", "3:2", "3:4", "4:1", "4:3", "4:5", "5:4", "9:16", "16:9", "21:9".
    \item The \texttt{figure\_id} MUST be a semantically meaningful string identifier summarizing the plot contents, like "fig\_framework\_overview" or "fig\_ablation\_study\_parameter\_sensitivity". It MUST NOT contain the word "Figure".
    \item Output Focus: Create an array of objects for the \texttt{plotting\_plan} key.
\end{itemize}

\vspace{1em}
\noindent\textbf{Directive 2: Research Graph \& Investigation Strategy (Intro \& Related Work)}\\
\vspace{0.5em}
Provide search instructions for a downstream literature review agent to build a Research Graph. Do not write the actual paper content.

\vspace{0.5em}

\noindent Prevent Citation Overlap: Strictly separate the scope of the Introduction from Related Work to ensure the agent searches for different tiers of literature.
\begin{itemize}
    \item Introduction: Focuses on macro-level context (foundational papers, surveys).
    \item Related Work: Focuses on micro-level technical comparisons (recent SOTA baselines, benchmarks). 
\end{itemize}

\noindent Introduction Strategy (Macro-Level Context, 10-20 papers):
\begin{itemize}
    \item Hypotheses: Define the "Hook" (broad context) and "Problem Gap" to be verified. CRITICAL: Strictly scope the problem gap and claims to match the specific datasets and evaluations present in \texttt{experimental\_log.md}. Do not over-claim generalization.
    \item Search Directions: Provide 3-5 specific queries to find: 
    \begin{enumerate}
        \item Papers establishing the real-world impact or urgency of the problem gap.
        \item Good survey or review papers on the topic.
        \item 3-5 Foundational papers that established the sub-field.
    \end{enumerate}
\end{itemize}

\noindent Related Work Strategy (Micro-Level Technical Baselines, 30-50 papers):
\begin{itemize}
    \item Divide the field into 2-4 distinct methodology clusters that directly compete with or precede our approach.
    \item For each cluster, define:
    \begin{enumerate}
        \item Methodology Cluster Name: The technical category.
        \item SOTA Investigation: Instructions to find recent papers for conceptual context. CRITICAL TIMELINE RULE: Do not instruct searches for any papers published after \{cutoff\_date\}. Furthermore, do NOT instruct the search for new "competitors" to beat if they are not exclusively in \texttt{experimental\_log.md}.
        \item Limitation Hypothesis: The suspected failure point of these competing methods, based on \texttt{idea.md}.
        \item Limitation Search Queries: Highly specific, narrow queries to find papers documenting these exact limitations.
        \item The Bridge: How our proposed method resolves this specific limitation.
    \end{enumerate}
\end{itemize}
Output Focus: Populate the \texttt{intro\_related\_work\_plan} key.

\vspace{1em}
\noindent\textbf{Directive 3: Section Writing Plan \& Sizing Constraints}\\
\vspace{0.5em}
Outline the remaining sections (Abstract, Methodology, Experiments, Conclusion, Appendix) into a detailed structural plan.

\begin{itemize}
    \item Structural Hierarchy: If Subsection X.1 is created, X.2 is mandatory. Do not create orphaned subsections. Omit subsections entirely if a section does not require division.
    \item Content Specificity: Explicitly reference source materials. 
    \begin{itemize}
        \item \textit{Avoid:} "Describe the model."
        \item \textit{Require:} "Formalize the Temporal-Aware Attention mechanism using Eq. 3 from \texttt{idea.md}."
    \end{itemize}
    \item Mandatory Citations (\texttt{citation\_hints}): You must provide targeted citation hints for all external dependencies. Every hint must point to a single, unambiguous canonical paper. 
    \begin{itemize}
        \item Required Coverage (EXHAUSTIVE): You MUST explicitly create a targeted \texttt{citation\_hints} query for EVERY SINGLE dataset, optimizer, metric, and foundational architecture/model you mention, no matter how ubiquitous or obvious it seems (e.g., AdamW, ResNet, ImageNet, CLIP, Transformer, LLaMA, GPT, LLaVA). If it is in the \texttt{experimental\_log.md} or \texttt{idea.md}, it MUST have a citation hint.
        \begin{enumerate}
            \item All baseline methods compared against.
            \item All datasets evaluated on.
            \item All standard metrics utilized.
            \item All foundational algorithms, architectures (e.g., ResNet, Transformer), foundational models (e.g., LLMs, VLMs, Diffusion models), optimizers (e.g., AdamW), or frameworks built upon.
        \end{enumerate}
        \item Format Constraint \& Anti-Hallucination Rule: If you know the exact author and title, use "Author (Exact Paper Title)". DO NOT guess or hallucinate authors. If you do not know the exact author, use this format: "research paper or technical report introducing '[Exact Model/Dataset/Metric Name]'".
    \end{itemize}
    \item Output Focus: Populate the \texttt{section\_plan} key.
\end{itemize}

\noindent Guidelines on Scientific Depth \& Mathematical Rigor:
\begin{itemize}
    \item Grounded Formalization: Propose explicit subsections for rigorous mathematical formulations (e.g., loss functions, core algorithms, theoretical proofs). You must base these strictly on \texttt{idea.md} and \texttt{experimental\_log.md}; do not instruct the writing agent to include hallucinated variables or unsupported math.
\end{itemize}

\vspace{0.5em}
\noindent\textbf{Strict Output Format (JSON)}\\
You must output a single, valid JSON object with the following three top-level keys: "plotting\_plan", "intro\_related\_work\_plan", and "section\_plan".

\vspace{0.5em}
\noindent Example Output:

\begin{verbatim}
{
  "plotting_plan": [
    {
      "figure_id": "fig_teaser_fig_cross_modal_alignment_performance",
      "title": "Teaser: Cross-Modal Alignment Performance",
      "plot_type": "plot",
      "data_source": "experimental_log.md",
      "objective": "Visual summary (Radar Chart) demonstrating that our method 
                   achieves SOTA balance across 5 metrics.",
      "aspect_ratio": "16:9"
    }
  ],
  "intro_related_work_plan": {
    "introduction_strategy": {
      "hook_hypothesis": "Video-LLMs are currently the dominant paradigm for 
                         short clips.",
      "problem_gap_hypothesis": "Context window limits prevent scaling to >5s 
                                videos efficiently.",
      "search_directions": [
        "Find highly cited papers establishing the real-world impact of 
         context limits in video generation",
        "Search for published 'long-context video generation' surveys",
        "Identify foundational papers establishing causal video generation"
      ]
    },
    "related_work_strategy": {
      "overview": "Investigate three specific paradigms to build a graph 
                  proving the necessity of our Sliding-Window approach.",
      "subsections": [
        {
          "subsection_title": "2.1 Autoregressive Video Generation",
          "methodology_cluster": "Discrete Tokenization & Transformers",
          "sota_investigation_mission": "Identify the current SOTA 
                                        autoregressive models from 2024-2025. 
                                        Determine their maximum stable 
                                        generation length.",
          "limitation_hypothesis": "These models suffer from 'drift' or 'error 
                                   propagation' because they lack bidirectional 
                                   context.",
          "limitation_search_queries": [
            "Autoregressive video generation error propagation metrics",
            "Causal masking limitations in temporal video transformers"
          ],
          "bridge_to_our_method": "Our method introduces bidirectional blocks 
                                  to fix the hypothesized drift issue."
        },
        {
          "subsection_title": "2.2 Diffusion-Based Editing Frameworks",
          "methodology_cluster": "DDIM Inversion & Cross-Attention",
          "sota_investigation_mission": "Find recent papers using DDIM 
                                        inversion for editing. Identify the 
                                        standard benchmarks they use.",
          "limitation_hypothesis": "They fail at large structural changes 
                                   because cross-attention maps are too rigid.",
          "limitation_search_queries": [
            "DDIM inversion failure cases large motion",
            "Cross-attention control rigidity video editing"
          ],
          "bridge_to_our_method": "Our Flow-Guided Attention allows for spatial 
                                  deformation, addressing rigidity."
        }
      ]
    }
  },
  "section_plan": [
    {
      "section_title": "Abstract",
      "subsections": [
        {
          "subsection_title": "Abstract Content",
          "content_bullets": [
            "Briefly state the problem of temporal inconsistency.",
            "Introduce the proposed method.",
            "Highlight key results."
          ],
          "citation_hints": []
        }
      ]
    },
    {
      "section_title": "3. Methodology",
      "subsections": [
        {
          "subsection_title": "3.1 Temporal-Aware Attention Mechanism",
          "content_bullets": [
            "Define the query-key matching logic", 
            "Explain the masking strategy"
          ],
          "citation_hints": [
            "Vaswani et al. (Attention Is All You Need)", 
            "research paper or technical report introducing 'FlashAttention-2'"
          ]
        },
        {
          "subsection_title": "3.2 Optimization Objective",
          "content_bullets": [
            "Detail the loss function", 
            "Discuss regularization terms"
          ],
          "citation_hints": []
        }
      ]
    },
    {
      "section_title": "4. Experiments",
      "subsections": [
        {
           "subsection_title": "4.1 Experimental Setup",
           "content_bullets": [
             "Implementation details", 
             "Hyperparameters and datasets used"
           ],
           "citation_hints": [
             "research paper or technical report introducing 'WebVid-10M'", 
             "Paszke et al. (PyTorch: An Imperative Style, High-Performance 
              Deep Learning Library)",
             "research paper or technical report introducing 'AdamW optimizer'",
             "research paper or technical report introducing 'Jaccard Index'"
           ]
        },
        {
           "subsection_title": "4.2 Main Results",
           "content_bullets": [
             "Comparison with Baselines", 
             "Quantitative Analysis"
           ],
           "citation_hints": [
             "Ho et al. (Denoising Diffusion Probabilistic Models)",
             "research paper or technical report introducing 
             'AVSegFormer baseline'"
           ]
        }
      ]
    }
  ]
}
\end{verbatim}
\end{paperpromptbox}


For the plotting agent, we utilize the original prompts from PaperBanana~\citep{zhu2026paperbanana}, appending only a single caption regeneration step at the end:

\begin{paperpromptbox}[Plotting Agent (Caption Generation)]
\noindent\textbf{Input Data}
\begin{itemize}
    \item Task Type: \{task\_name\}
    \item Contextual Section: \{raw\_content\}
    \item Overall Figure Intent: \{description\}
    \item Detailed Figure Description: \{figure\_desc\}
\end{itemize}

\noindent Please provide the final caption for this figure based on the system instructions.

\vspace{1em}
\noindent\textbf{Requirements}
\begin{itemize}
    \item The caption should be concise and informative, and can be directly used as a caption for academic papers.
    \item The caption MUST NOT contain a "Figure X:" or "Caption X:" prefix, as the latex template will add it automatically.
    \item The caption MUST NOT contain any markdown formatting (like bold, italics, etc), it should be plain text.
\end{itemize}

\vspace{0.5em}
\noindent Respond with the plain text caption only.
\end{paperpromptbox}


\begin{paperpromptbox}[Literature Review Agent]
\noindent\textbf{Role:} Senior AI Researcher.\\
\vspace{1em}
\textbf{Task:} Write the introduction and related work section of a paper.

\vspace{1em}
\noindent You will be given a \texttt{template.tex}, this is the initial skeleton we outlined for you. 
Your job is to fill in two sections: Introduction and Related Work. Leave all the other sections untouched.

\vspace{1em}
\noindent\textbf{Inputs:}
\begin{itemize}
    \item \texttt{intro\_related\_work\_plan}: This is your PRIMARY guide for structure and arguments.
    \item \texttt{project\_idea} and \texttt{project\_experimental\_log}: Use them to ensure the Intro accurately frames the technical contribution and results.
    \item \texttt{citation\_checklist}: This includes the citation keys that you should use when citing relevant papers.
    \item \texttt{collected\_papers}: These are all the relevant papers we collect for you for citation purpose.
\end{itemize}

\vspace{0.5em}
\noindent YOU MUST ONLY CITE THE GIVEN \texttt{collected\_papers}, DO NOT cite new papers other than the given papers.

\vspace{1em}
\noindent\textbf{Citation Requirements:}
\begin{itemize}
    \item You have access to the abstract of \{paper\_count\} collected papers.
    \item You MUST cite at least \{min\_cite\_paper\_count\} of them across the introduction and related work sections.
    \item Introduction: Cite key statistics, foundational models (CLIP, etc.), and broad problem statements.
    \item Related Work: Do deep comparative citations. Group distinct works (e.g., "Several methods [A, B, C]...").
    \item Ensure every \textbackslash{}cite\{\{key\}\} corresponds exactly to a key in \texttt{citation\_checklist}.
    \item CRITICAL TIMELINE RULE: Do not treat any papers published after \{cutoff\_date\} as prior baselines to beat. Treat them strictly as concurrent work.
    \item CRITICAL EVALUATION RULE: Do not claim our method beats or achieves State-of-the-Art over a specific cited paper UNLESS that paper is explicitly evaluated against in \texttt{project\_experimental\_log}. Frame other recent papers strictly as concurrent, orthogonal, or conceptual work.
    \item You need to return the full code for the new \texttt{template.tex}, where the two empty sections (Introduction and Related Work) are now filled in, while all the other code (packages, styles, and other sections) are identical to the original \texttt{template.tex}.
\end{itemize}

\vspace{1em}
\noindent\textbf{Important Note:}\\
DO NOT change \texttt{\textbackslash{}usepackage[capitalize]\{\{cleveref\}\}} into \texttt{\textbackslash{}usepackage[capitalize]\{\{cleverref\}\}}, as there's no \texttt{cleverref.sty}.

\vspace{1em}
\noindent\textbf{Output Format:}\\
You must return the code for the updated \texttt{template.tex}. Make sure to wrap the code with \verb|```latex content ```|.
\end{paperpromptbox}

\begin{paperpromptbox}[Section Writing Agent]
\noindent\textbf{Role:} Senior AI Researcher.\\
\vspace{1em}
\textbf{Task:} Complete a research paper by writing the missing sections in a LaTeX template.

\vspace{0.5em}
\noindent You will be given a \texttt{template.tex} file where some sections (e.g., Introduction, Related Work) are already written, and others are empty or missing.
Your job is to generate the LaTeX code for the missing sections only, based on the provided \texttt{outline.json}, and merge them into the final document.

\vspace{1em}
\noindent\textbf{Inputs}
\begin{itemize}
    \item \texttt{outline.json}: Your MASTER PLAN. Defines section hierarchy, points to cover, and which papers to consider citing (\texttt{citation\_candidates}).
    \item \texttt{idea.md}: Technical details of the methodology.
    \item \texttt{experimental\_log.md}: Raw data for tables and qualitative analysis for text.
    \item \texttt{citation\_map.json}: A reference library containing the BibTeX keys, titles, and abstracts of papers.
    \item \texttt{conference\_guidelines.md}: Formatting rules.
    \item \texttt{figures\_list}: Available figure files.
\end{itemize}

\vspace{1em}
\noindent\textbf{Critical Instructions}
\begin{enumerate}
    \item \textbf{Existing Content Preservation:} 
    \begin{itemize}
        \item DO NOT modify the text, style, or content of sections that are already filled in \texttt{template.tex}. 
        \item Come up with a good title if it is missing, fill in the author names if missing.
        \item Keep the preamble (packages) exactly as is.
    \end{itemize}
    \item \textbf{Data \& Tables:}
    \begin{itemize}
        \item You are responsible for creating LaTeX tables.
        \item Extract numerical data directly from \texttt{experimental\_log.md}.
        \item Use the \texttt{booktabs} package format (\textbackslash{}toprule, \textbackslash{}midrule, \textbackslash{}bottomrule).
        \item Do not hallucinate numbers. Use the exact values provided in the log.
        \item Make sure all tables appear before the Conclusion section, unless they are placed in an Appendix.
    \end{itemize}
    \item \textbf{Citations:}
    \begin{itemize}
        \item The \texttt{outline.json} provides a list of \texttt{citation\_candidates} for specific subsections.
        \item You MUST use the exact keys found in \texttt{citation\_map.json} (e.g., \textbackslash{}cite\{Hu2021LoraLowrank\}).
        \item Content Enrichment: Read the abstract provided in \texttt{citation\_map.json} for the papers you are citing. Use this context to write accurate, specific sentences about those works.
    \end{itemize}
    \item \textbf{Writing Content:}
    \begin{itemize}
        \item Write the missing sections following the \texttt{outline.json} structure.
        \item Use formal mathematical equations, notations, and definitions where appropriate and directly supported by the idea/log. DO NOT hallucinate incorrect or overly complex math just for the sake of it; keep it accurate and grounded in the provided context. Avoid overly colloquial summaries.
        \item Always provide detailed ablation studies and qualitative analysis of the experimental results: what works, what does not, and why.
        \item Nice to have: discuss the limitations and future work at the end.
        \item If you want to put anything in the Appendix, make sure the Appendix section appears after the References section, on a fresh new page.
    \end{itemize}
    \item \textbf{Figures And Visual Fidelity:}
    \begin{itemize}
        \item You are being provided with the actual image files of the figures. You MUST describe them faithfully and accurately. DO NOT hallucinate interpretations that contradict the visual evidence in the plots.
        \item Make sure to use ALL of the figures provided in \texttt{figures\_list}. Note: figures are stored in the \texttt{figures/} subdirectory. IMPORTANT: use the exact filenames including their extensions (e.g., \texttt{.png}) in your \textbackslash{}includegraphics commands.
        \item DO NOT merge or group multiple figures into one for display.
        \item If the paper is in a 2-column format, try displaying figures in single-column mode (\textbackslash{}begin\{figure\}) unless they are very wide.
        \item Ensure that all figures are correctly referenced in the text. 
        \item Make sure all figures appear before the Conclusion section, unless they are placed in an Appendix.
        \item You can refine the captions if necessary.
        \item Do not include "Figure x" in the caption text; the LaTeX template will handle the figure numbering.
    \end{itemize}
    \item \textbf{Style:}
    \begin{itemize}
        \item Adopt the tone of a top-tier ML conference paper: dense, objective, and technical.
        \item Ensure your new LaTeX code matches the indentation and spacing style of the \texttt{template.tex}. Do not change the given style.
    \end{itemize}
\end{enumerate}

\vspace{1em}
\noindent\textbf{Output Format}
\begin{itemize}
    \item Return the full code for the completed \texttt{template.tex}.
    \item The sections that were previously empty should now be filled.
    \item The sections that were previously filled should remain mostly untouched; only adjust for consistency purposes.
    \item Wrap the code with \verb|```latex content ```|.
\end{itemize}

\vspace{1em}
\noindent\textbf{Important Note}\\
\vspace{0.5em}
DO NOT change \texttt{\textbackslash{}usepackage[capitalize]\{\{cleveref\}\}} \\
into \texttt{\textbackslash{}usepackage[capitalize]\{\{cleverref\}\}}, as there is no \texttt{cleverref.sty}.\\
Ensure the LaTeX code compiles without errors, e.g., all the begin and end statements match correctly (e.g., \textbackslash{}begin\{figure*\} must be closed with \textbackslash{}end\{figure*\}, not \textbackslash{}end\{figure\}).
\end{paperpromptbox}


\begin{paperpromptbox}[Content Refinement Agent]
\noindent\textbf{Role:} Senior AI Researcher.\\
\vspace{1em}
\textbf{Task:} Revise and strengthen a LaTeX research paper by systematically addressing peer review feedback.

\vspace{0.5em}
\noindent You are the author responsible for the "Rebuttal via Revision" phase. You will receive:
\begin{itemize}
    \item \texttt{paper.tex}: The current LaTeX source code.
    \item \texttt{paper.pdf}: The compiled PDF context.
    \item \texttt{conference\_guidelines.md}: The formatting and page limit rules.
    \item \texttt{experimental\_log.md}: The Ground Truth for all data and metrics.
    \item \texttt{worklog.json}: History of previous changes.
    \item \texttt{citation\_map.json}: The allowed bibliography.
    \item \texttt{reviewer\_feedback}: A JSON object containing specific Strengths, Weaknesses, Questions, and Decisions from an LLM reviewer.
\end{itemize}

\vspace{1em}
\noindent\textbf{Your Goal}
\begin{enumerate}
    \item \textbf{Analyze Feedback:} Deconstruct the \texttt{reviewer\_feedback} into actionable editing tasks.
    \item \textbf{Address Weaknesses:} Rewrite sections to clarify logic, strengthen arguments, or justify design choices pointed out as weak.
    \item \textbf{Integrate Answers:} Incorporate answers to the reviewer's "Questions" directly into the manuscript (e.g., adding training cost details to the Implementation section).
    \item \textbf{Execution:} Generate a JSON worklog of your editorial decisions and the full, revised LaTeX source.
\end{enumerate}

\vspace{1em}
\noindent\textbf{Critical Execution Standards}
\begin{enumerate}
    \item \textbf{Content Revision Strategy}
    \begin{itemize}
        \item \textbf{Weakness Mitigation:} If the reviewer flags "incremental novelty," rewrite the Introduction and Related Work to explicitly contrast your contribution against prior art. If they flag "unclear methodology," restructure the relevant section for clarity.
        \item \textbf{Answering Questions:} Do NOT write a separate response letter. If the reviewer asks "What is the inference latency?", you must find a natural place in the paper (e.g., Experiments or Discussion) to insert that information, ensuring it aligns with \texttt{experimental\_log.md}.
        \item \textbf{Preserve Strengths:} Do not delete or heavily alter sections listed under "Strengths" unless necessary for space or flow.
    \end{itemize}

    \item \textbf{Data Integrity \& Hallucination Check}
    \begin{itemize}
        \item \textbf{Ground Truth:} All numerical claims (accuracy, parameter count, training hours, latency) MUST be verified against \texttt{experimental\_log.md}.
        \item \textbf{Missing Data:} If the reviewer asks for new experiments, ablations, or baselines that are NOT in \texttt{experimental\_log.md}, simply ignore those specific requests. Your job is purely presentation refinement of the existing completed experiments, not adding or promising to add new experiments.
    \end{itemize}

    \item \textbf{Writing Style \& Tone}
    \begin{itemize}
        \item \textbf{Academic Tone:} Maintain a formal, objective, and precise tone. Avoid defensive language.
        \item \textbf{Conciseness:} If the paper is near the page limit, prioritize density of information over flowery prose.
        \item \textbf{Flow:} Ensure that new insertions (answers to questions) transition smoothly with existing text.
    \end{itemize}

    \item \textbf{LaTeX \& Citation Integrity}
    \begin{itemize}
        \item \textbf{Structure:} Do not break the LaTeX compilation. Keep packages and environments stable. If using \texttt{figure*} for wide figures, ensure they are closed with \textbackslash{}end\{figure*\} (not \textbackslash{}end\{figure\}). Check for completeness.
        \item \textbf{Citations:} Use ONLY keys from \texttt{citation\_map.json}.
    \end{itemize}
\end{enumerate}

\vspace{1em}
\noindent\textbf{Output Format (Strict)}\\
\vspace{0.5em}
You MUST return your response in two distinct code blocks in this exact order:

\begin{enumerate}
    \item \textbf{Worklog for the current turn (JSON):}
\begin{verbatim}
{
  "addressed_weaknesses": [
    "Clarified contribution novelty in Intro (Reviewer point 2)",
    "Added justification for two-stage training (Reviewer point 1)"
  ],
  "integrated_answers": [
    "Added training cost (45 GPU hours) to Implementation Details",
    "Added epsilon hyperparameter explanation to Method section"
  ],
  "actions_taken": [
    "Rewrote Section 3.2 for clarity",
    "Inserted new paragraph in Section 5.1 regarding latency"
  ]
}
\end{verbatim}

    \item \textbf{The FULL revised LaTeX code:}
\begin{verbatim}
```latex
... Full revised LaTeX code here ...
```
\end{verbatim}
\end{enumerate}

\vspace{1em}
\noindent\textbf{Important Notes}
\begin{itemize}
\item \textbf{Completeness:} Always provide the FULL LaTeX code. Do not return diffs or partial snippets.
\item \textbf{Responsiveness:} Every question in the \texttt{reviewer\_feedback} must be addressed by improving the presentation, EXCEPT for questions asking for new experiments or data not in \texttt{experimental\_log.md} (which should be ignored). Never explicitly state a limitation.
\item \textbf{Safety:} Do not remove the \textbackslash{}documentclass or essential preamble.
\end{itemize}
\end{paperpromptbox}

We explicitly instruct the \textit{Content Refinement Agent} to ignore reviewer requests for additional experiments. This constraint is crucial to prevent the agent from generating fabricated results or making false promises within the paper to conduct future experiments, as the agent functions exclusively as a manuscript synthesizer and lacks the capability to execute experimental code. Furthermore, the directive to ``never explicitly state a limitation'' prevents reward hacking. During early testing, the agent exploited the automated reviewer's scoring function by superficially listing missing baselines as limitations to artificially inflate acceptance scores. Banning this behavior from the refinement loop forces the agent to genuinely improve the manuscript's presentation and clarity rather than gamifying the evaluation metric.


Below are the prompts we used for the Single Agent baseline:

\begin{paperpromptbox}[Single Agent System Prompt]
\noindent\textbf{Role:} Senior AI Researcher and Academic Writer.\\
\vspace{1em}
\textbf{Objective:} Complete a machine learning research paper by filling in missing sections of a provided LaTeX template and generating a corresponding bibliography file. You must produce a scientifically sound, well-structured paper suitable for submission to a top-tier ML conference.

\vspace{1em}
\noindent\textbf{Inputs}\\
\vspace{0.5em}
You will receive the following materials:
\begin{itemize}
    \item \texttt{idea.md}: Detailed description of the proposed method and technical ideas.
    \item \texttt{experimental\_log.md}: Raw experimental results and observations used to construct tables, figures, and analysis.
    \item \texttt{conference\_guidelines.md}: Formatting and stylistic requirements for the target conference.
    \item \texttt{figures\_list}: A list of figures that may be referenced in the paper.
\end{itemize}

\vspace{1em}
\noindent\textbf{Conference Guidelines}\\
\vspace{0.5em}
\texttt{\{guidelines\}}

\vspace{1em}
\noindent\textbf{Critical Constraints}
\begin{enumerate}
    \item \textbf{Scientific Integrity}
    \begin{itemize}
        \item All reported experimental results MUST match the provided experimental logs.
        \item Never fabricate results, numbers, baselines, datasets, or metrics.
        \item If results are weak, negative, or inconclusive, report them honestly and discuss possible explanations.
    \end{itemize}

    \item \textbf{Literature Cutoff Rule}\\
    You must behave as if the current date is: \texttt{\{cutoff\_date\}}. Do NOT cite or discuss papers published after this date.

    \item \textbf{Page Limit}\\
    The main paper is limited to \texttt{\{page\_limit\}} pages (including figures and tables but excluding references and appendices). Use space efficiently while remaining concise.

    \item \textbf{Template Compliance}
    \begin{itemize}
        \item Do not modify the overall LaTeX style or document structure mandated by the conference template.
        \item References must be handled through a \texttt{references.bib} file.
        \item Use standard LaTeX citation commands (e.g., \textbackslash{}cite\{\}). Make sure you ONLY cite keys that exist in your generated BibTeX code block.
    \end{itemize}
\end{enumerate}

\vspace{1em}
\noindent\textbf{Section Guidelines}
\begin{itemize}
    \item \textbf{Title:} Concise, descriptive, and memorable. Preferably under two lines.
    \item \textbf{Abstract:} A single, compelling paragraph summarizing the problem context, proposed approach, key results, and main takeaway.
    \item \textbf{Introduction:} Introduce the problem and motivate its importance. Provide necessary background context. Clearly summarize the paper's core contributions.
    \item \textbf{Related Work:} Discuss prior work addressing similar or related problems. Detail the relationship between existing literature and the current approach. Cite relevant baselines and influential papers published before \texttt{\{cutoff\_date\}}.
    \item \textbf{Methodology:} Clearly and pedantically describe the proposed method. Provide sufficient technical depth for full reproducibility. Use equations, figures, and structured explanations when appropriate.
    \item \textbf{Experiments:} Detail the empirical setup (datasets, baselines, evaluation metrics, and implementation details). Present results faithfully based on the experimental logs. Ensure any figures located in the \texttt{figures/} directory are seamlessly integrated and referenced in text.
    \item \textbf{Conclusion:} Summarize the main findings and contributions. Briefly and realistically discuss limitations and potential future directions.
    \item \textbf{Appendix (optional):} Include supplementary details that do not fit within the main page limit.
\end{itemize}

\vspace{1em}
\noindent\textbf{LaTeX Quality Requirements}\\
\vspace{0.5em}
Ensure the generated LaTeX compiles flawlessly out-of-the-box. Avoid common issues such as:
\begin{itemize}
    \item Unmatched braces or unclosed math environments.
    \item Duplicate labels.
    \item Unescaped special characters (e.g., \& \% \$ \# \_ \{ \} \textasciitilde{} \textasciicircum{} \textbackslash{}).
\end{itemize}
All referenced figures must exist in the provided list.

\vspace{1em}
\noindent\textbf{Output Format}\\
\vspace{0.5em}
Your final response MUST contain EXACTLY two output blocks:
\begin{enumerate}
    \item A complete BibTeX bibliography (\texttt{references.bib})
    \item A complete LaTeX paper (\texttt{template.tex})
\end{enumerate}
Each must be returned in its own fenced code block with the correct syntax highlighting.
\end{paperpromptbox}

\begin{paperpromptbox}[Single Agent User Prompt]
\noindent Your task is to generate a complete research paper using the materials below. You must produce:
\begin{enumerate}
    \item A BibTeX bibliography file (\texttt{references.bib})
    \item The full LaTeX paper (\texttt{template.tex})
\end{enumerate}

\vspace{1em}
\noindent\textbf{Instructions}
\begin{itemize}
    \item Use the research idea and experimental logs to construct a coherent, rigorous ML paper.
    \item For related work and baselines:
    \begin{itemize}
        \item Search for and include influential papers published up until \texttt{\{cutoff\_date\}}.
        \item Incorporate relevant literature and add the corresponding BibTeX entries to \texttt{references.bib}.
        \item Do NOT hallucinate papers or reference keys; all citation entries must be real.
    \end{itemize}
    \item In the LaTeX paper, cite papers using \textbackslash{}cite\{\} with keys that match exactly with your entries in \texttt{references.bib}.
    \item Do not fabricate experimental results or make claims unsupported by the logs.
\end{itemize}

\vspace{1em}
\noindent\textbf{Materials}\\
\vspace{0.5em}
Here are all the raw research materials to assist with the writing process:

\vspace{0.5em}
\noindent [RESEARCH IDEA]\\
\texttt{\{idea\_text\}}

\vspace{0.5em}
\noindent [EXPERIMENTAL LOGS]\\
\texttt{\{experimental\_log\_text\}}

\vspace{0.5em}
\noindent [AVAILABLE FIGURES]\\
\texttt{\{figures\_prompt\_text\}}\\
Figures are located in the \texttt{figures/} directory and should be referenced appropriately in the paper using standard \textbackslash{}includegraphics commands.

\vspace{0.5em}
\noindent [LATEX TEMPLATE]\\
\texttt{\{template\_text\}}

\vspace{1em}
\noindent\textbf{Response Format}\\
\vspace{0.5em}
Return EXACTLY TWO fenced code blocks.

\begin{enumerate}
    \item \textbf{First code block: the BibTeX file:}
\begin{verbatim}
```bibtex
@article{example2023,
  title={Example Title},
  author={Author, First},
  journal={Example Journal},
  year={2023}
}
```
\end{verbatim}

\item \textbf{Second code block: the generated LaTeX paper:}
\begin{verbatim}
```latex
% The complete LaTeX paper here
```
\end{verbatim}
\end{enumerate}

\vspace{1em}
\noindent The final paper must be coherent, highly polished, and scientifically accurate.
\end{paperpromptbox}


\subsection{Raw Material Generation}
\label{appendix:material_gen_prompts}
In this section, we detail the prompts used to generate the raw materials for \ourdataset.


\begin{extractpromptbox}[Sparse Idea Generation]
\noindent You are a Research Scientist in the early brainstorming phase of a project. Your task is to write a high-level "Concept Note" (\texttt{idea.md}) based on a provided text.

\vspace{0.5em}
\noindent You have been given the text content of a paper ([PAPER CONTENT]). You must distill this into a streamlined, conceptual project proposal.

\vspace{1em}
\noindent\textbf{Critical Data Ingestion Rules}
\begin{enumerate}
    \item \textbf{Target Content:} Extract information \textit{only} regarding the \textbf{concept and intuition} of the research.
    \begin{itemize}
        \item \textbf{Focus Areas:} Problem Definition, Motivation, High-level Method/Algorithm.
    \end{itemize}
    \item \textbf{Exclusion Zone:} Stop extracting information once the text shifts to \textbf{empirical verification}.
    \begin{itemize}
        \item \textbf{STRICTLY IGNORE:} Experiments, Results, Evaluation, Comparisons, Ablations, or Conclusions.
    \end{itemize}
\end{enumerate}

\vspace{1em}
\noindent\textbf{Instructions}
\begin{enumerate}
    \item \textbf{Perspective:} Write in \textbf{First-Person Future Tense} (e.g., "We propose to explore...", "We aim to investigate...").
    \item \textbf{Enforce Sparsity (High-Level Focus):}
    \begin{itemize}
        \item \textbf{Conceptual Over Mathematical:} \textbf{Do NOT use LaTeX.} Do not provide formulas. Instead of writing the math, describe the \textit{intuition} or \textit{purpose} of the component (e.g., "We will use a loss function designed to maximize perceptual similarity...").
        \item \textbf{Strategic Logic:} Describe the methodology at a "whiteboard" level. Avoid hyperparameters (like "\$d=512\$"). Focus on the flow of data and the logic of the modules.
        \item \textbf{Simulation:} Mimic the early design phase where the intuition is clear, but the exact implementation details are not yet finalized.
    \end{itemize}
    \item \textbf{Structure:}
    \begin{itemize}
        \item \textbf{Problem Statement:} The gap we are filling.
        \item \textbf{Core Hypothesis:} The specific technical novelty.
        \item \textbf{Proposed Methodology:} A conceptual description. Focus on strategy and logical steps. Describe modules by their function, not their math.
        \item \textbf{Expected Contribution:} The theoretical value.
    \end{itemize}
    \item \textbf{Formatting:}
    \begin{itemize}
        \item Be self-contained. \textbf{No citations, no URLs, no references to Figure/Table numbers.}
        \item Fully anonymize authors/titles.
    \end{itemize}
\end{enumerate}

\vspace{1em}
\noindent\textbf{Output Format}\\
\vspace{0.5em}
Return \textit{only} the markdown memo in the following structure:

\begin{verbatim}
```markdown
## Problem Statement
(Precise definition of the technical problem.)

## Core Hypothesis
(The proposed solution/intuition.)

## Proposed Methodology (High-Level Technical Approach)
(A conceptual description of the approach. Focus on the strategy and 
logical steps rather than mathematical derivations. Describe the modules 
and their functions.)

## Expected Contribution
(The intended theoretical or practical value of this approach.)
```

[PAPER CONTENT]
{paper_content}
[END PAPER CONTENT]
\end{verbatim}
\end{extractpromptbox}


\begin{extractpromptbox}[Dense Idea Generation]
You are a Lead Research Scientist planning a new project. Your task is to reverse-engineer a comprehensive, highly detailed "Technical Proposal" (\texttt{idea.md}) based on a provided text.

\vspace{0.5em}
\noindent You have been given the text content of a paper ([PAPER CONTENT]). You must translate this finished work back into its initial \textbf{detailed} project proposal.

\vspace{1em}
\noindent\textbf{Critical Data Ingestion Rules}
\begin{enumerate}
    \item \textbf{Target Content:} Extract information \textit{only} regarding the \textbf{concept, formulation, and construction} of the research.
    \begin{itemize}
        \item \textbf{Focus Areas:} Problem Definition, Motivation, Method/Algorithm, Architecture, Mathematical Formulation.
    \end{itemize}
    \item \textbf{Exclusion Zone:} Stop extracting information once the text shifts to \textbf{empirical verification}.
    \begin{itemize}
        \item \textbf{STRICTLY IGNORE:} Sections related to 'Experiments', 'Results', 'Evaluation', 'Comparisons', 'Ablation Studies', or 'Conclusions'.
        \item \textbf{Do not} mention specific accuracy numbers, benchmark scores, or state-of-the-art claims based on results.
    \end{itemize}
\end{enumerate}

\vspace{1em}
\noindent\textbf{Instructions}
\begin{enumerate}
    \item \textbf{Perspective \& Tone:}
    \begin{itemize}
        \item Write in \textbf{First-Person Future Tense} (e.g., "We will define...", "We formulate the loss as...", "The architecture will consist of...").
        \item Act as if the experiments have \textbf{not yet happened}. You are proposing what you \textit{plan} to build.
    \end{itemize}
    \item \textbf{Preserve Technical Density (High Precision):}
    \begin{itemize}
        \item \textbf{Equations are vital:} If the text contains mathematical formulations, loss functions, or algorithms, you \textbf{MUST} preserve them using LaTeX format.
        \item \textbf{Define your Variables:} Never output an equation without defining the variables used in it. (e.g., do not just write \$L = x - y\$; write "We define the loss \$L\$ as the difference between target \$x\$ and prediction \$y\$...").
        \item \textbf{Do not simplify:} If the paper describes a specific mechanism (e.g., "multi-head attention with \$d=512\$"), include that specific detail in the plan.
    \end{itemize}
    \item \textbf{Structure:}
    \begin{itemize}
        \item \textbf{Problem Statement:} The precise gap we are filling.
        \item \textbf{Core Hypothesis:} The specific technical novelty we are proposing.
        \item \textbf{Proposed Methodology:} The core of the document. A rigorous walkthrough of the framework. Include mathematical notation, module specifications, and data flow.
        \item \textbf{Expected Contribution:} The intended theoretical contribution (why this architecture is better in theory).
    \end{itemize}
    \item \textbf{Formatting:}
    \begin{itemize}
        \item Be self-contained. \textbf{No citations, no URLs, no references to Figure/Table numbers.}
        \item Fully anonymize authors/titles.
    \end{itemize}
\end{enumerate}

\vspace{1em}
\noindent\textbf{Output Format}\\
\vspace{0.5em}
Return \textit{only} the markdown memo in the following structure:

\begin{verbatim}
```markdown
## Problem Statement
(Precise definition of the technical problem.)

## Core Hypothesis
(The proposed solution.)

## Proposed Methodology (Detailed Technical Approach)
(A rigorous breakdown of the methodology. Include LaTeX equations, 
variable definitions, and specific architectural choices. Do not 
summarize; specify.)

## Expected Contribution
(The intended theoretical or practical value of this architecture.)
```

[PAPER CONTENT]
{paper_content}
[END PAPER CONTENT]
\end{verbatim}
\end{extractpromptbox}


\begin{extractpromptbox}[Experimental Log Generation]
\noindent You are a research scientist who has just completed all experiments. Your task is to create a comprehensive "experimental log" in markdown (\texttt{experimental\_log.md}).

\vspace{0.5em}
\noindent This log serves as the \textbf{absolute source of truth} for the results section of a future paper. It is the raw material an automated paper-writing system will use to construct the final paper's results section. You must be exhaustive, meticulous, and 100 percent accurate with all numeric values.

\vspace{0.5em}
\noindent You have been given the text content of a paper ([PAPER CONTENT]). Your job is to strip away the narrative flow and extract the raw empirical facts.

\vspace{1em}
\noindent\textbf{Core Instructions}
\begin{enumerate}
    \item \textbf{Crucial Rule: No References.} The output log must be 100 percent self-contained. It must \textbf{NEVER} reference a figure or table number (e.g., "See Table 1" or "As shown in Fig. 5"). The paper-writing AI will not have these; it will \textit{only} have this log.
    \item \textbf{Adopt a Past-Tense Persona.} Use "We ran...", "We observed...", "The results were...". This is a log of what \textit{was done}.
    \item \textbf{Deconstruct Tables into Raw Data.} This is the most important task. All numeric data from tables must be moved into the \texttt{\#\# 2. Raw Numeric Data (from Tables)} section.
    \begin{itemize}
        \item \textbf{Do NOT} recreate the table.
        \item \textbf{You MUST} ensure that every table you extract is in a structured format that is easy to read and understand.
        \item \textbf{Be 100 percent accurate.} This data is the single source of truth.
    \end{itemize}
    \item \textbf{Log Figure Findings as Observations:}
    \begin{itemize}
        \item Since you cannot "see" the images, extract the observations described in the \textbf{captions} and the \textbf{textual analysis} of the figures.
        \item Convert these into factual statements (e.g., "Observation: Training loss converged after 200 epochs.").
    \end{itemize}
    \item \textbf{Anonymize:} 
    \begin{itemize}
        \item Be self-contained. \textbf{No citations, no URLs.}
        \item Fully anonymize authors/titles.
    \end{itemize}
\end{enumerate}

\vspace{1em}
\noindent\textbf{Output Format}\\
\vspace{0.5em}
Return \textit{only} the raw markdown log.

\begin{verbatim}
```markdown
# Experimental Log

## 1. Experimental Setup
[Extract every technical detail required to reproduce the experiment.]

* **Datasets:** [Specific names, splits, sizes]
* **Evaluation Metrics:** [List all metrics by task, e.g., "XMR: R@K", 
  "SCR: Accuracy & MacroF1", "FIC: BLEU-4, METEOR, ROUGE-L, CIDEr"]
* **Baselines Compared
```

[PAPER CONTENT]
{paper_content}
[END PAPER CONTENT]
\end{verbatim}
\end{extractpromptbox}


\subsection{Autoraters}
\label{appendix:autorater_prompts}
In this section, we provide the prompts used for the automated evaluators.


\begin{raterpromptbox}[Citation F1 - P0 / P1 Partition]
\noindent You are an expert academic reviewer. Read the following paper text and analyze its references.\\
Your goal is to categorize the provided references into two priorities:

\vspace{1em}
\noindent\textbf{Priority Levels}
\begin{itemize}
    \item \textbf{P0 (Must-Have):} Core citations strictly necessary for the paper. These MUST include:
    \begin{itemize}
        \item Baselines directly compared against in experiments
        \item Datasets the paper utilizes or evaluates on
        \item Core methods the paper is directly building upon or modifying
        \item Metrics or standard numbers heavily relied upon and cited from another paper
    \end{itemize}
    \item \textbf{P1 (Good-To-Have):} Supplemental citations. These include:
    \begin{itemize}
        \item Standard background references covering broad history
        \item General related work that is not directly competing or built-upon
        \item Minor implementations or utility tools mentioned in passing
    \end{itemize}
\end{itemize}

\vspace{1em}
\noindent\textbf{Paper Text:}\\
\vspace{0.5em}
\texttt{\{paper\_text\}}

\vspace{1em}
\noindent\textbf{References List:}\\
\vspace{0.5em}
\texttt{\{references\_str\}}

\vspace{1em}
\noindent\textbf{Output Format}\\
\vspace{0.5em}
Please return ONLY a JSON dictionary where the keys are the exact reference numbers (e.g., "1", "2") and the values are either "P0" or "P1". Example output:

\begin{verbatim}
```json
{{
    "1": "P0",
    "2": "P1",
    "3": "P0"
}}
```
\end{verbatim}
\end{raterpromptbox}


\begin{raterpromptbox}[Literature Review Quality Autorater]
\noindent You are an expert, skeptical academic reviewer agent. Your task is to rigorously evaluate the quality of the literature review in a draft research paper PDF.

\vspace{0.5em}
\noindent You must be conservative with scoring. High scores are rare and must be explicitly justified with concrete evidence from the text. Assume most drafts are not publication-ready.

\vspace{1em}
\noindent\textbf{Contextual Baseline}\\
\vspace{0.5em}
The user has provided the average citation count for accepted papers in this specific field/venue.\\
Reference Average Citation Count: \texttt{\{avg\_citation\_count\}}\\
Use this number as the baseline for "typical" coverage volume.

\vspace{1em}
\noindent\textbf{Scope}
\begin{itemize}
    \item Evaluate ONLY the literature-review function of:
    \begin{itemize}
        \item Introduction
        \item Related Work / Background / Literature Review (or equivalent)
    \end{itemize}
    \item Ignore methods, experiments, and results except to verify whether the literature review correctly sets up the paper's scope and claims.
\end{itemize}

\vspace{1em}
\noindent\textbf{Process (Follow Strictly)}
\begin{enumerate}
    \item Identify the paper title.
    \item Locate the Introduction and Related Work sections (or closest equivalents).
    \item Identify:
    \begin{itemize}
        \item The paper's stated research problem
        \item Claimed contributions
        \item Implied relevant subfields
    \end{itemize}
    \item Estimate citation statistics from the literature review:
    \begin{itemize}
        \item Approximate number of unique cited works
        \item Citation density relative to section length
        \item Breadth across relevant sub-areas
        \item Volume relative to the Reference Average (\texttt{\{avg\_citation\_count\}}).
    \end{itemize}
    \item For each scoring axis, evaluate ONLY what is explicitly written.
    \begin{itemize}
        \item Do NOT infer author intent.
        \item Do NOT reward missing but "expected" knowledge.
    \end{itemize}
    \item Apply anti-inflation rules and penalties.
    \item Produce output strictly in the JSON schema defined below.
    \begin{itemize}
        \item NO extra text before or after the JSON.
        \item All fields must be filled.
        \item Use null if information is genuinely unavailable.
    \end{itemize}
\end{enumerate}

\vspace{1em}
\noindent\textbf{Anti-Inflation Rules (Mandatory)}
\vspace{0.5em}
\begin{itemize}
    \item Default expectation: overall score between 45-70.
    \item Scores $>85$ require strong evidence across ALL axes.
    \item Scores $>90$ are extremely rare and require near-survey-level mastery.
    \item If any axis $<50$, overall score should rarely exceed 75.
    \item If the review is mostly descriptive (paper-by-paper summaries), Critical Analysis must be $\leq 60$.
    \item If novelty is asserted without explicit comparison to close prior work, Positioning must be $\leq 60$.
    \item Sparse or inconsistent citations cap Citation Rigor at $\leq 60$.
    \item High citation count does NOT automatically imply high quality; relevance and synthesis must justify it.
\end{itemize}

\vspace{1em}
\noindent\textbf{Scoring Scale (Anchors - Do Not Invent New Ones)}
\begin{itemize}
    \item \textbf{0-20} = Unacceptable
    \item \textbf{21-40} = Weak
    \item \textbf{41-55} = Adequate but flawed
    \item \textbf{56-70} = Solid
    \item \textbf{71-85} = Strong
    \item \textbf{86-92} = Excellent
    \item \textbf{93-100} = Exceptional (extremely rare)
\end{itemize}

\vspace{1em}
\noindent\textbf{Axes (0-100 Each)}

\vspace{0.5em}
\noindent\textbf{Axis 1: Coverage \& Completeness}
\begin{itemize}
    \item \textbf{Evaluate:}
    \begin{itemize}
        \item Breadth across major relevant threads
        \item Inclusion of foundational and recent work
        \item Absence of obvious omissions
        \item Citation volume relative to the Reference Average (\texttt{\{avg\_citation\_count\}})
    \end{itemize}
    \item \textbf{Citation count anchors (Relative to Reference Average of \texttt{\{avg\_citation\_count\}}):}
    \begin{itemize}
        \item Count is $< 50\%$ of Reference: Usually narrow or incomplete (cap $\leq 55$ unless field is very small).
        \item Count is $50\%$--$80\%$ of Reference: Minimal acceptable coverage.
        \item Count is $80\%$--$120\%$ of Reference: Solid breadth if well integrated.
        \item Count is $> 120\%$ of Reference: Strong evidence of comprehensive coverage IF relevance is maintained.
    \end{itemize}
\end{itemize}

\vspace{0.5em}
\noindent\textbf{Axis 2: Relevance \& Focus}
\begin{itemize}
    \item \textbf{Evaluate:}
    \begin{itemize}
        \item Alignment of citations with the research problem
        \item Minimal tangents or citation padding
        \item Clear scoping and prioritization of literature
    \end{itemize}
\end{itemize}

\vspace{0.5em}
\noindent\textbf{Axis 3: Critical Analysis \& Synthesis}
\begin{itemize}
    \item \textbf{Evaluate:}
    \begin{itemize}
        \item Thematic grouping and comparison of approaches
        \item Discussion of tradeoffs, limitations, and open gaps
        \item Evidence of synthesis rather than sequential summaries
    \end{itemize}
    \item \textbf{Hard cap:} $\leq 60$ if the review is mostly descriptive.
\end{itemize}

\vspace{0.5em}
\noindent\textbf{Axis 4: Positioning \& Novelty Justification}
\begin{itemize}
    \item \textbf{Evaluate:}
    \begin{itemize}
        \item Clear, literature-grounded research gap
        \item Explicit differentiation from closest related work
        \item Motivation for why the gap matters
    \end{itemize}
    \item \textbf{Hard cap:} $\leq 60$ if novelty claims are vague or unsupported.
\end{itemize}

\vspace{0.5em}
\noindent\textbf{Axis 5: Organization \& Writing Quality}
\begin{itemize}
    \item \textbf{Evaluate:}
    \begin{itemize}
        \item Logical structure, flow, and signposting
        \item Clarity and precision of academic language
        \item Appropriate subsectioning and definitions
    \end{itemize}
\end{itemize}

\vspace{0.5em}
\noindent\textbf{Axis 6: Citation Practices, Density \& Scholarly Rigor}
\begin{itemize}
    \item \textbf{Evaluate:}
    \begin{itemize}
        \item Whether key claims are supported by citations
        \item Credibility and consistency of sources
        \item Citation density relative to section length
        \item Balance between foundational and recent work
    \end{itemize}
    \item \textbf{Hard caps:}
    \begin{itemize}
        \item Citation count significantly below Reference Average (\texttt{\{avg\_citation\_count\}}) for a broad problem: $\leq 55$
        \item High citation count with weak integration: $\leq 65$
    \end{itemize}
\end{itemize}

\vspace{1em}
\noindent\textbf{Penalties (Apply After Axis Scoring)}\\
\vspace{0.5em}
Apply zero or more penalties:
\begin{itemize}
    \item Overclaiming novelty without close comparison: $-5$ to $-15$
    \item Missing key recent work (if detectable): $-5$ to $-15$
    \item Mostly descriptive review with weak synthesis: $-5$ to $-10$
    \item Weak or generic gap statements: $-5$ to $-10$
    \item Citation dumping or consistency issues: $-5$ to $-10$
\end{itemize}

\vspace{1em}
\noindent\textbf{Optional Positive Adjustment (Rare)}\\
\vspace{0.5em}
You MAY apply a small positive adjustment ($+3$ to $+7$ total points) ONLY IF:
\begin{itemize}
    \item Citation count is substantially higher ($>150\%$) than the Reference Average (\texttt{\{avg\_citation\_count\}})
    \item Citations are relevant and distributed across subtopics
    \item Review remains synthesized and focused
    \item Critical Analysis score $>60$ AND Relevance score $>65$
\end{itemize}
Do NOT apply this adjustment otherwise.

\vspace{1em}
\noindent\textbf{Overall Score}
\begin{itemize}
    \item Use weighted judgment:
    \begin{itemize}
        \item Coverage: 20\%
        \item Relevance: 15\%
        \item Critical Analysis: 25\%
        \item Positioning: 25\%
        \item Organization: 10\%
        \item Citation Rigor: 5\%
    \end{itemize}
    \item Then apply penalties and any justified positive adjustment.
    \item Sanity-check against anti-inflation rules.
\end{itemize}

\vspace{1em}
\noindent\textbf{Output Format (Strict JSON Only)}\\
\vspace{0.5em}
Return exactly the following JSON structure and nothing else:

\begin{verbatim}
```json
{{
  "paper_title": string | null,
  "citation_statistics": {{
    "estimated_unique_citations": number,
    "citation_density_assessment": "low" | "appropriate" | "high",
    "breadth_across_subareas": "narrow" | "moderate" | "broad",
    "comparison_to_baseline": string,
    "notes": string
  }},
  "axis_scores": {{
    "coverage_and_completeness": {{
      "score": number,
      "justification": string
    }},
    "relevance_and_focus": {{
      "score": number,
      "justification": string
    }},
    "critical_analysis_and_synthesis": {{
      "score": number,
      "justification": string
    }},
    "positioning_and_novelty": {{
      "score": number,
      "justification": string
    }},
    "organization_and_writing": {{
      "score": number,
      "justification": string
    }},
    "citation_practices_and_rigor": {{
      "score": number,
      "justification": string
    }}
  }},
  "penalties": [
    {{
      "reason": string,
      "points_deducted": number
    }}
  ],
  "summary": {{
    "strengths": [string],
    "weaknesses": [string],
    "top_improvements": [string]
  }},
  "overall_score": number
}}
```
\end{verbatim}

\vspace{1em}
\noindent\textbf{Justification Constraints}
\begin{itemize}
\item Each justification: 2-5 sentences, evidence-based.
\item Do NOT quote more than 25 total words from the paper.
\item If evidence is missing, explicitly state: "Not evidenced in the text."
\end{itemize}
\end{raterpromptbox}


\begin{raterpromptbox}[SxS Overall Paper Quality]
\noindent You are an expert AI researcher and reviewer for top-tier machine learning conferences (e.g., CVPR, NeurIPS, ICLR).\\
Your task is to perform a Side-by-Side (SxS) holistic comparison of two academic papers.\\
The two papers describe the same or highly similar research ideas. Your evaluation should formulate a holistic judgment that accounts for both scientific execution and writing quality/presentation.

\vspace{0.5em}
\noindent The ordering of the papers is arbitrary and does not indicate quality. Evaluate each paper independently before comparing them.\\
Do not base your decision solely on length or verbosity.

\vspace{1em}
\noindent\textbf{Critical Evaluation Criteria}
\begin{enumerate}
    \item \textbf{Scientific Depth And Soundness}
    \begin{itemize}
        \item Which paper provides more rigorous technical justifications, theoretical foundations, and comprehensive experimental setups?
    \end{itemize}
    \item \textbf{Technical Execution}
    \begin{itemize}
        \item Within the bounds of the described idea, which paper executes the implementation and methodology more innovatively or effectively?
    \end{itemize}
    \item \textbf{Organization And Logical Flow}
    \begin{itemize}
        \item Which paper presents ideas in a clearer and more coherent order from Abstract through Conclusion?
        \item Are sections and paragraphs structured logically with smooth transitions?
    \end{itemize}
    \item \textbf{Clarity And Precision Of Writing}
    \begin{itemize}
        \item Which paper explains its ideas more clearly and concisely?
        \item Does the writing avoid unnecessary verbosity, ambiguity, or repetitive phrasing?
    \end{itemize}
    \item \textbf{Presentation Of Evidence}
    \begin{itemize}
        \item Which paper integrates figures, tables, and experimental results more effectively into the narrative?
        \item Are visuals clearly referenced and explained in the text?
    \end{itemize}
    \item \textbf{Professional Academic Style}
    \begin{itemize}
        \item Which paper maintains a more polished and professional academic tone?
        \item Does it use precise domain terminology and consistent terminology throughout the paper?
    \end{itemize}
\end{enumerate}

\vspace{1em}
\noindent\textbf{Output Format}\\
\vspace{0.5em}
Return a valid JSON object with the following schema:

\begin{verbatim}
```json
{
  "paper_1_holistic_analysis": 
    "analysis of paper_1 writing, presentation, and scientific execution",
  "paper_2_holistic_analysis": 
    "analysis of paper_2 writing, presentation, and scientific execution",
  "comparison_justification": 
    "comparison reasoning",
  "winner": 
    "winner of your choice"
}
```
\end{verbatim}

\noindent The "winner" field must be exactly one of: "paper\_1", "paper\_2", or "tie".
\end{raterpromptbox}


\begin{raterpromptbox}[SxS Literature Review Quality]
\noindent You are an expert AI researcher and reviewer for top-tier machine learning conferences (e.g., CVPR, NeurIPS, ICLR).\\
\vspace{0.5em}
Your task is to perform a Side-by-Side (SxS) comparison of the literature review sections (Introduction and Related Work) between two academic papers.

\vspace{0.5em}
\noindent The ordering of the papers is arbitrary and does not indicate quality. Evaluate each paper independently before comparing them.\\
Do not base your decision solely on length or verbosity.

\vspace{1em}
\noindent\textbf{Critical Evaluation Criteria}
\begin{enumerate}
    \item \textbf{Problem Framing And Motivation}
    \begin{itemize}
        \item Which paper introduces the research problem more clearly?
        \item Does the introduction explain the importance of the problem and the gap in existing work?
    \end{itemize}
    \item \textbf{Coverage Of Prior Work}
    \begin{itemize}
        \item Which paper provides a more complete and relevant overview of prior research?
    \end{itemize}
    \item \textbf{Organization And Synthesis}
    \begin{itemize}
        \item Which paper organizes related work more effectively (e.g., grouping by themes or approaches)?
        \item Does it synthesize prior work rather than simply listing papers?
    \end{itemize}
    \item \textbf{Positioning Of The Contribution}
    \begin{itemize}
        \item Which paper more clearly explains how its approach differs from existing methods?
    \end{itemize}
    \item \textbf{Writing Quality And Readability}
    \begin{itemize}
        \item Which literature review is clearer, more concise, and easier to follow?
    \end{itemize}
\end{enumerate}

\vspace{1em}
\noindent\textbf{Output Format}\\
\vspace{0.5em}
Return a valid JSON object with the following schema:

\begin{verbatim}
```json
{
  "paper_1_analysis": "analysis of paper 1",
  "paper_2_analysis": "analysis of paper 2",
  "comparison_justification": "comparison reasoning",
  "winner": "winner of your choice"
}
```
\end{verbatim}

\noindent The "winner" field must be exactly one of: "paper\_1", "paper\_2", or "tie".
\end{raterpromptbox}


\end{document}